%% file: iclr2027_conference.tex
\documentclass{article} 
\usepackage{iclr2027_conference,times}

\input{math_commands.tex}

\usepackage[utf8]{inputenc} 
\usepackage[T1]{fontenc}    
\usepackage{hyperref}       
\usepackage{url}            
\usepackage{booktabs}       
\usepackage{amsfonts}       
\usepackage{nicefrac}       
\usepackage{microtype}      
\usepackage{xcolor}         

\usepackage[dvipsnames]{xcolor}
\usepackage{colortbl}
\usepackage{multirow}
\usepackage{graphicx}
\usepackage{amsmath}
\usepackage{amssymb}
\usepackage{booktabs}
\usepackage{multirow}
\usepackage{bm}
\usepackage{comment}
\usepackage{color, colortbl}
\usepackage{xcolor}
\usepackage{hyperref}
\usepackage{cleveref}
\definecolor{Gray}{gray}{0.95}
\newcolumntype{g}{>{\columncolor{Gray}}c}

\newcommand{\ie}{\textit{i}.\textit{e}.}
\newcommand{\eg}{\textit{e}.\textit{g}.}
\usepackage{pifont}
\usepackage{wrapfig}
\usepackage{caption}
\usepackage{subcaption}
\def\paper{AnyGroundBench}
\definecolor{NavyBlue}{rgb}{0.0, 0.0, 0.5}
\usepackage{amssymb}      
\usepackage{fontawesome5}
\usepackage{etoc}
\usepackage[most]{tcolorbox}
\definecolor{iclgray}{gray}{0.92}
\definecolor{DeepIndigo}{HTML}{4E27A2}
\definecolor{RoyalNavy}{HTML}{143690}
\definecolor{ForestGreen}{HTML}{3B8550}
\definecolor{OceanBlue}{HTML}{34768F}
\definecolor{DeepViolet}{HTML}{301186}
\definecolor{AnimalColor}{HTML}{017B9B}
\definecolor{IndustryColor}{HTML}{00C0BA}
\definecolor{SportsColor}{HTML}{FD8BA8}
\definecolor{SurgeryColor}{HTML}{FDBD6C}
\definecolor{SecurityColor}{HTML}{8F6F55}
\definecolor{zeroshotColor}{HTML}{BD1D1C}
\definecolor{twoshotColor}{HTML}{1472BC}
\definecolor{gtshotColor}{HTML}{438539}
\usepackage{tabularx}
\usepackage{makecell}

\definecolor{negred}{HTML}{D98B8E}
\definecolor{purple}{HTML}{8991B7} %
\newcommand{\poscol}[1]{\textcolor{purple}{\textbf{#1}}}
\newcommand{\negcol}[1]{\textcolor{negred}{\textbf{#1}}}

\newcommand{\appref}[1]{\hyperref[#1]{Appendix~\ref*{#1}}}
\definecolor{BrickRed}{rgb}{0.6, 0.1, 0.1}

\hypersetup{
    colorlinks=true,
    citecolor=MidnightBlue,
    linkcolor=Maroon,
    urlcolor=MidnightBlue,
}

\title{A Multi-Domain Adaptation Benchmark for Video Grounding in VLMs}

\author{%
  Rintaro Otsubo$^{1,2*}$ \quad Ryo Fujii$^{1,2*}$ \quad Reina Ishikawa$^{1,2}$ \quad Taiki Kanaya$^{1,2}$ \quad Kanta Sawafuji$^{1,2}$ \\
  \textbf{Hiroki Kajita}$^{1,3}$ \quad \textbf{Shigeki Sakai}$^{1,3}$ \quad \textbf{Hideo Saito}$^{1,2}$ \quad \textbf{Ryo Hachiuma}$^{4}$ \\
  $^1$Keio University \quad $^2$Keio AI Research Center
  $^3$Keio University School of Medicine \quad $^4$NVIDIA \\
  \small $^*$Equal contribution
}

\iclrfinalcopy 
\begin{document}

\maketitle

\begin{center}
  \small
  \href{https://rinost081.github.io/AnyGroundBench-page}{\faGlobe\ Project Page}
  \quad
  \href{https://github.com/rinost081/AnyGroundBench}{\faGithub\ GitHub}
  \quad
  \href{https://huggingface.co/datasets/rinost081/AnyGroundBench}{\faDatabase\ Hugging Face}
\end{center}

\begin{abstract}
Vision-Language Models (VLMs) have shown strong performance in Spatio-Temporal Video Grounding (STVG), yet they are still evaluated mostly in a zero-shot manner on general-purpose benchmarks of everyday scenes. This creates a critical disconnect from real-world applications in specialized domains, where models inevitably encounter rare visual or textual concepts. Since exhaustive pre-training across infinite data distributions is infeasible, the ability to adapt to novel domains with limited data is essential. To bridge this gap, we introduce \paper, a domain-adaptation benchmark designed to shift the STVG evaluation paradigm from static zero-shot testing to rigorous domain adaptation. Targeting five specialized domains (animal, industry, sports, surgery, and public security), \paper\ pairs newly captured, expert-annotated videos with established datasets, unifying them through dense, high-fidelity spatio-temporal annotations. Crucially, the benchmark provides dedicated limited training subsets, enabling systematic evaluation of domain adaptability under limited training data. We benchmark 23 state-of-the-art VLMs in the zero-shot setting and further evaluate five adaptation strategies, spanning training-free and fine-tuning-based approaches, on representative models, assessing their zero-shot generalization and adaptation capacity. Our results show that current VLMs remain far from practical performance in the zero-shot setting, that training-free adaptation produces highly variable effects depending on the model and domain, and that fine-tuning-based adaptation, though more effective, still falls short of real-world requirements, with gains varying markedly across domains. These findings expose fundamental limitations in current VLMs' spatio-temporal reasoning, pointing to concrete directions for future research.
\end{abstract}

\begin{figure*}[h] 
\centering
\includegraphics[width=\linewidth]{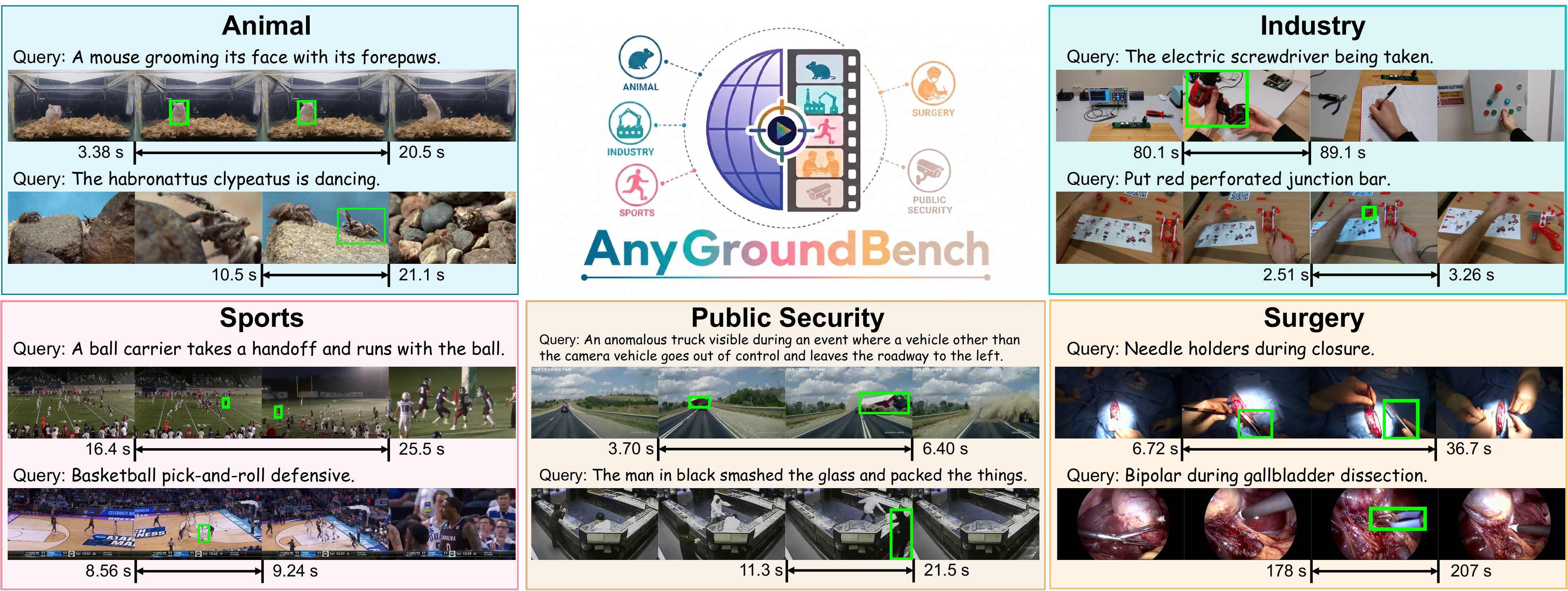}
\vspace{-1.em}
\caption{\textbf{\paper\ examples across five specialized domains.} \paper\ integrates newly captured, expert-annotated videos with established public datasets, unifying them through dense, new high-fidelity spatio-temporal annotations and language queries.}
\label{fig:overview}
\vspace{-1.em}
\end{figure*}

\section{Introduction}

Spatio-Temporal Video Grounding (STVG) is a fundamental task in the video perception field that requires localizing a target object across both spatial and temporal dimensions in the video based on natural language queries. Its importance is underscored by its role in high-stakes applications, such as video retrieval~\citep{gu2024context},  video reasoning~\citep{cheng2025vstar,meng2025openo3}, and autonomous driving~\citep{Zeng2025futuresightdrive}. Recently, the field has transitioned from specialized model architectures~\citep{Su2021STVGBert,Tang2022HC-STVGv2,Wasim2024VideoGrounding-DINO,Zhang2020VidSTG,Kim_2023_WACV,Gu2025Knowing,Gu2026ARTSTVG} toward general Vision-Language Models (VLMs)~\citep{ahmad2025videomolmo,gu2025thinking,Li2025LLaVA-ST,Li2024groundinggpt,Vidi2026vidi25,Wang2026SpaceVLLM,zhang2026stvgr,Heo2025CVPR,Pramanick_2025_ICCV}. By incorporating the grounding capability in a single unified model, VLMs can extend this capability to tasks that require both perception and reasoning, such as grounded video question answering~\citep{Xiao_2024_CVPR,shen2025zoomzero,yan2025videochatr15}, grounded reasoning~\citep{meng2025openo3}, and agentic tool-calling~\citep{li2026videothinkerbuilds}.


Despite recent progress, most established benchmarks~\citep{Chen2019Weakly,gao2025omniground,Kurita2023RefEgo,Li2025LLaVA-ST,Liang2025EgoMask,shi2026videoloom,Tang2022HC-STVGv2,xu2025tog,Yamaguchi2017STPR,yao2026omnistvg,Zhang2020VidSTG} remain predominantly focused on everyday scenarios and general objects, as summarized in \autoref{tab:dataset_comparison}. However, real-world deployments inevitably extend beyond such settings. For example, the ability to spatio-temporally localize specific mouse behaviors, such as scratching, is crucial for dermatology or neurology research~\citep{Segalin2021MARS}. In such highly specialized domains, both the visual appearance and the textual representation deviate significantly from those found in everyday activities. This naturally raises a fundamental question: \textit{Can existing VLMs effectively adapt to ground anything in these rare domains?}

Addressing this question is hindered not only by the lack of benchmarks for specialized domains but also by evaluation protocols, which rely excessively on zero-shot performance~\citep{Yang2025Unleashing}. While VLMs have demonstrated remarkable foundational power, it is practically infeasible for any model to \emph{pre-learn} the near-infinite variety of data distributions encountered, or to be encountered in the future, in the wild. As novel domains and tasks continuously emerge in the real world, such as the aforementioned laboratory or industrial settings~\citep{Francesco2023MECCANO,Ragusa2024Enigma51}, a model's capability cannot be measured solely by its static pre-trained knowledge. Instead, the ability to seamlessly adapt to a rare domain becomes a strict requirement for real-world deployment. Therefore, to understand whether a model can perceive data from specialized domains, the evaluation paradigm must shift from testing zero-shot performance to measuring adaptability. Moreover, due to the inherent difficulty of collecting data in specialized domains, adaptation should be conducted under \emph{limited-data} conditions.

To bridge these two crucial gaps in domain coverage and adaptability evaluation, we introduce \textbf{\paper}, a domain-adaptation benchmark that shifts the evaluation paradigm for STVG from general zero-shot evaluation to adaptation to specialized domains, as illustrated in \autoref{fig:overview}. \paper\ comprises $2{,}042$ videos, natural language queries, and corresponding spatio-temporal bounding boxes. \paper\ offers two key advantages. (1) \textbf{Diverse and High-Quality Data in Specialized Domains.} \paper\ targets five specialized domains: animal, industry, sports, surgery, and public security. For data sourcing, we aggregate videos from entirely novel captures (including American football from the sports domain and medical expert-curated mouse scratching from the animal domain) alongside new annotations added to existing public datasets. To ensure high-fidelity annotations, we manually provide dense spatio-temporal bounding boxes for ungrounded videos and transform existing labels for the STVG task if applicable. (2) \textbf{Dedicated Training Sets for Domain Adaptation.} Unlike previous benchmarks that rely exclusively on general-domain training, \paper\ provides tailored training subsets for each domain. This design explicitly enables researchers to systematically evaluate VLMs' limited-data adaptability: how effectively they adjust to the unique spatio-temporal dynamics inherent to specialized domains.

We evaluate $23$ VLMs on our benchmark, including both open-source and proprietary models with diverse architectures and model sizes. We further decompose STVG tasks into isolated Spatial Video Grounding (SVG), which focuses on spatial localization within ground-truth time spans, and Temporal Video Grounding (TVG), which targets temporal boundary detection, thereby pinpointing precisely where current spatio-temporal reasoning breaks down. Furthermore, to serve as a reference for future benchmark users, we evaluate a suite of adaptation strategies spanning training-free approaches and fine-tuning-based approaches, covering both practical backpropagation-free settings and higher-performing but more resource-intensive settings.

We reveal three critical findings about VLMs' grounding capability in specialized domains:
\begin{itemize}
    \item \textbf{Limited Spatio-Temporal Grounding Capability of Current VLMs:} Even the latest proprietary VLMs (\eg, GPT-6-Astra) fall short of practical STVG performance, while most open-source VLMs remain near zero in most domains, lagging far behind their proprietary counterparts. This highlights the need to improve the STVG capability of VLMs in general, and of open-source VLMs in particular.
    \item \textbf{Bottleneck Shifts from Spatial to Joint Spatio-Temporal Grounding:} For open-source and earlier proprietary models, SVG lags far behind TVG in most domains, making spatial grounding the primary bottleneck. The strongest proprietary models perform well on both SVG and TVG, yet their STVG remains far below both, indicating that jointly aligning spatial and temporal predictions becomes the limiting factor.
    \item \textbf{Limited and Inconsistent Gains from Adaptation:} Training-free adaptation yields inconsistent effects across models and domains. Fine-tuning improves temporal and spatial grounding individually, with gains that vary greatly across domains. However, these improvements lead to only small gains in STVG. This suggests the need for new adaptation methods that robustly improve grounding across domains.
\end{itemize}

\begin{table}[tb]
\centering
\caption{\textbf{Comparison with existing video grounding datasets.}
\paper is the only dataset that spans multiple specialized domains while providing a training set for spatio-temporal grounding adaptation.
\faDatabase\ Public Dataset,\ \faGlobe\ Internet-Collected,\ \faCamera\ Self-Collected.
}
\label{tab:dataset_comparison}
\resizebox{\linewidth}{!}{%
\begin{tabular}{l l c c c c c c c c}
\toprule
\textbf{Dataset} & \textbf{Domain} & \textbf{\# Videos} & \textbf{Total Duration} & \textbf{\# Queries} & \textbf{Data Source} & \textbf{SVG} & \textbf{TVG} & \textbf{STVG} & \textbf{Train Set} \\
\midrule
\multicolumn{10}{l}{\textit{General Domain}} \\
\midrule
TACoS~\citep{Regneri2013TACoS}            & Cooking          & 0.1K & 10.1h & 18.2K & \faDatabase &    & \checkmark &    & \checkmark \\
Charades-STA~\citep{Gao2018Charades-STA}     & Activity         & 6.7K & 57.1h & 16.1K & \faDatabase &    & \checkmark &    & \checkmark \\
YouCook2~\citep{Zhou2018YouCook2}         & Cooking          & 2.0K & 176h  & 13.8K & \faDatabase &    & \checkmark &    & \checkmark \\
VidSTG~\citep{Zhang2020VidSTG}           & General          & 6.9K & 69.1h & 99.9K & \faDatabase & \checkmark & \checkmark & \checkmark & \checkmark \\
HC-STVG~\citep{Tang2022HC-STVGv2}         & Human Action     & 5.7K & 31.4h & 5.7K  & \faDatabase & \checkmark & \checkmark & \checkmark & \checkmark \\
VUE-STG~\citep{Vidi2025vidi2}       & General          & 0.9K & 205h & 1.6K  & \faGlobe    & \checkmark & \checkmark & \checkmark &    \\
OmniGround~\citep{gao2025omniground}       & General          & 3.5K & 17.6h & 3.5K  & \faDatabase\,\faGlobe    & \checkmark & \checkmark & \checkmark &    \\
\midrule
\multicolumn{10}{l}{\textit{Specialized Domain}} \\
\midrule
TUMTraffic~\citep{zhou2025tumtraf}   & Roadside                                                   & 1.0K &      & 5.7K  & \faCamera   & \checkmark & \checkmark & \checkmark & \checkmark \\
UAV-SVG~\citep{Zhan2025UAV-SVG}              & Aerial / Drone                                             & 3.6K & 18.7h & 17.8K & \faDatabase & \checkmark &    &    & \checkmark \\
QV-M$^2$~\citep{cao2025when}             & Vlog / News                                                & 1.3K &   55.9h   & 2.2K  & \faDatabase &    & \checkmark &    &    \\
\rowcolor{iclgray}
\textbf{\paper (Ours)}  & \makecell[l]{Animal / Industry / Sports \\ Surgery / Public Security}     & 2.0K & 18.3h & 3.5K  & \faCamera\,\faDatabase & \checkmark & \checkmark & \checkmark & \checkmark \\
\bottomrule
\end{tabular}%
}\vspace{-1.0em}
\end{table}

\section{Related Work}
\noindent \textbf{Benchmark for Spatio-Temporal Video Grounding.} 
Spatio-temporal video grounding (STVG) aims to localize a target object specified with the text query in both the space and time axes. Progress in this field has been heavily driven by large-scale datasets~\citep{Chen2019Weakly,Kurita2023RefEgo,Liang2025EgoMask,xu2025tog,Yamaguchi2017STPR,yao2026omnistvg,Zhang2020VidSTG}. Traditionally, benchmarks like VidSTG~\citep{Zhang2020VidSTG} (built upon VidOR~\citep{shang2019VidOR}) and HC-STVG~\citep{Tang2022HC-STVGv2} (sourced from AVA~\citep{Gu2018AVA} and YouTube) have served as primary testbeds, focusing mainly on generic objects and human-centric activities. More recently, LLaVA-ST~\citep{Li2025LLaVA-ST} enhanced VidSTG with LLM-generated text queries, and OmniGround~\citep{gao2025omniground} was introduced to cover a wider range of object categories by collecting videos from Pexels\footnote{\url{https://www.pexels.com}} and RVOS~\citep{Ding2023MeViS,Seo2020URVOS}. 

However, existing benchmarks share a major limitation: owing to their reliance on general internet platforms and existing generic datasets, they remain heavily confined to common daily-life domains. This narrow scope leaves a critical void for evaluating models in highly specialized, real-world scenarios. To bridge this gap, \paper\ explicitly targets specialized fields. Crucially, we differentiate our work by combining \textit{newly captured} videos curated from experts with established domain-specific datasets with unified, high-fidelity annotations. This exclusive combination provides a benchmark for evaluating STVG domain-adaptation capabilities.

\noindent \textbf{Spatio-Temporal Video Grounding via VLMs.} 
Recently, VLMs have dominated not only video understanding tasks~\citep{fu2025videomme,zhou2024mlvu} but also video perception tasks, such as STVG~\citep{ahmad2025videomolmo,Li2024groundinggpt,Li2025LLaVA-ST,Vidi2026vidi25,Wang2026SpaceVLLM,zhang2026stvgr}. Yet, current evaluation protocols for these models rely almost exclusively on zero-shot performance on the general-domain benchmarks~\citep{Yang2025Unleashing}. While measuring foundational knowledge is useful, it ignores the reality of real-world deployments. Since it is impossible to pre-train a model on every conceivable domain, practical systems must rapidly adapt to unseen, specialized data. Currently, the field lacks standardized and unified benchmarks designed to explicitly measure domain adaptability (\eg, via In-Context Learning (ICL)~\citep{brown2020language,dong2026demoicl,Fujii2026Viola,Kim2025VideoICL, rubin2022learning, xie2022an,xue2026personal,yu2024eliciting}). \paper\ is the first benchmark to fill both of these critical gaps by providing highly specialized target domains equipped with standardized adaptation training sets.

\begin{table}[tb]
    \centering
    \caption{\textbf{Overview of \paper.} Our benchmark emphasizes domain diversity and high-fidelity manual annotations for specialized scenarios. Avg. Dur. is the mean video length per vid. \faVideo: Newly captured videos, \faArrowsAltH: Temporal span annotation, \faClone[regular]: Spatial box annotation, \faPen: Text query annotation.}
    \label{tab:dataset_overview}
    \resizebox{0.8\linewidth}{!}{%
    \begin{tabular}{lcccccccc}
    \toprule
     \multirow{2.5}{*}{Domain} & \multirow{2.5}{*}{\# Videos} & \multirow{2.5}{*}{Avg. Dur. (sec)} & \multicolumn{2}{c}{\# QA Pairs} & \multicolumn{4}{c}{Additional Efforts} \\
     \cmidrule(lr){4-5} \cmidrule(lr){6-9}
    & & & Train & Test & \faVideo & \faArrowsAltH & \faClone[regular]  & \faPen  \\

    \midrule
    \textcolor{AnimalColor}{Animal} & 286 & 47 & 583 & 157 & \checkmark & \checkmark & \checkmark & \checkmark \\
    \textcolor{IndustryColor}{Industry} & 254 & 55 & 410 & 169 & & & \checkmark & \checkmark \\
    \textcolor{SportsColor}{Sports} & 557 & 20 & 460 & 163 & \checkmark & \checkmark & \checkmark & \checkmark \\
    \textcolor{SurgeryColor}{Surgery} & 193 & 79 & 559 & 216 & & & & \checkmark \\
    \textcolor{SecurityColor}{Public Security} & 752 & 16 & 558 & 250 & & & \checkmark & \\
    \bottomrule
    \end{tabular}%
    }\vspace{-1.0em}
\end{table}

\section{\paper}
We introduce \paper, a benchmark designed to evaluate STVG capability as well as the adaptation across five specialized domains that present distinct challenges and high real-world relevance: animal, industry, sports, surgery, and public security. To evaluate the adaptability, \paper\ provides training subsets for each domain. An overview of the selected domains and statistics are summarized in \autoref{tab:dataset_overview}. Details are provided in \appref{app:datasetdetails}.

\subsection{Benchmark Tasks}
\label{subsec:benchmarktasks}
\paper\ evaluates models on three interconnected tasks defined over a video and a corresponding text query in a zero-shot and via adaptation with limited data. The same VLM is used across all three tasks; the task is induced solely by a task-specific system prompt. We first introduce a unified notation to formalize each task.

\noindent \textbf{Notation.}
Let $V = \{I_t\}_{t=1}^{T}$ denote a video of $T$ frames with $I_t \in \mathbb{R}^{H \times W \times 3}$, and let $Q$ be a natural-language query referring to a target object and its associated event. A per-frame axis-aligned bounding box\footnote{Several VLMs (\eg, the Gemini or Qwen series) emit bounding-box predictions only at a discrete subset of timestamps $\{t_l\}_{l=1}^{L} \subseteq [\hat{t}_s, \hat{t}_e]$ rather than at every frame.} is written as $b_t  \in \mathbb{R}^{4}$, and a spatio-temporal tube on an interval $[t_s, t_e] \subseteq [1, T]$ as
\begin{equation}
\tau \;=\; \bigl\{\,(t, b_t)\,\bigr\}_{t=t_s}^{t_e}.
\end{equation}
Each test instance is paired with a ground-truth tube $\tau^{*} = \{(t, b_t^{*})\}_{t=t_s^{*}}^{t_e^{*}}$. We denote the evaluated VLM by $\mathcal{F}_\theta$ with parameters $\theta$. All three tasks invoke the same $\mathcal{F}_\theta$,
\begin{equation}
\hat{y} \;=\; \mathcal{F}_\theta(V, Q;\, p),
\label{eq:model}
\end{equation}
where $p \in \{p_{\text{STVG}},\, p_{\text{SVG}},\, p_{\text{TVG}}\}$ is the task-specific system prompt (full templates in \appref{app:prompts}) and $\hat{y}$ is the output type appropriate to the task. The three tasks correspond to estimating different subsets of $\tau^{*}$ from $(V, Q)$.

\noindent \textbf{Spatio-Temporal Video Grounding (STVG)} requires localizing the target object matching the query in both the spatial and temporal axes. Given $(V, Q)$ and the system prompt $p_{\text{STVG}}$, the model predicts the full tube as:
\begin{equation}
\hat{\tau} \;=\; \mathcal{F}_\theta(V, Q;\, p_{\text{STVG}})
\;=\; \bigl\{\,(t, \hat{b}_t)\,\bigr\}_{t=\hat{t}_s}^{\hat{t}_e},
\end{equation} 
\ie, both the temporal interval $[\hat{t}_s, \hat{t}_e]$ and the bounding box $\hat{b}_t$ in every frame inside it.

\noindent \textbf{Spatial Video Grounding (SVG)} focuses on spatial localization within a temporally trimmed video, isolating the spatial component of grounding from temporal-boundary estimation. Given the trimmed video $V_{[t_s^{*}, t_e^{*}]} = \{I_t\}_{t=t_s^{*}}^{t_e^{*}}$, the query $Q$, and the system prompt $p_{\text{SVG}}$, the model predicts a per-frame bounding-box sequence as:
\begin{equation}
\bigl\{\hat{b}_t\bigr\}_{t=t_s^{*}}^{t_e^{*}}
\;=\;
\mathcal{F}_\theta\!\left(V_{[t_s^{*}, t_e^{*}]},\, Q;\, p_{\text{SVG}}\right).
\end{equation}

\noindent \textbf{Temporal Video Grounding (TVG)} aims to determine the correct temporal boundaries of the queried event, effectively performing temporal segment localization. Given $(V, Q)$ and the system prompt $p_{\text{TVG}}$, the model predicts only the temporal interval
\begin{equation}
[\hat{t}_s, \hat{t}_e] \;=\; \mathcal{F}_\theta(V, Q;\, p_{\text{TVG}}).
\end{equation}

\subsection{Adaptation Protocol}
\label{subsec:adaptationprotocol}

\paper\ provides per-domain training sets $\mathcal{T}_\mathcal{D}^{\text{train}} = \{(V_j, Q_j, \tau_j^{*})\}_{j=1}^{K}$ that enable evaluation of \emph{adaptation} alongside zero-shot generalization. Let $\mathcal{F}_\theta$ be a base VLM with parameters $\theta$, and let $\mathcal{A}$ denote any adaptation operator (\eg, PEFTs~\citep{hu2022lora,liu2024doraweightdecomposed,jia2022vpt}, or ICL~\citep{brown2020language,Kim2025VideoICL} that produces an adapted predictor $g$ from $\mathcal{F}_\theta$ and $\mathcal{T}_\mathcal{D}^{\text{train}}$,
\begin{equation}
g \;=\; \mathcal{A}\!\left(\mathcal{F}_\theta,\, \mathcal{T}_\mathcal{D}^{\text{train}}\right),
\qquad
g : (V, Q) \mapsto \hat{y},
\label{eq:adapt_op}
\end{equation}
where $\hat{y} \in \{\hat{\tau},\, \{\hat{b}_t\},\, [\hat{t}_s, \hat{t}_e]\}$ is the task-specific output selected by the system prompt $p$ as in \autoref{subsec:benchmarktasks}. The benchmark is agnostic to the choice of $\mathcal{A}$: any method that consumes $\mathcal{T}_\mathcal{D}^{\text{train}}$ to produce a predictor for $\mathcal{T}_\mathcal{D}^{\text{test}}$ (whether by augmenting the input with retrieved demonstrations, updating a subset of $\theta$, or full fine-tuning) is directly comparable under \paper's metrics.

\subsection{Domain and Data Source}

\noindent \textbf{Animal.}
To advance the understanding of non-human behaviors, we include Animal Kingdom~\citep{Ng2022AnimalKingdom}, capturing diverse species in natural environments. We complement this macroscopic view with a curated Mouse Scratching dataset, collected in a university medical department. This dataset focuses on fine-grained, high-frequency motions (\eg, distinguishing rapid scratches from continuous bouts), presenting challenges for precise spatio-temporal localization in clinical contexts.

\noindent \textbf{Industry.}
Industrial workflows demand precise, multi-step human-object interactions. We use two egocentric datasets, MECCANO~\citep{Francesco2023MECCANO} and ENIGMA-51~\citep{Ragusa2024Enigma51}, to capture professional activities like mechanical assembly and electrical repair using specialized tools.

\noindent \textbf{Sports.}
The high-velocity sports domain draws on MultiSports~\citep{Li2021MultiSports}, covering disciplines like basketball and gymnastics, where fine-grained coordination is central. To elevate complexity, we introduce a novel American Football dataset capturing highly tactical, domain-specific plays.

\noindent \textbf{Surgery.}
The surgical domain requires extreme precision and strict procedural adherence. We encompass two fundamental modalities: EgoSurgery~\citep{Fujii2022Surgicaltool,Fujii2024Egosurgeryphase,Fujii2024Egosurgerytool} for direct manual interactions in egocentric open surgery, and CholecTrack20~\citep{Nwoye2025CholecTrack20} for instrument-mediated interventions in laparoscopic procedures. This combination covers the full spectrum of modern surgical practice, from physical manipulation to technology-assisted techniques.

\noindent \textbf{Public Security.}
The public security domain focuses on critical incidents in public and transit environments. We utilize UCA~\citep{Yuan2024UCA} to provide an exocentric, multimodal perspective on urban surveillance, linking visual anomalies with semantic interpretations. This is complemented by DoTA~\citep{Yao2023DoTA}, which addresses highly dynamic traffic accidents from an egocentric driving perspective. Together, these benchmarks span the diverse challenges of security and risk monitoring, bridging fixed urban observation and mobile accident perception.

\subsection{Data Annotation}
\label{subsec:dataannotation}

\noindent \textbf{Annotators.} To ensure high-quality annotations, we assembled a diverse team tailored to the specific requirements of each dataset. Annotation and data processing for the existing datasets were conducted by graduate students specializing in computer vision. For domains requiring highly specialized knowledge, we engaged domain experts. Specifically, the Mouse Scratching dataset was annotated by \emph{two medical experts to guarantee the clinical fidelity of the captured behaviors}. The newly captured American Football dataset was annotated by \emph{individuals with years of active playing experience}, ensuring the accurate identification and localization of complex, domain-specific actions.

\begin{figure}[t]
  \centering
  \setlength{\tabcolsep}{0pt} 
  
  \begin{tabular}{ccc}
    \begin{minipage}{0.33\textwidth}
      \centering
      \includegraphics[width=0.7\linewidth]{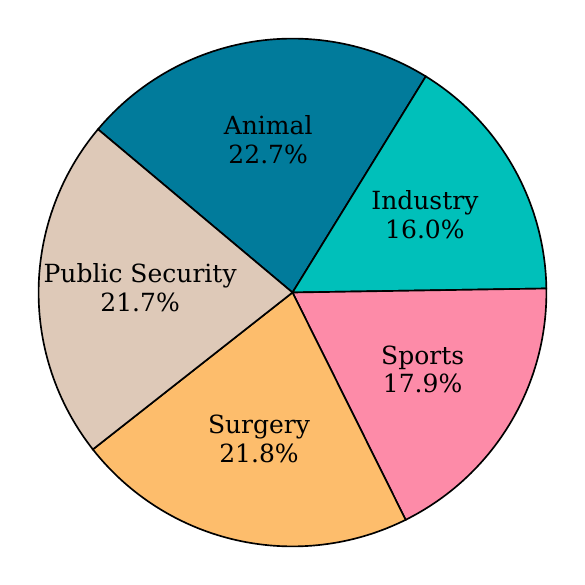} \\
      \vspace{-0.4em} {\small (a) Distribution of train set domains}
    \end{minipage} &
    \begin{minipage}{0.33\textwidth}
      \centering
      \includegraphics[width=0.7\linewidth]{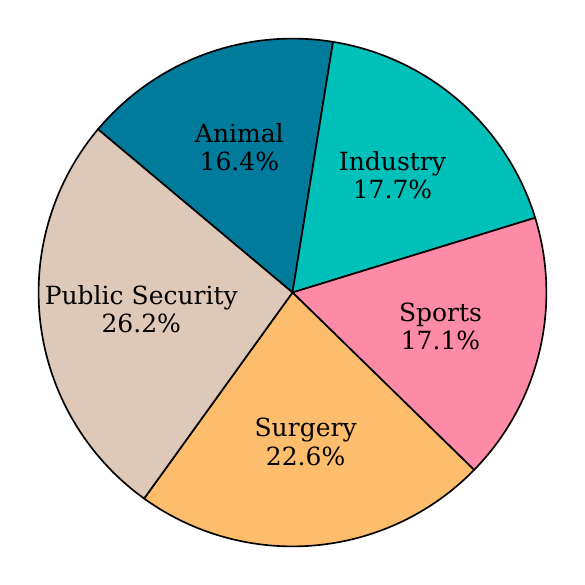} \\
      \vspace{-0.4em} {\small (b) Distribution of test set domains}
    \end{minipage} &
    \begin{minipage}{0.33\textwidth}
      \centering
      \includegraphics[width=0.98\linewidth]{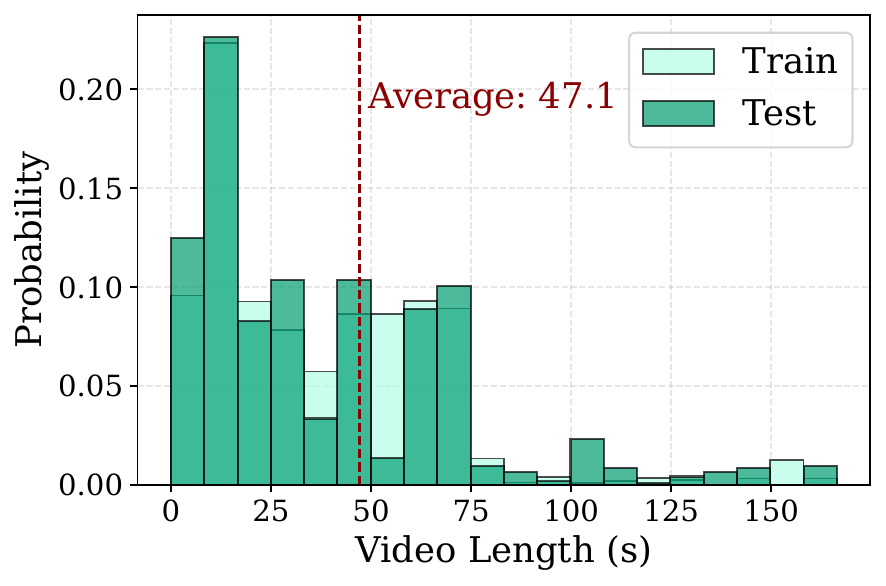} \\
      \vspace{-0.3em} {\small (c) Distribution of video length}
    \end{minipage} \\

    \\[-0.5em] 

    \begin{minipage}{0.33\textwidth}
      \centering
      \includegraphics[width=0.98\linewidth]{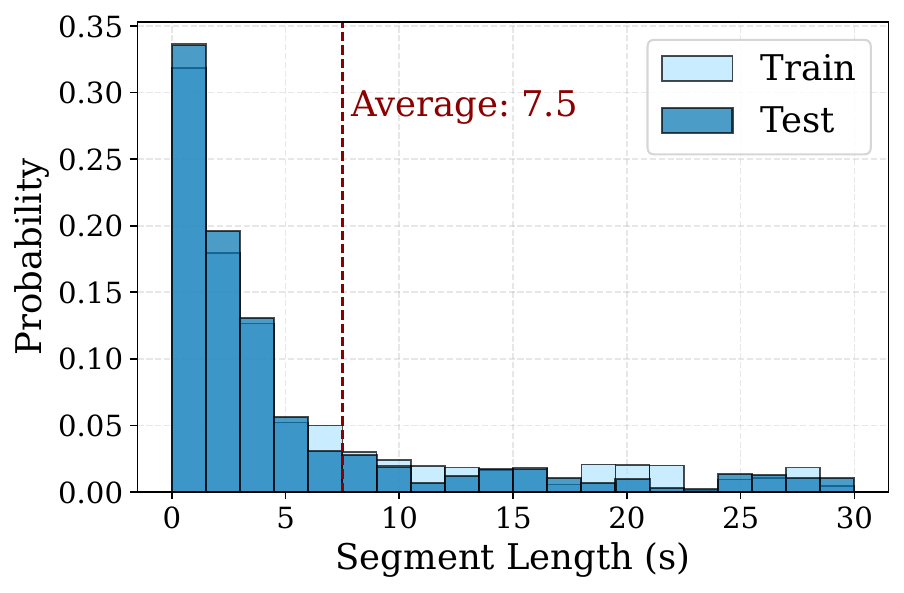} \\
      \vspace{-0.3em} {\small (d) Distribution of segment length}
    \end{minipage} &
    \begin{minipage}{0.33\textwidth}
      \centering
      \includegraphics[width=0.98\linewidth]{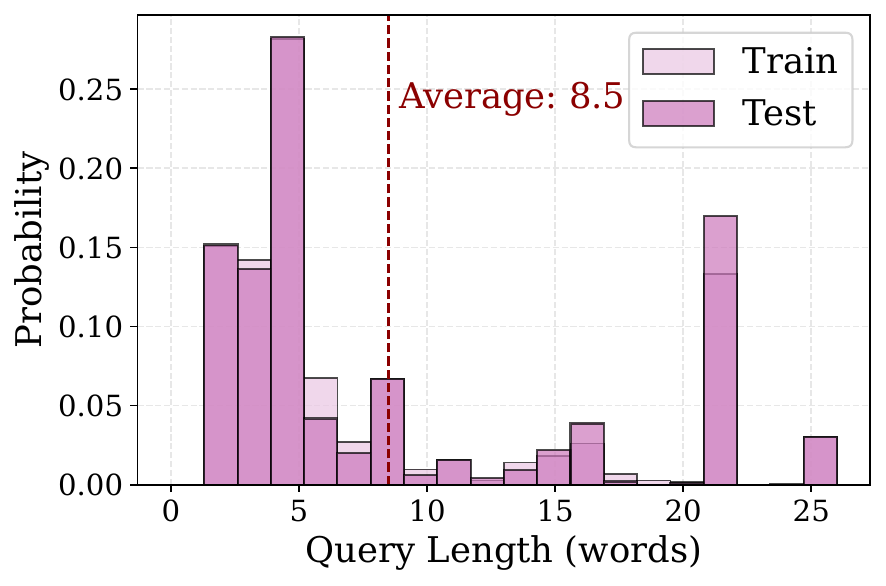} \\
      \vspace{-0.3em} {\small (e) Distribution of query length}
    \end{minipage} &
    \begin{minipage}{0.33\textwidth}
      \centering
      \includegraphics[width=0.98\linewidth]{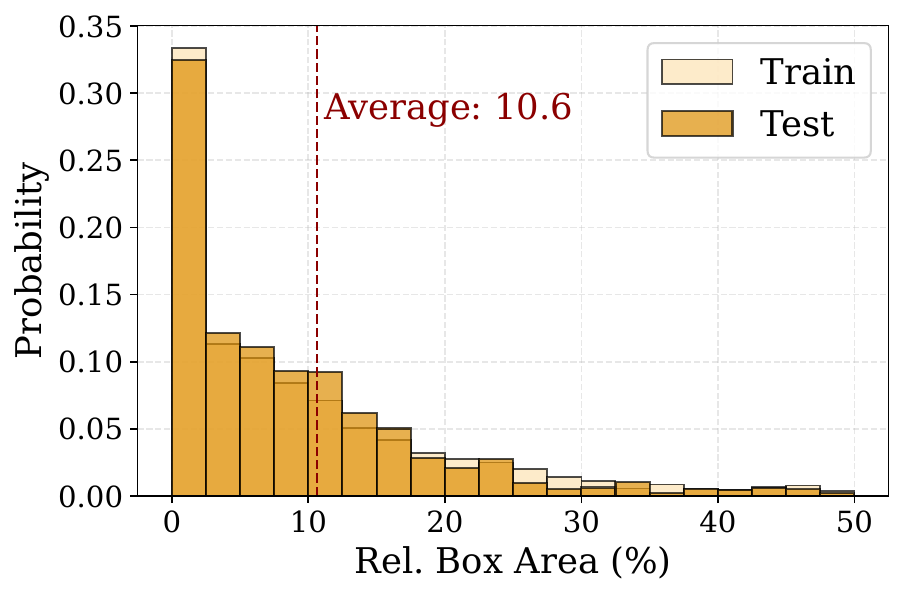} \\
      \vspace{-0.3em} {\small (f) Distribution of box area}
    \end{minipage}
  \end{tabular}
  \caption{\textbf{Representative statistics on \paper}, including distributions of training set domains in (a), test set domains in (b), video length (in seconds) in (c), temporal segment length (in seconds) in (d), textual query length (in words) in (e), and box area in (f).}
  \label{fig:statistics}
 \vspace{-1.5em}
\end{figure}

\noindent \textbf{Annotation Process and Quality Control.}
Our annotation pipeline utilizes advanced vision foundation models with rigorous human verification to ensure high-fidelity spatio-temporal grounding. The annotation process consists of the following key stages. (1) \textit{Textual Query and Corresponding Time Span}: For datasets equipped with existing temporal grounding annotations, we directly utilize their provided text queries. For datasets featuring other types of annotations, such as action category or anomaly localization, we manually paraphrase or annotate textual queries based on their respective labels. For datasets without grounding annotations, annotators provide the temporal span and corresponding textual query. (2) \textit{Spatial Bounding Boxes}: We take a detection-then-tracking approach to label the bounding boxes within the annotated time spans in a fully automated, semi-automated, and manual manner, based on the zero-shot grounding capability of the open-vocabulary object detection (Grounding DINO~\citep{Liu2024GroundingDINO}) and tracking (SAM2~\citep{ravi2025sam}) models for each data source. The fully automated approach applies Grounding DINO to the first annotated frame with the text query input and tracks the box until the annotated end frame. In the semi-automated approach, the human annotator manually labels the corresponding box to the first frame and applies tracking. In the manual approach, the human annotators manually label the bounding box per frame. For datasets that already possess dense spatial tracking annotations, we directly adapt their existing annotations to fit our unified format. (3) \textit{Manual Refinement}: All generated spatio-temporal boxes undergo a comprehensive manual inspection. Human annotators visually review the bounding boxes and manually correct any inaccurate bounding boxes, tracking drifts, or missing frames. (4) \textit{Quality Control}: \label{step:qc}As a final quality control measure, refined samples are independently double-checked by at least one additional annotator, ensuring that all corrections are accurate and guarantee high reliability, annotation consistency, and fidelity in the benchmark.  The details are in \appref{app:datasetdetails}.

\subsection{Benchmark Statistics}
We display representative statistics in \autoref{fig:statistics}. From \autoref{fig:statistics} (a) and (b), the 3,525 queries are evenly distributed across five specialized domains, providing a balanced testbed for domain adaptation. From \autoref{fig:statistics} (c) and (d), the average video length per query is $47.13$s while the target segment is $7.51$s ($16\%$ of the video length), requiring precise temporal localization within videos. Furthermore, \autoref{fig:statistics} (f) shows that the average relative box area is merely $10.63\%$, demonstrating the challenge of \paper{} in grounding small, specialized visual concepts.

\begin{table*}[tb]
\centering
\caption{\textbf{Zero-shot results on Spatio-Temporal (STVG), Temporal (TVG), and Spatial Video Grounding (SVG) tasks.}
Each cell reports STVG / TVG / SVG, where STVG uses $v\mathrm{IoU}@0.3$, TVG uses $t\mathrm{IoU}@0.3$, and SVG uses $s\mathrm{IoU}@0.3$.
\textbf{Bold} and \underline{underlined} fonts represent the best and second-best results, respectively.}

\label{tab:VLM_zero-shot_results}
\resizebox{\textwidth}{!}{
\begin{tabular}{l *{5}{c}}
\toprule
\textbf{Models} & \textcolor{AnimalColor}{\textbf{Animal}} & \textcolor{IndustryColor}{\textbf{Industry}} & \textcolor{SportsColor}{\textbf{Sports}} & \textcolor{SurgeryColor}{\textbf{Surgery}} & \textcolor{SecurityColor}{\textbf{Public Security}} \\
\midrule
\multicolumn{6}{c}{\textit{Proprietary VLMs}} \\
\midrule
GPT-4o           & 0.00 / 16.6 / 7.64 & 2.96 / 8.88 / 14.8 & 0.61 / 17.2 / 0.00 & 0.00 / 13.4 / 0.00 & 0.80 / 56.0 / 4.40 \\
GPT-5.1          & 3.18 / 19.7 / 55.4 & 5.33 / 9.47 / 28.4 & 0.00 / 25.8 / 1.84 & 1.39 / 22.7 / 0.67 & 4.80 / 61.2 / 35.6 \\
GPT-5.4          & 6.37 / 21.7 / 65.6 & 4.73 / 12.4 / 33.1 & 0.61 / 32.5 / 5.52 & 1.39 / \underline{38.9} / 2.68 & 2.00 / 68.0 / 53.6 \\
GPT-5.6-Sol      & 15.3 / 25.5 / \underline{92.4} & \textbf{10.7} / 24.3 / \underline{47.9} & \textbf{6.13} / \underline{44.2} / \underline{45.4} & 7.87 / 32.4 / 27.5 & 20.8 / 67.2 / \underline{88.4} \\
GPT-5.6-Terra    & 13.4 / 22.3 / 86.0 & 6.51 / 17.2 / 43.8 & \underline{4.91} / 31.9 / 31.3 & 4.17 / 32.4 / 10.1 & 18.4 / 56.8 / 83.2 \\
GPT-6-Astra      & \textbf{19.1} / 36.3 / \textbf{97.5} & 7.10 / 24.9 / \textbf{53.8} & \underline{4.91} / \textbf{46.0} / \textbf{58.3} & \textbf{25.9} / \textbf{47.7} / \textbf{61.7} & 18.8 / \textbf{76.0} / \textbf{93.6} \\
Gemini-2.5-Flash & 2.55 / 26.1 / 15.3 & 0.59 / 21.3 / 4.73 & 1.84 / 31.9 / 3.68 & 0.00 / 22.7 / 0.67 & 0.40 / 51.2 / 2.00 \\
Gemini-2.5-Pro   & 8.28 / 36.9 / 20.4 & 1.18 / \textbf{39.6} / 17.2 & 0.61 / 37.4 / 8.59 & 1.39 / 31.5 / 3.36 & 4.00 / 65.9 / 22.8 \\
Gemini-3-Flash   & 14.6 / 36.3 / 52.2 & 5.92 / \underline{30.8} / 20.1 & 2.45 / 29.4 / 2.45 & 0.93 / 38.0 / 13.4 & 11.2 / 66.8 / 45.6 \\
Gemini-3.1-Pro   & 16.6 / 37.6 / 70.7 & \underline{7.69} / 21.9 / 41.4 & 1.23 / 26.4 / 16.6 & 4.17 / 32.9 / 34.2 & \underline{22.8} / 69.5 / 52.0 \\
Gemini-3.5-Flash & 16.6 / \textbf{41.4} / 66.9 & 5.92 / 19.5 / 47.3 & \underline{4.91} / 30.1 / 15.3 & 4.17 / 30.1 / 29.5 & 22.4 / 61.6 / 52.0 \\
Gemini-3.8-Flash & \underline{17.2} / \underline{38.9} / 66.2 & \underline{7.69} / 20.1 / 37.3 & 4.29 / 36.2 / 12.9 & \underline{9.26} / 35.6 / \underline{41.6} & \textbf{28.4} / \underline{75.2} / 56.0 \\
\midrule
\multicolumn{6}{c}{\textit{Open-source Specialized VLMs}} \\
\midrule
LLaVA-ST         & 12.1 / 19.7 / 53.5 & 0.00 / 9.47 / 12.4 & 0.79 / 8.59 / 3.68 & 0.00 / 12.0 / 0.00 & 0.80 / 35.6 / 13.2 \\
\midrule
\multicolumn{6}{c}{\textit{Open-source General-Purpose VLMs}} \\
\midrule
Qwen3-VL-4B      & 5.73 / 25.5 / 42.0 & 0.00 / 4.14 / 13.0 & 0.00 / 8.59 / 0.00 & 0.00 / 13.4 / 0.00 & 0.40 / 39.2 / 0.00 \\
Qwen3-VL-8B      & 3.82 / 19.7 / 17.8 & 0.00 / 10.7 / 7.69 & 0.00 / 11.7 / 0.00 & 0.00 / 15.7 / 0.00 & 0.80 / 46.0 / 3.60 \\
Qwen3.5-4B       & 2.55 / 30.6 / 13.4 & 0.00 / 14.8 / 7.69 & 0.00 / 12.9 / 0.00 & 0.00 / 28.2 / 1.34 & 0.40 / 49.2 / 2.00 \\
Qwen3.5-9B       & 4.46 / 35.0 / 20.4 & 0.59 / 17.2 / 12.4 & 0.61 / 14.7 / 0.00 & 0.00 / 28.2 / 1.34 & 0.40 / 50.4 / 4.80 \\
Qwen3.8-27B      & 12.7 / 26.1 / 12.7 & 5.33 / 18.3 / 28.4 & 0.00 / 11.0 / 1.23 & 0.00 / 35.2 / 10.1 & 13.6 / 56.4 / 2.80 \\
Eagle2.5-8B      & 0.00 / 24.8 / 1.27 & 0.00 / 7.69 / 2.37 & 0.00 / 7.98 / 0.00 & 0.00 / 15.3 / 0.00 & 0.00 / 47.6 / 0.40 \\
InternVL3-8B     & 0.64 / 15.3 / 3.82 & 0.00 / 5.92 / 1.18 & 0.00 / 6.79 / 0.00 & 0.00 / 10.6 / 0.00 & 0.00 / 4.40 / 0.80 \\
InternVL3-14B    & 0.64 / 17.8 / 7.64 & 0.00 / 5.33 / 1.18 & 0.00 / 7.36 / 0.00 & 0.00 / 10.6 / 0.00 & 0.00 / 13.2 / 0.00 \\
InternVL3.5-8B   & 0.00 / 11.5 / 0.00 & 0.00 / 2.96 / 1.18 & 0.00 / 6.13 / 0.00 & 0.00 / 7.41 / 0.00 & 0.00 / 3.60 / 0.00 \\
Gemma4-12B       & 5.10 / 27.4 / 3.82 & 4.14 / 11.2 / 18.3 & 0.00 / 9.20 / 0.00 & 2.31 / 25.0 / 2.01 & 1.60 / 43.2 / 0.00 \\
\bottomrule
\end{tabular}\vspace{-3.0em}}
\end{table*}

\section{Experiments}
\subsection{Experimental setup}
\label{sec:exp_setup}
\noindent \textbf{Models.} To ensure a comprehensive evaluation, we benchmark a diverse set of VLMs, including \textit{Proprietary models}: the GPT series (GPT-4o~\citep{achiam2023gpt}, GPT-5.1~\citep{singh2025gpt5}, GPT-5.4~\citep{openai2026gpt5_4}, GPT-5.6-Sol/Terra~\citep{openai2026gpt5_6}, and GPT-6-Astra~\citep{openai2026gpt6astra}) and the Gemini series (Gemini-2.5-Flash/Pro~\citep{comanici2025gemini2_5}, Gemini-3-Flash~\citep{google2025gemini3flash}, Gemini-3.1-Pro~\citep{google2026gemini3_1pro}, Gemini-3.5-Flash~\citep{google2026gemini3_5flash}, and Gemini-3.8-Flash~\citep{google2026gemini3_8flash}), \textit{Open-source Specialized VLMs}: LLaVA-ST~\citep{Li2025LLaVA-ST}, and \textit{Open-source General-Purpose VLMs}: the Qwen series (Qwen3-VL-4B/8B~\citep{Qwen3-VL}, Qwen3.5-4B/9B~\citep{qwen3.5}, and Qwen3.8-27B~\citep{qwen38}), Eagle2.5-8B~\citep{chen2025eagle2_5}, the InternVL series (InternVL3-8B/14B and InternVL3.5-8B~\citep{wang2025internvl3_5}), and Gemma4-12B~\citep{gemmateam2026gemma4}. Detailed inference configurations and full prompt templates are provided in \appref{app:implementation} and \appref{app:prompts}, respectively.

\begin{table*}[tb]
\centering
\caption{\textbf{Adaptation results on Spatio-Temporal (STVG), Temporal (TVG), and Spatial Video Grounding (SVG) tasks.}
Each cell reports STVG / TVG / SVG, where STVG uses $v\mathrm{IoU}@0.3$, TVG uses $t\mathrm{IoU}@0.3$, and SVG uses $s\mathrm{IoU}@0.3$.
For each model, the first row shows the zero-shot baseline, followed by adaptation using the provided per-domain training subsets. \colorbox{gray!12}{Shaded} rows denote fine-tuning-based adaptation, while unshaded adaptation rows are training-free.
\poscol{Purple} denotes performance improvements or ties and \negcol{red} denotes degradations relative to the zero-shot baseline.
}
\label{tab:VLM_adaptation_results}
\resizebox{\textwidth}{!}{
\begin{tabular}{l *{5}{c}}
\toprule
\textbf{Models} & \textcolor{AnimalColor}{\textbf{Animal}} & \textcolor{IndustryColor}{\textbf{Industry}} & \textcolor{SportsColor}{\textbf{Sports}} & \textcolor{SurgeryColor}{\textbf{Surgery}} & \textcolor{SecurityColor}{\textbf{Public Security}} \\
\midrule
\multicolumn{6}{c}{\textit{Proprietary VLM (Gemini-3.1-Pro)}} \\
\midrule
Zero-shot        & 16.6 / 37.6 / 70.7 & 7.69 / 21.9 / 41.4 & 1.23 / 26.4 / 16.6 & 4.17 / 32.9 / 34.2 & 22.8 / 69.5 / 52.0 \\
~+ICL            & \negcol{12.7} / \negcol{33.1} / \negcol{60.5} & \poscol{11.8} / \poscol{27.8} / \negcol{39.6} & \poscol{1.23} / \poscol{39.9} / \negcol{7.36} & \poscol{9.72} / \poscol{42.1} / \poscol{34.2} & \negcol{22.4} / \poscol{77.2} / \negcol{41.6} \\
~+GEPA           & \poscol{17.2} / \negcol{34.4} / \negcol{60.5} & \poscol{13.0} / \poscol{38.5} / \poscol{51.5} & \poscol{2.45} / \poscol{39.3} / \negcol{12.9} & \poscol{21.8} / \poscol{39.8} / \poscol{45.6} & \negcol{19.2} / \poscol{78.4} / \poscol{53.2} \\
\midrule
\multicolumn{6}{c}{\textit{Open-source VLM (Qwen3-VL-8B)}} \\
\midrule
Zero-shot        & 3.82 / 19.7 / 17.8 & 0.00 / 10.7 / 7.69 & 0.00 / 11.7 / 0.00 & 0.00 / 15.7 / 0.00 & 0.80 / 46.0 / 3.60 \\
~+ICL            & \poscol{3.82} / \poscol{28.0} / \poscol{20.4} & \poscol{3.55} / \poscol{14.8} / \poscol{15.4} & \poscol{0.00} / \poscol{23.3} / \poscol{0.61} & \poscol{0.93} / \poscol{34.3} / \poscol{13.4} & \poscol{4.00} / \poscol{65.6} / \negcol{3.20} \\
~+GEPA           & \negcol{3.18} / \negcol{17.2} / \poscol{35.7} & \poscol{0.00} / \poscol{11.8} / \poscol{11.8} & \poscol{0.00} / \poscol{17.8} / \poscol{0.61} & \poscol{0.00} / \poscol{39.4} / \poscol{0.00} & \negcol{0.40} / \poscol{48.8} / \poscol{4.80} \\
\rowcolor{gray!12}
~+SFT LoRA       & \poscol{8.92} / \poscol{21.0} / \poscol{26.1} & \poscol{0.59} / \negcol{7.69} / \poscol{21.3} & \poscol{0.00} / \poscol{19.0} / \poscol{1.23} & \poscol{0.46} / \poscol{36.1} / \poscol{12.8} & \poscol{4.00} / \poscol{64.8} / \negcol{0.40} \\
\rowcolor{gray!12}
~+SFT Full Fine-tuning & \poscol{10.2} / \negcol{16.6} / \poscol{28.7} & \poscol{2.96} / \negcol{7.69} / \poscol{24.3} & \poscol{0.61} / \poscol{12.9} / \poscol{5.52} & \poscol{4.63} / \poscol{27.8} / \poscol{32.9} & \poscol{8.00} / \poscol{71.2} / \poscol{16.0} \\
\rowcolor{gray!12}
~+GRPO LoRA        & \poscol{11.5} / \poscol{34.4} / \poscol{31.2} & \poscol{0.59} / \poscol{18.3} / \poscol{25.4} & \poscol{0.61} / \poscol{33.1} / \poscol{3.07} & \poscol{3.70} / \poscol{54.2} / \poscol{36.2} & \poscol{7.20} / \poscol{83.6} / \negcol{2.40} \\
\bottomrule
\end{tabular}\vspace{-3.0em}}
\end{table*}

\noindent \textbf{Adaptation Methods.}
We investigate five adaptation strategies spanning training-free adaptation
(ICL, as detailed in \appref{app:icl_setup}, and
GEPA~\citep{agrawal2025gepa}) and fine-tuning-based adaptation
(SFT via LoRA~\citep{hu2022lora} and full
fine-tuning, and GRPO~\citep{shao2024deepseekmath} with LoRA using IoU-based
rewards). ICL and GEPA are prompt-level and
applicable to all models, so we evaluate them on both Gemini-3.1-Pro and
Qwen3-VL-8B, while SFT, and GRPO require parameter access and are
therefore evaluated only on Qwen3-VL-8B.

\noindent \textbf{Evaluation Metrics.} In accordance with established protocols~\citep{Huang2024VTimeLLM,Li2025LLaVA-ST,Ren2024TimeChat,Yang2022TubeDETR}, we report $v\mathrm{IoU}@0.3$ for STVG, $t\mathrm{IoU}@0.3$ for TVG, and $s\mathrm{IoU}@0.3$ for SVG.
Results for additional threshold-based metrics, including $v\mathrm{IoU}$, $s\mathrm{IoU}$, and $t\mathrm{IoU}$ at multiple thresholds, are detailed in~\appref{app:detailed_results}.

\subsection{Zero-Shot Results}
\label{sec:zero-shot_results}
As shown in \autoref{tab:VLM_zero-shot_results}, current VLMs remain far from practical STVG performance in specialized domains.
Even the best model in each domain achieves at most 28.4 $v\mathrm{IoU}@0.3$ (public security, Gemini-3.8-Flash), and only 6.13 in sports.
Open-source VLMs lag far behind: all except Gemma4-12B and Qwen3.8-27B score below 1.0 STVG outside animal, and scaling helps only partially, as the largest, Qwen3.8-27B, improves in animal, industry, and public security but still scores 0.00 in sports and surgery.
These results highlight the need to improve the STVG capability of VLMs, particularly open-source ones. Decomposing STVG reveals where grounding breaks down.
For open-source and earlier proprietary models, SVG lags far behind TVG in most domains (\eg, 56.4 TVG vs.\ 2.80 SVG for Qwen3.8-27B in public security), making spatial localization the primary bottleneck.
In contrast, the strongest proprietary models perform well on both sub-tasks, \eg, GPT-6-Astra reaches 76.0 TVG and 93.6 SVG in public security, yet only 18.8 STVG.
This indicates that for these models the bottleneck shifts to jointly aligning spatial and temporal predictions.
\vspace{-0.5em}
\subsection{Adaptation Results}
\label{sec:adaptation_results}
As shown in \autoref{tab:VLM_adaptation_results}, training-free adaptation yields inconsistent effects across models and domains.
For Gemini-3.1-Pro, ICL and GEPA improve TVG in four of five domains but often degrade SVG, and GEPA substantially improves STVG in surgery (4.17$\rightarrow$21.8) while degrading it in public security.
For Qwen3-VL-8B, ICL improves nearly all metrics, whereas GEPA yields mixed results, and STVG gains remain at most 4.00. Fine-tuning is more effective, particularly GRPO LoRA, whose TVG in surgery and public security surpasses all zero-shot models including GPT-6-Astra.
The gains, however, vary greatly across domains: GRPO LoRA raises SVG from 0.00 to 36.2 in surgery but not in public security (3.60$\rightarrow$2.40).
Moreover, these sub-task gains translate into only small STVG improvements, \eg, only 3.70 STVG in surgery despite the above TVG and SVG.
These results call for adaptation methods that robustly improve joint spatio-temporal grounding across domains.
\begin{figure*}[tb]
\centering
\includegraphics[width=0.95\linewidth]{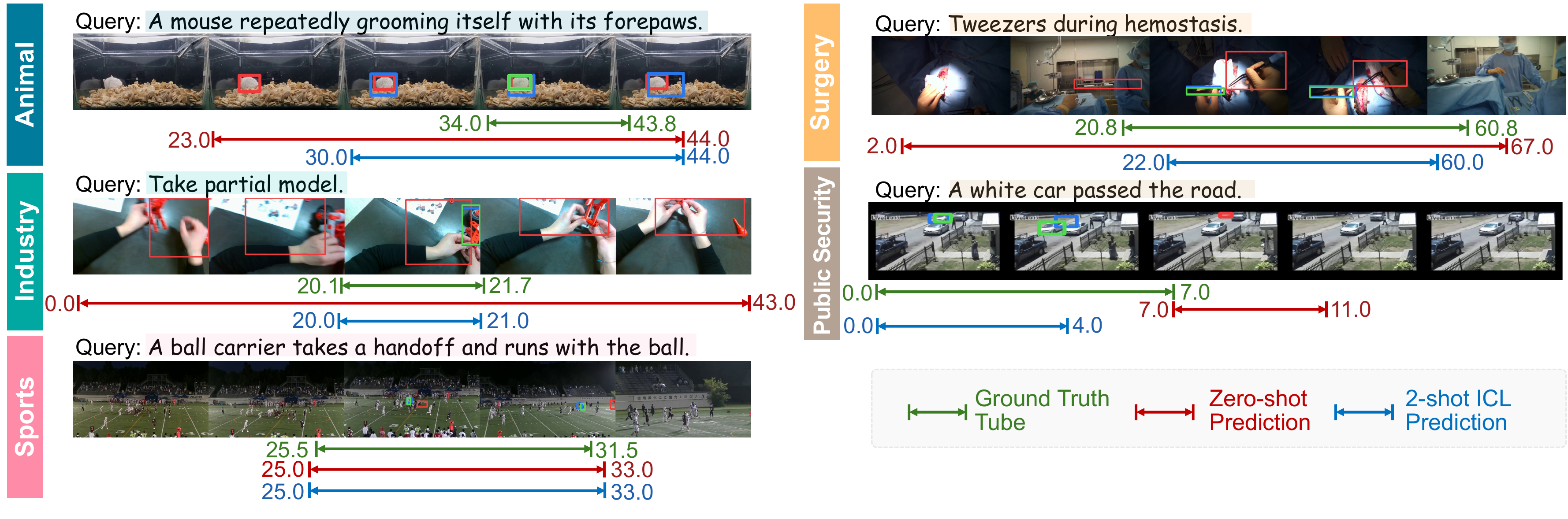}
\caption{\textbf{Qualitative STVG results of Gemini-3.1-Pro across five specialized domains on \paper.}
Each example compares the \textcolor{gtshotColor}{ground-truth tube}, \textcolor{twoshotColor}{2-shot ICL prediction}, and \textcolor{zeroshotColor}{zero-shot prediction} for the same query.
The temporal boundaries are shown in seconds.
2-shot ICL can improve localization on some samples, but the gains are inconsistent, and spatial grounding remains fragile in specialized domains.}
\label{fig:qualitative_stvg_gemini31pro}
\end{figure*}

\subsection{Additional Analysis}
\begin{figure*}[tb]
\centering
\includegraphics[width=\textwidth]{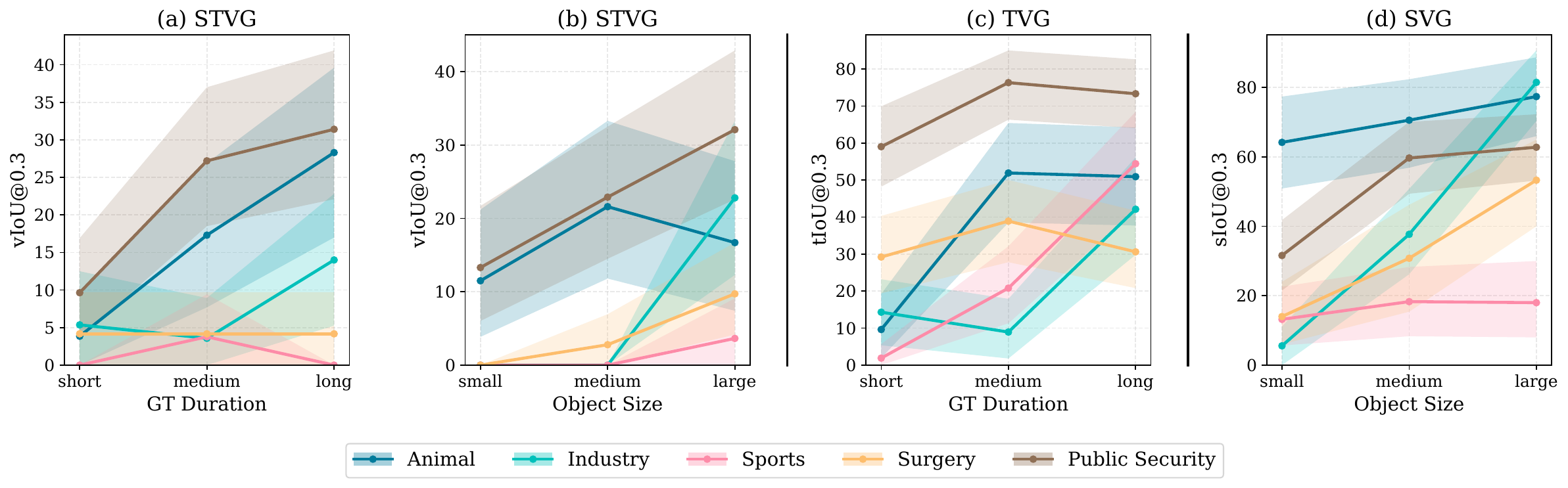}
\caption{\textbf{Sensitivity analysis of temporal and spatial scales.} 
Within each domain, temporal events are split into terciles by ground-truth duration (short/medium/long), and targets are split into terciles by relative box area (small/medium/large). Error bars denote bootstrap $95\%$ confidence intervals computed per domain. All results use Gemini-3.1-Pro. \vspace{-1.0em}
}
\label{fig:integrated_sensitivity}
\end{figure*}

\noindent \textbf{Sensitivity to Temporal and Spatial Scales.}
We further analyze sensitivity to two fundamental sources of difficulty in STVG: ground-truth event duration and target bounding box size, as shown in \autoref{fig:integrated_sensitivity}.
For this analysis, we group sampled evaluation examples into comparable temporal-duration and object-size bins.
Longer events tend to be easier to ground: both TVG and STVG generally improve as the ground-truth duration increases ((a), (c)).
Likewise, larger targets tend to be easier to ground: both SVG and STVG generally improve as the target box size increases ((b), (d)).
These trends, however, do not always carry over to STVG, since it requires both the temporal interval and the spatial boxes to be correct.
In Sports, for instance, temporal localization improves with event duration, but spatial localization remains poor regardless of duration, so that $v\mathrm{IoU}@0.3$ stays near zero and the duration trend is not reflected in the metric. Additional analyses are in \appref{app:further_analysis}.

\section{Conclusion}
This paper presented \textbf{\paper}, a domain-specialized benchmark for STVG in VLMs.
By spanning five specialized domains and providing both training and evaluation subsets per domain, \paper\ enables evaluating the zero-shot grounding and adaptation capability of VLMs.
Experiments with 23 VLMs show that specialized-domain STVG remains far from practical performance.
Spatial localization is the main bottleneck for most models, whereas the strongest ones struggle to align spatial and temporal predictions jointly.
Neither training-free nor fine-tuning-based adaptation robustly improves STVG across domains.

\noindent \textbf{Broader Impacts.}\label{sec:broader_impacts}
\paper\ provides a standardized benchmark for STVG on five specialized domains, aiming to evaluate how VLMs adapt to specialized domains.
By releasing \paper\ with expert-annotated data in a unified format, we expect researchers to develop robust domain-adaptation techniques for video grounding in VLMs under limited training data.

\subsection*{AI use statement}
In this work, we used generative AI tools to aid and polish our writing,
including improving phrasing, grammar, and clarity of the manuscript text.
We have not used generative AI tools to generate synthetic data sets, help
develop theoretical models or conceptual frameworks, formulate mathematical
claims, provide critical ingredients for proving mathematical claims, assist
in the writing of proofs, propose or refine hypotheses, design or provide
feedback on research methodology or experiments, implement methods, assist
with translation, clean and reformat our dataset, support qualitative and
thematic data analysis, or interpret results; these are not applicable to
this work. We have reviewed all AI-assisted work: three of the authors
independently reviewed and edited the AI-polished text to ensure accuracy,
clarity, and faithfulness to the original content and claims. We take
responsibility for the final content of this work, including text, claims
or artifacts produced with the aid of generative AI.

\subsection*{Ethics statement}

Eight of the ten components of \paper\ are existing public datasets, for which
we release derived annotations only; we do not redistribute their videos, and
users must obtain each source from its official provider under the original
terms (\appref{app:licenses}). These sources differ in their handling of
personal data: some are de-identified by their providers (\eg, EgoSurgery),
while others consist of unmodified surveillance, dashcam, or broadcast
footage. We rely on the de-identification, consent, and access arrangements
established by each provider and do not alter them. The surgical sources
(EgoSurgery, CholecTrack20) additionally constitute medical data. We use all
of these as secondary data under each provider's license and data-use
agreement, which we confirmed permits third-party re-annotation and the
release of derived annotations. We did not attempt to re-identify or link any
individual, and collected no demographic or clinical attributes.

Two datasets were collected by us. The Mouse Scratching recordings were
obtained observationally from animals already enrolled in an ongoing study at
the authors' university school of medicine, approved by its institutional
animal experimentation committee and conducted under the
institution's animal experimentation regulations. No additional procedure or
animal was introduced for this benchmark. The American Football recordings
were filmed at public university-level games, with approval subsequently
obtained from the host university. Both are released under CC BY-NC-SA 4.0.

One aspect of our own processing warrants note. In the public-security
subsets we add bounding boxes to people appearing in surveillance and dashcam
footage; a box marks the primary object referred to by the query and does not
assert that any depicted individual committed an offence. These sources
contain violent incidents and real collisions, which we retain as they
constitute the events being evaluated.

\paper\ supports research on spatio-temporal grounding and domain adaptation;
it is not evidence of clinical, operational, or forensic validity. Accuracy
here does not indicate readiness for intraoperative use, deployed
surveillance, judgements about specific individuals, or validated behavioural
scoring.

\subsection*{Reproducibility statement}
To facilitate reproducibility, we release our code via an
\href{https://github.com/anonymous7bhblm-cpu/7bnblm}{Anonymous GitHub}
repository and our benchmark data via an anonymized
\href{https://huggingface.co/datasets/anonymous7bhblm/7bnblm}{Hugging Face}
dataset repository. The code includes our data processing pipeline,
implementations of the five adaptation strategies evaluated in this work, and
the evaluation scripts used to produce the reported results. Machine-readable dataset metadata in the Croissant format is available from
the Hugging Face dataset repository. Details of the
model configurations, adaptation strategies, and evaluation protocols are
provided in the main text and appendix.



\bibliography{iclr2027_conference}
\bibliographystyle{iclr2027_conference}

\newpage
\appendix
\section*{Appendix}
\input{appendix.tex}

\end{document}

%% file: math_commands.tex
\usepackage{amsmath,amsfonts,bm}

\def\eqref#1{equation~\ref{#1}}

\def\1{\bm{1}}

\DeclareMathAlphabet{\mathsfit}{\encodingdefault}{\sfdefault}{m}{sl}
\SetMathAlphabet{\mathsfit}{bold}{\encodingdefault}{\sfdefault}{bx}{n}



%% file: appendix.tex
\section{Implementation Details}
\label{app:implementation}

\subsection{Inference Configuration}
\label{app:inference_config}
\noindent \textbf{Model-Specific Parameters.} Unless otherwise noted, we employed the default inference configurations for each model or API. Specific exceptions were made for the following models: for Qwen3.5~\cite{qwen3.5}, we disabled ``thinking mode'' because preliminary trials with five retries on a dataset showed that most output tokens were spent on free-form reasoning text, preventing the models from reliably following the required timestamp and bounding-box output format; for Qwen3.8~\cite{qwen38}, served with vLLM, we also disabled thinking mode and used the sampling parameters recommended in its model card for non-thinking mode (a temperature of $0.7$, \texttt{top\_p=0.8}, \texttt{top\_k=20}, and a presence penalty of $1.5$); for Eagle 2.5~\cite{chen2025eagle2_5}, five preliminary experiments with the default setting failed to follow the required output format, and therefore we used \texttt{do\_sample=True}, \texttt{top\_p=0.95}, a temperature of $0.8$, and a maximum of 1024 new tokens to allow natural generation while improving instruction following; and for LLaVA-ST~\cite{Li2025LLaVA-ST}, we used \texttt{do\_sample=True}, a temperature of $0.01$, and 1 beam. For all other models, output-length limits were set to 64 tokens for temporal grounding and 4096 tokens for spatial and spatio-temporal grounding.

\noindent \textbf{Video Preprocessing.} The default preprocessing involved an explicit sampling rate of 1~fps with a maximum of 120 frames, which was applied to Qwen3-VL~\cite{Qwen3-VL}, Qwen3.5, Qwen3.8, Eagle 2.5, InternVL-3~\cite{zhu2025internvl3}, InternVL-3.5~\cite{wang2025internvl3_5} and Gemma 4~\cite{gemmateam2026gemma4}; for Qwen3.8, frame sampling with this setting was performed by the vLLM server. For Gemini models~\cite{comanici2025gemini2_5,google2025gemini3flash,google2026gemini3_1pro,google2026gemini3_5flash,google2026gemini3_8flash}, we passed each mp4 videos directly to the API without any additional preprocessing. For GPT models~\cite{achiam2023gpt,singh2025gpt5,openai2026gpt5_4,openai2026gpt5_6,openai2026gpt6astra}, videos were sampled at 1~fps with a maximum of 120 frames; for GPT-5.4, GPT-5.6-Sol/Terra, and GPT-6-Astra, the maximum was reduced to 32 frames because of their high API cost. All sampled frames were resized to a maximum resolution of 512 pixels on the longer side. For LLaVA-ST, query videos were processed with a fixed budget of 100 sampled frames. These preprocessing configurations were consistently applied to both query and in-context videos. For InternVL-3/3.5, the sampled frame counts of demonstration videos were dynamically adjusted to match those of the query video during the in-context learning process.

\subsection{Computational Requirements}
\label{app:compute}
All local open-source model experiments were conducted on two internal GPU servers.
One internal GPU server was equipped with NVIDIA RTX 5090 (32GB) and NVIDIA RTX PRO 5000 Blackwell (48GB) GPUs.
Another internal GPU server was equipped with NVIDIA A100 (80GB) GPUs. API-based experiments with GPT and Gemini models were conducted through their official APIs.

\subsection{In-Context Learning Setup}
\label{app:icl_setup}
To assess the model's adaptability, we employ ICL as a simple baseline. Despite its simplicity, it can be universally applied to both proprietary (closed-source) and open-source models, unlike PEFTs~\cite{hu2022lora,liu2024doraweightdecomposed,jia2022vpt}.

For our few-shot evaluations, we provide $m=2$ demonstrations retrieved from the domain-specific training split. Following \cite{Kim2025VideoICL}, we employ similarity-based selection, computing a hybrid retrieval score $S$ defined as the weighted sum of visual and textual cosine similarities:
\begin{equation}
    S = (1 - \alpha)s_{\mathrm{visual}} + \alpha s_{\mathrm{text}}
\end{equation}
where $s_{\mathrm{visual}}$ and $s_{\mathrm{text}}$ denote the cosine similarities computed using InternVideo2~\cite{wang2024internvideo2} and SentenceBERT~\cite{reimers-gurevych-2019-sentence} embeddings, respectively. We set the weighting coefficient $\alpha = 0.5$ across all experiments to balance the influence of both modalities, and select the top $m=2$ examples with the highest score $S$ for each query.

\subsection{GEPA Setup}
\label{app:gepa_setup}
We use GEPA~\cite{agrawal2025gepa} to optimize the base prompt. 
GEPA uses two models: a task model, which performs grounding with the current prompt, and a reflection model, which reads the task model's outputs and proposes a revised prompt. In our experiments, each model serves as both its own task model and its own reflection model (Gemini-3.1-Pro and Qwen3-VL-8B). 
The scoring metric is $t\mathrm{IoU}$ for TVG, $s\mathrm{IoU}$ for SVG, and $v\mathrm{IoU}$ for STVG. 
All other hyperparameters follow the official GEPA repository\footnote{\url{https://github.com/gepa-ai/gepa}}.

\subsection{Fine-Tuning SetUp}
\label{app:finetuning_setup}
\noindent \textbf{Supervised Fine-Tuning.} We fine-tune Qwen3-VL-8B with LoRA~\cite{hu2022lora} and with full-parameter updates, computing the cross-entropy loss only on the answer tokens. Both variants are trained for 5 epochs with a learning rate of $2\times10^{-5}$ and a batch size of 8. For LoRA, we use rank 16, $\alpha=32$, and dropout 0.05 on the attention and MLP projection layers.

\noindent \textbf{Reinforcement Learning Fine-Tuning.} We further apply GRPO~\cite{shao2024deepseekmath} with LoRA to the SFT models. For each query, we sample a group of four responses. The reward is the sum of an accuracy reward and a format reward,
\begin{equation}
r = r_{\mathrm{acc}} + \lambda\, r_{\mathrm{fmt}}.
\end{equation}
The accuracy reward $r_{\mathrm{acc}}$ is the task-specific IoU between a generated response $\hat{y}$ and the ground truth $y$: $t\mathrm{IoU}(\hat{y}, y)$ for TVG, $s\mathrm{IoU}(\hat{y}, y)$ for SVG, and $v\mathrm{IoU}(\hat{y}, y)$ for STVG. The format reward $r_{\mathrm{fmt}}$ is $1$ for a parsable response and $0$ otherwise, and $\lambda = 0.1$ weights it.

\section{Prompts}
\label{app:prompts}
To ensure reproducibility and foster future spatio-temporal grounding research, we provide the prompts used to evaluate open-source models (Qwen3-VL~\cite{Qwen3-VL}, Qwen3.5~\cite{qwen3.5}, Qwen3.8~\cite{qwen38}, InternVL-3~\cite{zhu2025internvl3}, InternVL-3.5~\cite{wang2025internvl3_5}, Eagle 2.5~\cite{chen2025eagle2_5}, and Gemma 4~\cite{gemmateam2026gemma4}), proprietary models (Gemini-2.5-Flash/Pro~\cite{comanici2025gemini2_5}, Gemini-3-Flash~\cite{google2025gemini3flash}, Gemini-3.1-Pro, GPT-4o~\cite{achiam2023gpt} and GPT-5.1~\cite{singh2025gpt5}), and the specialist LLaVA-ST model~\cite{Li2025LLaVA-ST}.

The requested response format differs across model families in temporal identifiers, bounding-box coordinate order, and coordinate scale.
For Gemini and GPT models, we followed their official API documentation.\footnote{Gemini: \url{https://ai.google.dev/gemini-api/docs/image-understanding} and \url{https://ai.google.dev/gemini-api/docs/video-understanding}; OpenAI: \url{https://platform.openai.com/docs/guides/vision}.}
For Qwen3-VL, InternVL-3, InternVL-3.5, and LLaVA-ST, we followed their technical reports, documentation, or released code~\cite{Qwen3-VL,zhu2025internvl3,wang2025internvl3_5,Li2025LLaVA-ST}.\footnote{InternVL documentation: \url{https://internvl.readthedocs.io/en/latest/get_started/chat_data_format.html}; LLaVA-ST code: \url{https://github.com/appletea233/LLaVA-ST}.}
For Gemma 4, we used the same prompt templates and response format as the Gemini family.
For Qwen3.5, Qwen3.8, and Eagle 2.5, we did not find a model-specific grounding-output schema in the released paper or code~\cite{chen2025eagle2_5}, so we used the same open-source prompt format selected by preliminary parsing trials; see \autoref{app:inference_config}.
\autoref{tab:output_conventions} summarizes the detailed response format used for each model family; all outputs are subsequently parsed and converted into a unified internal representation before evaluation.

\begin{table}[t]
\centering
\small
\caption{Model-specific response formats requested during inference.
The raw model outputs are parsed and normalized during evaluation, but the requested surface format differs by model family along three axes: temporal identifier, box order, and coordinate scale.}
\label{tab:output_conventions}
\resizebox{\linewidth}{!}{%
\begin{tabular}{llll}
\toprule
\textbf{Model family} & \textbf{Temporal identifier} & \textbf{Box order} & \textbf{Coordinate scale} \\
\midrule
Gemini and Gemma & Timestamp (MM:SS) & [yxyx] & 0--1000 normalized \\
GPT & Frame index & [xyxy] & 0--1 normalized \\
General open-source MLLMs & Timestamp in seconds & [xyxy] & 0--1000 normalized \\
LLaVA-ST & Temporal token & [xyxy] & temporal and spatial tokens in 0--99 \\
\bottomrule
\end{tabular}}
\end{table}

The full prompt text for each task and model family is listed below.

\subsection{Temporal Grounding} 
We use model-family-specific temporal grounding prompts.

\noindent\textbf{Gemini family and Gemma 4} (adapted from Vidi2.5~\cite{Vidi2026vidi25}).
\begin{tcolorbox}[colback=gray!10, title=\textbf{Prompt Template: Gemini Family (Temporal)}]
Answer with time ranges and do not output explanation. \\
What is the single time range corresponding to the text query: \{query\}? \\
Output format: [start, end]
\end{tcolorbox}

\noindent\textbf{GPT family} (adapted from Vidi2.5~\cite{Vidi2026vidi25}).

\begin{tcolorbox}[colback=gray!10, title=\textbf{Prompt Template: GPT Family (Temporal)}]
The input images are frames from a video. \\
Output the frame indexes that correspond to the text query: \{query\}. \\
Only output the index range (\eg, 2-4, 6-8).
\end{tcolorbox}

\noindent\textbf{General open-source MLLMs} (adapted from Qwen3-VL~\cite{Qwen3-VL}).
We use this template for Qwen3-VL, Qwen3.5, Qwen3.8, InternVL3, InternVL3.5, and Eagle 2.5.

\begin{tcolorbox}[colback=gray!10, title=\textbf{Prompt Template: General Open-Source MLLMs (Temporal)}]
Give you a textual query: \{query\} \\
When does the described content occur in the video? \\
Please return the timestamp in seconds. \\
Output format: [start, end].
\end{tcolorbox}

\noindent\textbf{LLaVA-ST~\cite{Li2025LLaVA-ST}.}

\begin{tcolorbox}[colback=gray!10, title=\textbf{Prompt Template: LLaVA-ST (Temporal)}]
Give you a textual query: \{query\} \\
When does the described content occur in the video? \\
Please return the start and end timestamps.
\end{tcolorbox}

\subsection{Spatial Grounding}
We use model-family-specific spatial grounding prompts.

\noindent\textbf{Gemini family and Gemma 4.}
\begin{tcolorbox}[colback=gray!10, title=\textbf{Prompt Template: Gemini Family (Spatial)}]
Answer with bounding boxes and do not output explanation. \\
Bounding box format: [y\_min, x\_min, y\_max, x\_max] with values in [0, 1000] normalized to the video frame. \\
What are all the positions corresponding to the text query: \{query\}? \\
Output format: \texttt{[{"timestamp":"00:30", "box\_2d":[100, 200, 300, 400]}, ...]}
\end{tcolorbox}

\noindent\textbf{GPT family.}
\begin{tcolorbox}[colback=gray!10, title=\textbf{Prompt Template: GPT Family (Spatial)}]
Given the frames of video, please find all the objects corresponding to the text query: \texttt{"\{query\}"} \\
Output only a JSON array. \\
Frame index format: 0-based integer. \\
Bounding box format: [x\_min, y\_min, x\_max, y\_max] with normalized value in 0.000 -- 1.000. \\
Example: \texttt{[{"frame\_id": 1, "bbox\_2d": [x\_min, y\_min, x\_max, y\_max]}, ...]}
\end{tcolorbox}

\noindent\textbf{General open-source MLLMs}.

\begin{tcolorbox}[colback=gray!10, title=\textbf{Prompt Template: General Open-Source MLLMs (Spatial)}]
Given the query "\{query\}", for each frame, detect and localize the visual content described by the textual query. \\
If the visual content does not exist in a frame, skip that frame. \\
Output the results in JSON format. \\
Example: \texttt{[{"timestamp": 1.0, "bbox\_2d": [x\_min, y\_min, x\_max, y\_max]}, ...]}
\end{tcolorbox}

\noindent\textbf{LLaVA-ST~\cite{Li2025LLaVA-ST}.}

\begin{tcolorbox}[colback=gray!10, title=\textbf{Prompt Template: LLaVA-ST (Spatial)}]
Between \texttt{\{<TEMP-000><TEMP-099>\}}, "\{query\}". \\
Please describe the location of the corresponding subject/object in this video. \\
Please give the spatial bounding box corresponding to each timestamp in the time period.
\end{tcolorbox}

\subsection{Spatio-Temporal Grounding}
We use model-family-specific spatio-temporal grounding prompts.

\noindent\textbf{Gemini family and Gemma 4} (adapted from Vidi2.5~\cite{Vidi2026vidi25}).
\begin{tcolorbox}[colback=gray!10, title=\textbf{Prompt Template: Gemini Family (Spatio-Temporal)}]
Answer with timestamps and bounding boxes and do not output explanation. \\
Output only a JSON array. \\
Timestamp format: MM:SS with zero-padding (00--59 for MM and SS). \\
Bounding box format: [y\_min, x\_min, y\_max, x\_max] with values in [0, 1000] normalized to the video frame. \\
Example: \texttt{[{"timestamp":"00:30", "box\_2d":[100, 200, 300, 400]}, ...]} \\
What are all the timestamps and positions corresponding to the text query: \texttt{"\{query\}"}?
\end{tcolorbox}

\noindent\textbf{GPT family} (adapted from Vidi2.5~\cite{Vidi2026vidi25}).

\begin{tcolorbox}[colback=gray!10, title=\textbf{Prompt Template: GPT Family (Spatio-Temporal)}]
Given the frames of video, please find all the objects corresponding to the text query: \texttt{"\{query\}"} \\
Output only a JSON array. \\
Frame index format: 0-based integer. \\
Bounding box format: [x\_min, y\_min, x\_max, y\_max] (0.000 -- 1.000 normalized). \\
Example: \texttt{[{"frame": 3, "box": [0.051, 0.252, 0.323, 0.954]}, ...]}
\end{tcolorbox}

\noindent\textbf{General open-source MLLMs} (adapted from Qwen3-VL~\cite{Qwen3-VL}).

\begin{tcolorbox}[colback=gray!10, title=\textbf{Prompt Template: General Open-Source MLLMs (Spatio-Temporal)}]
Given the query "\{query\}", for each frame, detect and localize the visual content described by the textual query. \\
If the visual content does not exist in a frame, skip that frame. \\
Output the results in JSON format. \\
Example: \texttt{[{"timestamp": 1.0, "bbox\_2d": [x\_min, y\_min, x\_max, y\_max]}, ...]}
\end{tcolorbox}

\noindent\textbf{LLaVA-ST~\cite{Li2025LLaVA-ST}.}

\begin{tcolorbox}[colback=gray!10, title=\textbf{Prompt Template: LLaVA-ST (Spatio-Temporal)}]
When does "\{query\}" occur in the video? \\
Please describe the location of the corresponding subject/object in this video. \\
Please firstly give the timestamps, and then give the spatial bounding box corresponding to each timestamp in the time period.
\end{tcolorbox}

\section{Details of Constituent Datasets}
\label{app:datasetdetails}
\paper\ consists of ten datasets spanning five specialized domains: animal, industry, sports, surgery, and public security. The specific characteristics, curation processes, and annotation enhancements for each constituent dataset are detailed below. For newly captured videos, we apply random split at the video level, while for existing video sources, we adhere to the original splits. Although Mouse Scratching was recorded with four synchronized cameras, each video segment is taken from a single view, so the same moment never appears from more than one view.

\noindent \textbf{Animal Kingdom~\cite{Ng2022AnimalKingdom}.}
Animal Kingdom is a large-scale dataset featuring 850 species across 50 hours of video. While it provides multiple annotation types, including action recognition and pose estimation, we specifically utilize the temporal grounding subset. We extended the original temporal boundaries by annotating per-frame spatial bounding boxes for the target animals. This converts the original temporal-only labels into a Spatio-Temporal Video Grounding (STVG) format, focusing on diverse behaviors in complex natural environments.

\begin{figure*}[tb] 
\centering
\includegraphics[width=\linewidth]{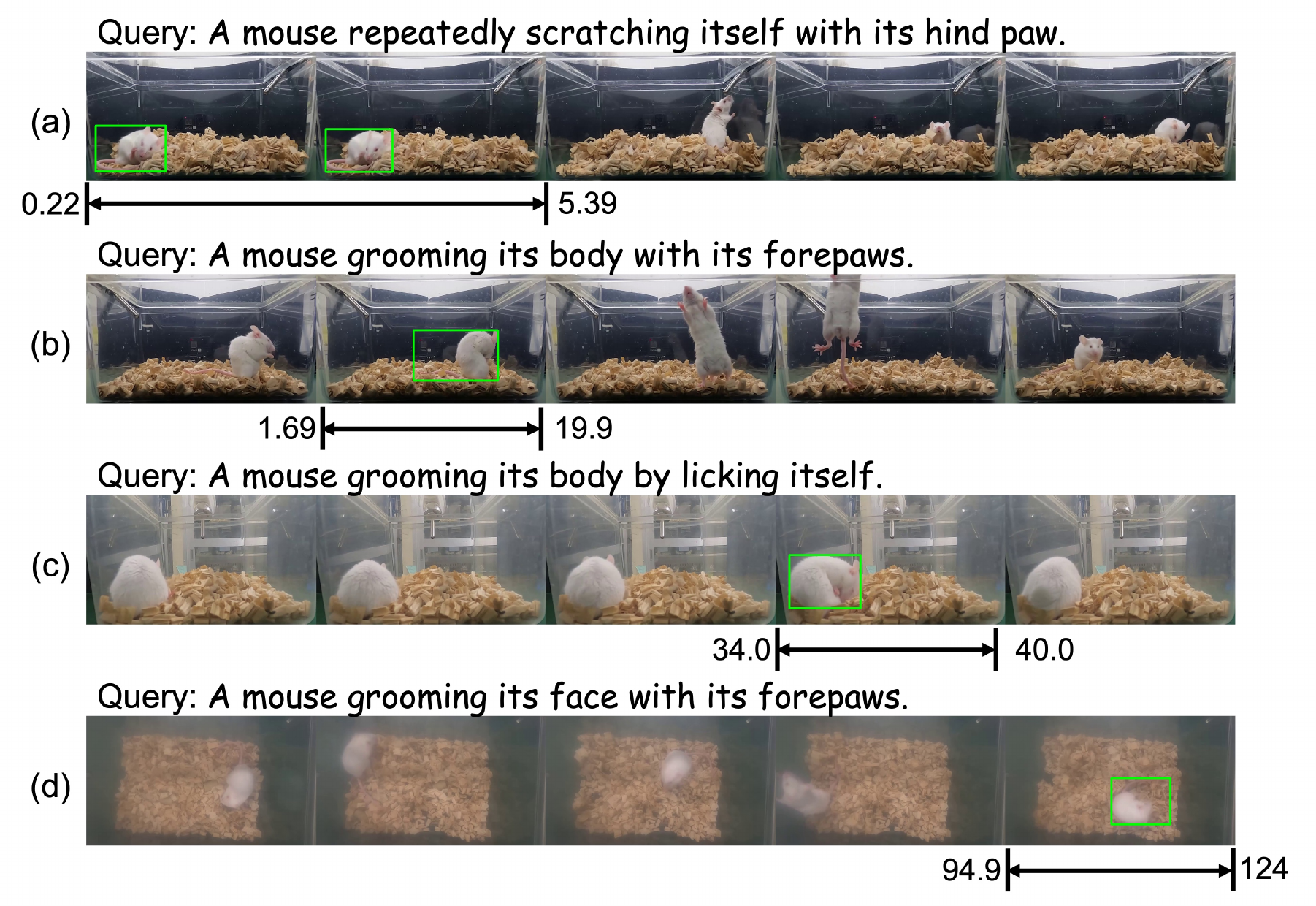}
\vspace{-1.em}
\caption{Examples from newly curated Mouse Scratching dataset.
The dataset features synchronized four perspectives—(a) Front view 1, (b) Front view 2, (c) Side view, and (d) Top-down spatial view—to capture scratching behaviors in diverse environments. The arrows and numbers (in seconds) in the figure indicate the temporal intervals in which events occur.}
\label{fig:mouse}
\vspace{-1.em}
\end{figure*}

\noindent \textbf{Mouse Scratching.} Mouse Scratching is a specialized, clinical-grade spatio-temporal grounding benchmark newly curated by two board-certified plastic surgeons to evaluate therapeutic efficacy in atopic dermatitis mouse models, with additional relevance to neurology and psychiatry for detecting nuanced behaviors like facial scratching. \autoref{fig:mouse} provides additional qualitative examples from the Mouse Scratching dataset, illustrating the diversity of behaviors and the four camera perspectives used for annotation. Comprising synchronized four-view video sequences, Mouse Scratching focuses on fine-grained, high-frequency motions—such as distinguishing rapid hind paw scratching from continuous forepaw grooming—that are often difficult to localize in clinical settings. The dataset introduces significant technical challenges, including motion blur and foggy environmental conditions, requiring models to perform robust spatio-temporal reasoning. The annotation schema follows a tripartite logic involving limb identification (hind paw vs. forepaw), target body parts, and action frequency (single vs. repeated), providing a high-fidelity resource for precise behavioral quantification.

\noindent \textbf{MECCANO~\cite{Francesco2023MECCANO}.}
MECCANO is an egocentric dataset capturing industrial-like assembly tasks, specifically the construction of a motorbike model. It provides dense annotations for human-object interactions (HOI), including both temporal action segments and spatial bounding boxes for active objects. We leverage these existing HOI labels, specifically utilizing the active object trajectories within their respective interaction windows, to establish the spatio-temporal grounding targets for the industrial domain.

\noindent \textbf{ENIGMA-51~\cite{Ragusa2024Enigma51}.}
ENIGMA-51 features egocentric videos of subjects repairing electrical boards using professional tools such as electric screwdrivers and oscilloscopes. While the original dataset provides rich interaction labels, spatial bounding boxes were primarily annotated only at discrete interaction key-frames. To adapt this for STVG, we established continuous spatio-temporal tubes by identifying strict interaction sessions (\eg, from \textit{take} to \textit{release}) and providing dense bounding box annotations for all intermediate frames where they were previously absent. Additionally, we synthesized object-centric natural language queries, such as ``the screwdriver being taken,'' to serve as the grounding targets. This effort transforms discrete action labels into a dense benchmark for evaluating fine-grained localization in technical industrial workflows.

\noindent \textbf{MultiSports~\cite{Li2021MultiSports}.}
MultiSports is a large-scale dataset for spatio-temporal action detection, covering 66 fine-grained action classes across four sports: basketball, volleyball, football, and aerobic gymnastics. It is characterized by multi-person scenes with concurrent actions and professional-level competition dynamics. We utilize the original high-quality, 25 fps frame-wise bounding boxes and temporal labels as the ground truth for our STVG benchmark. This dataset provides a rigorous test for models to handle high-velocity motions and distinguish between subtle, motion-dependent maneuvers in multi-agent environments.

\begin{figure*}[tb] 
\centering
\includegraphics[width=\linewidth]{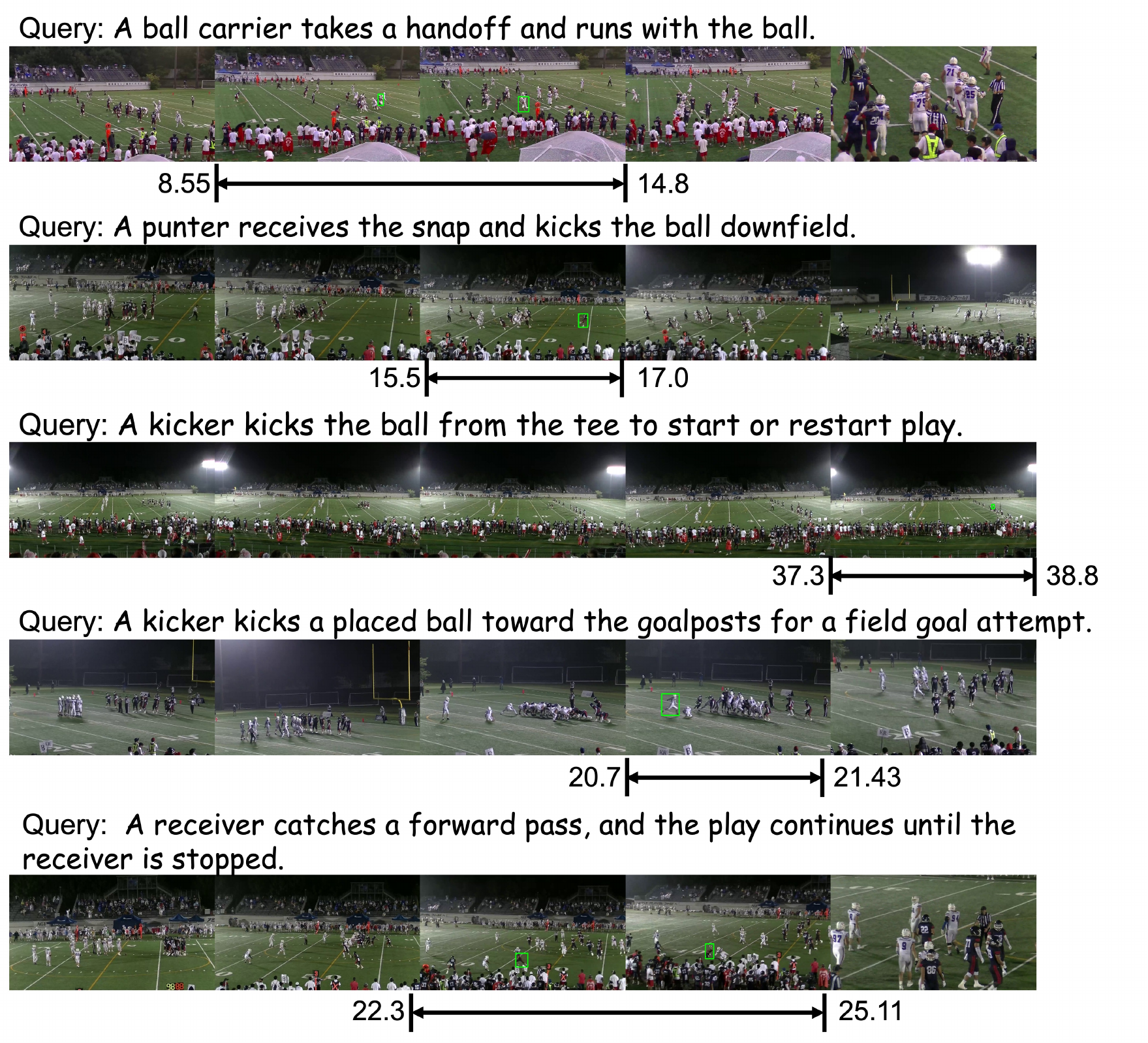}
\vspace{-1.em}
\caption{Examples from newly curated American Football dataset. The arrows and numbers (in seconds) in the figure indicate the temporal intervals in which events occur.}
\label{fig:amefoot}
\vspace{-1.em}
\end{figure*}

\noindent \textbf{American Football.} The American Football dataset, illustrated in \autoref{fig:amefoot}, is a specialized, expert-grade spatio-temporal grounding benchmark for American Football, newly curated by individuals with years of active playing experience to evaluate tactical execution and athletic performance. Comprising 123 play sequences totaling over one hour of footage, the dataset focuses on five core action categories—\textit{pass-to-end}, \textit{run-to-end}, \textit{snap-to-punt}, \textit{field-goal}, and \textit{kick-off}—that require precise identification of ball-handling transitions and player roles. The dataset introduces significant technical challenges, including high-speed player collisions, complex occlusions within the line of scrimmage, requiring models to perform robust spatio-temporal reasoning. The annotation schema follows a domain-specific logic that distinguishes nuanced movements, such as the exact moment of a hand-off in running plays, the punter's reception of the snap, or the specialized mechanics of place-kicking from the tee, providing a high-fidelity resource for automated football analysis and sports coaching.

\noindent \textbf{EgoSurgery~\cite{Fujii2024Egosurgeryphase,Fujii2024Egosurgerytool,Fujii2022Surgicaltool}.}
EgoSurgery is a large-scale dataset comprising egocentric open surgery videos captured from head-mounted cameras. It features dense annotations for surgical tools across 15 categories, hand-bounding boxes, and 9 surgical phases. We enriched this dataset for our benchmark by synthesizing textual queries that combine tool identities with their specific surgical context (\eg, "Needle holders during closure"). Characterized by significant tool-shape similarities and heavy occlusion by hands and tissues, EgoSurgery provides a highly challenging benchmark for STVG in real-world open surgical environments.

\noindent \textbf{CholecTrack20~\cite{Nwoye2025CholecTrack20}.}
CholecTrack20 is a specialized dataset for multi-class instrument tracking in laparoscopic surgery. While the original dataset provides dense spatial bounding boxes, track IDs, and surgical phases, it does not include natural language descriptions. We enriched this data by generating domain-specific textual queries that link instruments to their operative context (\eg, "Bipolar during gallbladder dissection"). By combining the existing high-quality spatial trajectories with these new queries, we establish a benchmark for STVG in technology-assisted surgical interventions.

\noindent \textbf{UCA~\cite{Yuan2024UCA,Sultani2018UCFCrime}.}
UCA provides fine-grained language descriptions for the UCF-Crime dataset, focusing on various anomalous events in public surveillance scenarios. We utilize the original textual queries and temporal segments as our grounding targets. To enable spatio-temporal evaluation, we supplemented these labels with new spatial bounding box annotations for the entities responsible for the anomalies. This dataset addresses the challenges of low-resolution, exocentric viewpoints typical in urban security monitoring. 

\noindent \textbf{DoTA~\cite{Yao2023DoTA}.}
DoTA is a large-scale egocentric dataset designed for the detection and localization of traffic accidents from a driving perspective. It contains 4,677 video sequences covering 18 anomaly categories, such as vehicle collisions and near-misses. For our benchmark, we utilize the original "when-where" annotations, which provide both the temporal boundaries of the anomalous events and the spatial bounding box tracklets of the involved participants. This dataset serves as a critical testbed for spatio-temporal grounding in highly dynamic, mobile environments characterized by complex ego-motion and rapid incident development.

\section{Data Collection and Annotation Details for Newly Captured Datasets}
\label{sec:newly_captured_dataset_annotation_details}
We provide additional details for the two newly captured datasets, Mouse Scratching and American Football, because their labels were created specifically for \paper.
In both cases, the annotation objective was to produce dense spatio-temporal tubes and query descriptions that are faithful to domain-specific event definitions while remaining usable for a unified STVG benchmark.

\subsection{Mouse Scratching}
\noindent \textbf{Video acquisition and textual query construction.}
Mouse Scratching videos were recorded in a medical-school environment using four synchronized GoPro cameras.
The four-view (two front views, side view, and top-down spatial view) setup was designed to reduce viewpoint ambiguity when distinguishing subtle behaviors around the face and forelimbs.
In mouse behavioral analysis~\cite{kalueff2016neurobiology}, forepaw-directed face touching is treated as grooming, whereas hind-paw-directed face touching is treated as scratching, and our annotations follow this distinction.
The textual queries were constructed by combining behavior type, contact limb, target region, and repetition pattern, as summarized in \autoref{tab:mouse_query_taxonomy}.

\begin{table}[t]
\centering
\caption{Taxonomy of textual queries in the Mouse Scratching dataset.}
\label{tab:mouse_query_taxonomy}
\resizebox{\linewidth}{!}{%
\begin{tabular}{llll}
\toprule
\textbf{Behavior} & \textbf{Limb / mode} & \textbf{Target / frequency} & \textbf{Example query} \\
\midrule
Grooming & licking & body & A mouse grooming its body by licking itself. \\
Grooming & forepaws & body & A mouse grooming its body with its forepaws. \\
Grooming & forepaws & face & A mouse grooming its face with its forepaws. \\
Grooming & forepaws & once & A mouse grooming itself once with its forepaws. \\
Grooming & forepaws & repeated & A mouse repeatedly grooming itself with its forepaws. \\
Scratching & hind paw & body & A mouse scratching its body with its hind paw. \\
Scratching & hind paw & face & A mouse scratching its face with its hind paw. \\
Scratching & hind paw & once & A mouse scratching itself once with its hind paw. \\
Scratching & hind paw & repeated & A mouse repeatedly scratching itself with its hind paw. \\
\bottomrule
\end{tabular}}
\end{table}

\noindent \textbf{Temporal annotation.}
Temporal ground truth was manually annotated by two medical-school professors who counted facial scratching events.
Following their annotation rule, one scratching event was defined from the moment the leg was raised to the moment it was lowered.
The temporal spans were labeled in ELAN, a manual multimodal annotation tool widely used for time-aligned video annotation.

\noindent \textbf{Spatial annotation.}
For spatial grounding, we used CVAT and created one annotation task for each temporally trimmed ground-truth segment.
In each task, we first manually annotated the mouse with a bounding box on the first frame of the segment.
We then applied SAM2 tracking to the temporally clipped frames, using the manually specified first-frame bounding box in pixel coordinates as the input prompt.
We binarized the SAM2 prediction logits with a threshold of 0 and converted the resulting masks into bounding boxes.
As quality control, we visually inspected the bounding boxes and manually corrected segments when clear tracking drift was observed.

\subsection{American Football}
\noindent \textbf{Video acquisition and textual query construction.}
The American Football dataset was recorded from university-level amateur games with handheld cameras.
Unlike professionally produced broadcast footage, these videos reflect real-world amateur recording conditions and often contain spectators and other non-player distractors in the scene.
The textual queries were constructed with an experienced American football player so that the linguistic descriptions were natural in the context of American football gameplay and aligned with football-specific event definitions.
The textual queries were organized around football-specific play outcomes and target roles, as summarized in \autoref{tab:football_query_taxonomy}.

\begin{table}[t]
\centering
\caption{Taxonomy of textual queries in the American Football dataset.}
\label{tab:football_query_taxonomy}
\resizebox{\linewidth}{!}{%
\begin{tabular}{lll}
\toprule
\textbf{Play type} & \textbf{Target role} & \textbf{Example query} \\
\midrule
Run play & ball carrier & A ball carrier takes a handoff and runs with the ball. \\
Pass play & receiver & A receiver catches a forward pass, and the play continues until the receiver is stopped. \\
Punt play & punter & A punter receives the snap and kicks the ball downfield. \\
Field goal & kicker & A kicker kicks a placed ball toward the goalposts for a field goal attempt. \\
Kickoff & kicker & A kicker kicks the ball from the tee to start or restart play. \\
\bottomrule
\end{tabular}}
\end{table}

\noindent \textbf{Temporal annotation.}
Temporal annotations were created by an annotator with several years of American football playing experience.
The videos were examined frame by frame at 30 fps to identify the start and end frames of each target event, and both the frame indices and their corresponding timestamps in seconds were recorded.

\noindent \textbf{Spatial annotation.}
For kick plays and pass plays, we first clipped the video to the annotated temporal interval, manually drew a bounding box on the first frame, and then tracked the target with the Ultralytics SAM3VideoPredictor.
The inputs to the tracker were the temporally clipped video and the manually annotated first-frame bounding box.
The model outputs both masks and bounding boxes, and we extracted the predicted spatial extent from the pixels with mask values greater than zero.
As quality control, segments with visible drift were manually corrected by additional bounding-box annotation.
For run plays, however, player density and severe occlusion made off-the-shelf tracking unreliable, so these segments were annotated fully by hand in CVAT.

\section{Licenses and Redistribution Constraints}
\label{app:licenses}
For curated benchmark components derived from existing datasets, each component remains subject to the license and terms of use of its source dataset.
Our release does not redistribute the original videos for these components; instead, users must obtain the source videos from the official dataset providers and comply with the corresponding source licenses.
We release the curated benchmark metadata needed for evaluation, including query definitions, splits, and derived grounding annotations where applicable.
For the newly captured Mouse Scratching and American Football datasets, we release the data under CC BY-NC-SA 4.0.

\begin{itemize}
    \item \textbf{Animal Kingdom}\footnote{Official repository:
    \url{https://github.com/sutdcv/Animal-Kingdom}.}: available through a dataset-use questionnaire.
    We filled out the questionnaire and obtained an official download link.
    We also emailed the authors about our use of the dataset in this paper.
    \item \textbf{Mouse Scratching}: newly captured dataset released under CC BY-NC-SA 4.0.
    \item \textbf{MECCANO}\footnote{Official repository:
    \url{https://github.com/fpv-iplab/MECCANO}.}: publicly available through the official provider.
    We emailed the authors about our use of the dataset.
    \item \textbf{ENIGMA-51}\footnote{Official repository:
    \url{https://github.com/fpv-iplab/ENIGMA-51}.}: publicly available through the official provider.
    We emailed the authors about our use of the dataset.
    \item \textbf{American Football}: newly captured dataset released under CC BY-NC-SA 4.0.
    \item \textbf{MultiSports}\footnote{Official repository:
    \url{https://github.com/MCG-NJU/MultiSports}.}: CC BY-NC 4.0;
    users should obtain the source dataset from the official provider and comply with the
    non-commercial license terms.
    \item \textbf{CholecTrack20}\footnote{Official repository:
    \url{https://github.com/CAMMA-public/cholectrack20}.}: CC BY-NC-SA 4.0
    and subject to the provider's data-use agreement; users should obtain the source dataset
    through the official access process.
    \item \textbf{UCA annotations}\footnote{Official repository:
    \url{https://github.com/Xuange923/Surveillance-Video-Understanding}.}: Apache-2.0, with academic/research-use restrictions
    stated by the dataset provider; the underlying UCF-Crime~\cite{Sultani2018UCFCrime} videos must be obtained
    separately from the original provider.
    \item \textbf{DoTA}\footnote{Official repository:
    \url{https://github.com/MoonBlvd/Detection-of-Traffic-Anomaly}.}: MIT; users should
    obtain the source dataset from the official provider and comply with any terms governing
    the released video clips and annotations.
    \item \textbf{EgoSurgery}\footnote{Official repository:
    \url{https://github.com/Fujiry0/EgoSurgery}.}: CC BY-NC-SA 4.0,
    limited by the provider to academic research and non-commercial use; users must obtain
    access through the official request process.
\end{itemize}

\section{Inter Annotator Agreement}
\label{app:iaa}
We report both raw percent agreement and Gwet's AC1~\citep{gwet2008computing} as a chance-corrected agreement statistic. We adopt AC1 over the more commonly used Cohen's $\kappa$~\citep{cohen1960coefficient}, since $\kappa$ is known to become unstable and can yield paradoxically low values when the underlying label distribution is highly skewed~\citep{feinstein1990high}, which is precisely the case in our setting, where the large majority of QC decisions are labeled as ``acceptable.'' Gwet's AC1 corrects for this prevalence effect while preserving a chance-corrected interpretation, making it better suited to our annotation setting. As shown in Table~\ref{tab:iaa}, agreement is consistently high across all domains, with AC1 ranging from 0.782 (Surgery) to 0.947 (Sports) and an overall value of 0.879, indicating substantial to near-perfect agreement according to standard interpretive guidelines~\citep{landis1977measurement}. Overall, these results confirm that our quality control decisions are highly reliable across all domains considered.

\begin{table}[h]
\centering
\caption{\textbf{Inter Annotator Agreement results.} Raw percent agreement and Gwet's AC1 between the original QC decision and an independent second annotator, per domain.}
\label{tab:iaa}
\begin{tabular}{lcc}
\toprule
Domain & Agreement (\%) & Gwet's AC1 \\
\midrule
Animal          & 90  & 0.885 \\
Industry        & 93  & 0.922 \\
Public Security & 87  & 0.843 \\
Sports          & 95  & 0.947 \\
Surgery         & 83  & 0.782 \\
\midrule
Overall         & 91 & 0.879 \\
\bottomrule
\end{tabular}
\end{table}

\begin{table*}[t]
\centering
\caption{\textbf{Zero-shot results of VLMs on the Spatio-Temporal Video Grounding (STVG) task.}
Each cell reports $m_{\text{t}}\mathrm{IoU}$ / $m_{\text{v}}\mathrm{IoU}$ / $v\mathrm{IoU}@0.3$ / $v\mathrm{IoU}@0.5$ (\%).
\textbf{Bold} and \underline{underlined} fonts represent the best and second-best results, respectively.} 
\label{tab:vlm_stvg_details}
\resizebox{\textwidth}{!}{
\begin{tabular}{l *{5}{c}}
\toprule
\textbf{Models} & \textcolor{AnimalColor}{\textbf{Animal}} & \textcolor{IndustryColor}{\textbf{Industry}} & \textcolor{SportsColor}{\textbf{Sports}} & \textcolor{SurgeryColor}{\textbf{Surgery}} & \textcolor{SecurityColor}{\textbf{Public Security}} \\
\midrule
\multicolumn{6}{c}{\textit{Proprietary VLMs}} \\
\midrule
GPT-4o & 9.71 / 1.59 / 0.00 / 0.00 & 10.8 / 2.55 / 2.96 / 0.59 & 18.8 / 0.77 / 0.61 / 0.00 & 11.0 / 1.10 / 0.00 / 0.00 & 28.1 / 4.89 / 0.80 / 0.00 \\
GPT-5.1 & 18.0 / 4.96 / 3.18 / 0.00 & 10.9 / 4.11 / 5.33 / 2.37 & 13.6 / 0.63 / 0.00 / 0.00 & 14.2 / 2.55 / 1.39 / 0.00 & 38.8 / 9.47 / 4.80 / 0.80 \\
GPT-5.4 & 22.4 / 8.99 / 6.37 / 2.55 & 10.5 / 3.27 / 4.73 / 1.18 & 19.4 / 1.38 / 0.61 / \underline{0.61} & 22.2 / 3.58 / 1.39 / 0.00 & 40.7 / 7.51 / 2.00 / 0.00 \\
GPT-5.6-Sol & 22.2 / 14.6 / 15.3 / \underline{8.28} & 16.8 / \textbf{7.33} / \textbf{10.7} / \textbf{5.33} & \textbf{38.1} / \textbf{6.95} / \textbf{6.13} / \textbf{1.23} & \underline{27.8} / \underline{9.14} / 7.87 / \underline{0.93} & 40.1 / \underline{20.6} / 20.8 / \underline{4.80} \\
GPT-5.6-Terra & 22.4 / 11.5 / 13.4 / 5.10 & 11.2 / 4.75 / 6.51 / \underline{3.55} & 35.2 / \underline{5.34} / \underline{4.91} / \textbf{1.23} & 23.9 / 6.07 / 4.17 / 0.00 & 40.8 / 18.2 / 18.4 / 2.80 \\
GPT-6-Astra & 23.3 / \textbf{17.5} / \textbf{19.1} / \textbf{10.2} & 11.8 / 5.93 / 7.10 / 2.37 & 17.3 / 5.12 / \underline{4.91} / \underline{0.61} & \textbf{31.1} / \textbf{18.3} / \textbf{25.9} / \textbf{11.1} & 36.7 / 19.4 / 18.8 / 3.20 \\
Gemini-2.5-Flash & 14.9 / 4.44 / 2.55 / 0.64 & 14.4 / 1.01 / 0.59 / 0.00 & 27.1 / 1.72 / 1.84 / \textbf{1.23} & 14.8 / 1.27 / 0.00 / 0.00 & 26.0 / 3.99 / 0.40 / 0.00 \\
Gemini-2.5-Pro & 23.6 / 8.44 / 8.28 / 3.18 & \textbf{21.8} / 1.92 / 1.18 / 0.00 & 35.4 / 2.17 / 0.61 / \underline{0.61} & 22.6 / 3.66 / 1.39 / 0.00 & 40.5 / 8.66 / 4.00 / 0.80 \\
Gemini-3-Flash & 29.2 / 12.8 / 14.6 / 5.73 & 15.5 / 4.94 / 5.92 / 2.37 & 32.8 / 2.78 / 2.45 / 0.00 & 12.8 / 3.44 / 0.93 / 0.00 & 38.8 / 13.2 / 11.2 / 1.60 \\
Gemini-3.1-Pro & 29.1 / 12.9 / 16.6 / 6.37 & \underline{19.7} / \underline{6.22} / \underline{7.69} / 2.37 & 31.2 / 3.26 / 1.23 / 0.00 & 19.0 / 6.55 / 4.17 / 0.46 & 40.5 / 17.0 / \underline{22.8} / 4.40 \\
Gemini-3.5-Flash & \underline{29.5} / 14.2 / 16.6 / \textbf{10.2} & 11.5 / 4.76 / 5.92 / 2.96 & 30.6 / 3.92 / \underline{4.91} / \textbf{1.23} & 11.2 / 5.26 / 4.17 / 0.46 & \underline{42.6} / 17.5 / 22.4 / 4.00 \\
Gemini-3.8-Flash & \textbf{30.3} / \underline{14.8} / \underline{17.2} / 7.01 & 12.8 / 5.27 / \underline{7.69} / \underline{3.55} & \underline{36.7} / 5.30 / 4.29 / 0.00 & 16.3 / 8.09 / \underline{9.26} / 0.46 & \textbf{46.2} / \textbf{22.3} / \textbf{28.4} / \textbf{8.80} \\
\midrule
\multicolumn{6}{c}{\textit{Open-source Specialized VLMs}} \\
\midrule
LLaVA-ST & 17.5 / 7.85 / 12.1 / 2.55 & 3.67 / 0.77 / 0.00 / 0.00 & 15.6 / 0.66 / 0.79 / 0.00 & 7.09 / 0.57 / 0.00 / 0.00 & 32.5 / 3.98 / 0.80 / 0.00 \\
\midrule
\multicolumn{6}{c}{\textit{Open-source General-Purpose VLMs}} \\
\midrule
Qwen3-VL-4B & 16.4 / 6.29 / 5.73 / 1.27 & 6.14 / 0.58 / 0.00 / 0.00 & 10.9 / 0.07 / 0.00 / 0.00 & 11.6 / 0.94 / 0.00 / 0.00 & 22.5 / 2.20 / 0.40 / 0.00 \\
Qwen3-VL-8B & 18.7 / 4.77 / 3.82 / 1.27 & 6.76 / 0.82 / 0.00 / 0.00 & 11.8 / 0.03 / 0.00 / 0.00 & 10.3 / 0.78 / 0.00 / 0.00 & 27.7 / 2.28 / 0.80 / 0.00 \\
Qwen3.5-4B & 24.1 / 4.29 / 2.55 / 0.64 & 10.8 / 0.46 / 0.00 / 0.00 & 15.3 / 0.13 / 0.00 / 0.00 & 13.3 / 0.66 / 0.00 / 0.00 & 31.7 / 1.56 / 0.40 / 0.00 \\
Qwen3.5-9B & 24.6 / 4.33 / 4.46 / 0.64 & 9.05 / 1.07 / 0.59 / 0.00 & 14.1 / 0.22 / 0.61 / 0.00 & 13.2 / 0.74 / 0.00 / 0.00 & 32.1 / 1.32 / 0.40 / 0.00 \\
Qwen3.8-27B & 24.8 / 10.4 / 12.7 / 5.73 & 13.0 / 4.49 / 5.33 / 2.96 & 20.6 / 0.79 / 0.00 / 0.00 & 15.3 / 2.45 / 0.00 / 0.00 & 39.0 / 13.6 / 13.6 / 2.40 \\
Eagle2.5-8B & 3.63 / 0.46 / 0.00 / 0.00 & 5.64 / 0.26 / 0.00 / 0.00 & 2.77 / 0.03 / 0.00 / 0.00 & 3.10 / 0.18 / 0.00 / 0.00 & 10.4 / 0.73 / 0.00 / 0.00 \\
InternVL3-8B & 11.0 / 1.04 / 0.64 / 0.00 & 3.69 / 0.19 / 0.00 / 0.00 & 4.13 / 0.00 / 0.00 / 0.00 & 6.30 / 0.20 / 0.00 / 0.00 & 7.00 / 0.31 / 0.00 / 0.00 \\
InternVL3-14B & 14.1 / 1.88 / 0.64 / 0.00 & 3.37 / 0.12 / 0.00 / 0.00 & 12.3 / 0.01 / 0.00 / 0.00 & 7.55 / 0.18 / 0.00 / 0.00 & 22.8 / 0.75 / 0.00 / 0.00 \\
InternVL3.5-8B & 15.5 / 0.33 / 0.00 / 0.00 & 5.53 / 0.03 / 0.00 / 0.00 & 7.64 / 0.01 / 0.00 / 0.00 & 5.89 / 0.13 / 0.00 / 0.00 & 13.0 / 0.06 / 0.00 / 0.00 \\
Gemma4-12B & 19.1 / 6.40 / 5.10 / 1.27 & 9.29 / 2.93 / 4.14 / 1.78 & 6.47 / 0.42 / 0.00 / 0.00 & 20.6 / 3.72 / 2.31 / 0.00 & 12.1 / 2.87 / 1.60 / 0.40 \\
\bottomrule
\end{tabular}}
\end{table*}

\begin{table*}[t]
\centering
\caption{\textbf{Zero-shot results of VLMs on the Temporal Video Grounding (TVG) task.}
Each cell reports $m_{\text{t}}\mathrm{IoU}$ / $t\mathrm{IoU}@0.3$ / $t\mathrm{IoU}@0.5$ / $t\mathrm{IoU}@0.7$ (\%).
\textbf{Bold} and \underline{underlined} fonts represent the best and second-best results, respectively.} 
\label{tab:vlm_tvg_details}
\resizebox{\textwidth}{!}{
\begin{tabular}{l *{5}{c}}
\toprule
\textbf{Models} & \textcolor{AnimalColor}{\textbf{Animal}} & \textcolor{IndustryColor}{\textbf{Industry}} & \textcolor{SportsColor}{\textbf{Sports}} & \textcolor{SurgeryColor}{\textbf{Surgery}} & \textcolor{SecurityColor}{\textbf{Public Security}} \\
\midrule
\multicolumn{6}{c}{\textit{Proprietary VLMs}} \\
\midrule
GPT-4o & 11.8 / 16.6 / 9.55 / 5.10 & 6.92 / 8.88 / 5.92 / 4.14 & 13.7 / 17.2 / 6.75 / 2.45 & 10.8 / 13.4 / 8.80 / 3.24 & 33.9 / 56.0 / 27.2 / 6.80 \\
GPT-5.1 & 14.1 / 19.7 / 12.7 / 7.01 & 6.22 / 9.47 / 6.51 / 3.55 & 18.5 / 25.8 / 12.3 / 3.07 & 17.1 / 22.7 / 18.1 / 9.26 & 38.6 / 61.2 / 34.4 / 12.8 \\
GPT-5.4 & 16.9 / 21.7 / 14.0 / 7.64 & 9.74 / 12.4 / 7.69 / 1.78 & 21.1 / 32.5 / 16.0 / 2.45 & 26.9 / \underline{38.9} / 25.5 / 12.5 & 43.8 / 68.0 / 41.6 / \underline{16.4} \\
GPT-5.6-Sol & 19.6 / 25.5 / 18.5 / 10.2 & 15.8 / 24.3 / 13.6 / 6.51 & \underline{29.0} / \underline{44.2} / \underline{27.6} / \underline{12.9} & 23.9 / 32.4 / 21.8 / 14.8 & 42.7 / 67.2 / 38.4 / \textbf{17.2} \\
GPT-5.6-Terra & 17.3 / 22.3 / 15.3 / 7.64 & 10.7 / 17.2 / 9.47 / 1.18 & 20.8 / 31.9 / 18.4 / 7.36 & 22.5 / 32.4 / 21.8 / 11.1 & 35.1 / 56.8 / 24.8 / 7.20 \\
GPT-6-Astra & 28.2 / 36.3 / \underline{27.4} / \textbf{19.1} & 17.2 / 24.9 / 11.2 / 4.73 & \textbf{32.4} / \textbf{46.0} / \textbf{29.4} / \textbf{16.0} & \textbf{36.9} / \textbf{47.7} / \textbf{38.4} / \textbf{27.8} & \textbf{46.2} / \textbf{76.0} / \textbf{44.0} / \underline{16.4} \\
Gemini-2.5-Flash & 20.8 / 26.1 / 20.4 / 13.4 & 14.7 / 21.3 / 10.7 / 4.73 & 19.8 / 31.9 / 14.1 / 4.29 & 18.0 / 22.7 / 14.8 / 8.33 & 31.3 / 51.2 / 28.4 / 9.60 \\
Gemini-2.5-Pro & 26.5 / 36.9 / 26.1 / 13.4 & \textbf{26.8} / \textbf{39.6} / \textbf{26.6} / \textbf{11.8} & 25.7 / 37.4 / 22.7 / 5.52 & 25.1 / 31.5 / 22.7 / 15.7 & 41.6 / 65.9 / 41.0 / 16.1 \\
Gemini-3-Flash & \underline{28.4} / 36.3 / 24.8 / 15.9 & \underline{21.0} / \underline{30.8} / \underline{16.6} / 8.28 & 22.4 / 29.4 / 15.3 / 6.13 & \underline{30.0} / 38.0 / \underline{30.6} / \underline{20.8} & 40.8 / 66.8 / 38.8 / 10.4 \\
Gemini-3.1-Pro & 26.4 / 37.6 / 24.8 / 15.9 & 18.4 / 21.9 / 13.0 / \underline{8.88} & 19.3 / 26.4 / 10.4 / 3.07 & 25.5 / 32.9 / 23.1 / 14.4 & 41.3 / 69.5 / 39.0 / 12.9 \\
Gemini-3.5-Flash & \textbf{30.3} / \textbf{41.4} / \textbf{29.9} / 15.9 & 10.8 / 19.5 / 8.88 / 3.55 & 21.6 / 30.1 / 12.9 / 3.68 & 23.2 / 30.1 / 24.1 / 13.9 & 36.8 / 61.6 / 32.8 / 10.8 \\
Gemini-3.8-Flash & 27.8 / \underline{38.9} / \underline{27.4} / \underline{16.6} & 11.8 / 20.1 / 7.10 / 1.78 & 25.0 / 36.2 / 17.2 / 5.52 & 28.6 / 35.6 / 29.2 / 15.3 & \underline{43.9} / \underline{75.2} / \underline{42.8} / 15.2 \\
\midrule
\multicolumn{6}{c}{\textit{Open-source Specialized VLMs}} \\
\midrule
LLaVA-ST & 15.9 / 19.7 / 14.6 / 8.28 & 6.55 / 9.47 / 4.73 / 0.59 & 6.27 / 8.59 / 1.84 / 0.61 & 9.51 / 12.0 / 5.09 / 2.78 & 24.4 / 35.6 / 17.6 / 5.20 \\
\midrule
\multicolumn{6}{c}{\textit{Open-source General-Purpose VLMs}} \\
\midrule
Qwen3-VL-4B & 18.9 / 25.5 / 15.9 / 10.8 & 5.13 / 4.14 / 3.55 / 1.18 & 10.6 / 8.59 / 3.68 / 0.61 & 10.8 / 13.4 / 6.48 / 4.17 & 24.8 / 39.2 / 12.0 / 4.00 \\
Qwen3-VL-8B & 14.1 / 19.7 / 11.5 / 5.73 & 8.34 / 10.7 / 5.92 / 0.59 & 10.6 / 11.7 / 1.23 / 0.00 & 10.6 / 15.7 / 5.09 / 3.70 & 29.4 / 46.0 / 16.0 / 2.00 \\
Qwen3.5-4B & 24.5 / 30.6 / 21.0 / 12.7 & 12.5 / 14.8 / 10.1 / 2.37 & 12.2 / 12.9 / 4.91 / 0.00 & 20.0 / 28.2 / 16.7 / 10.2 & 32.2 / 49.2 / 17.2 / 4.80 \\
Qwen3.5-9B & 27.3 / 35.0 / 24.2 / 15.3 & 12.4 / 17.2 / 7.69 / 1.18 & 13.1 / 14.7 / 4.91 / 0.00 & 23.6 / 28.2 / 21.3 / 17.6 & 33.0 / 50.4 / 16.0 / 5.20 \\
Qwen3.8-27B & 22.4 / 26.1 / 15.3 / 8.92 & 13.4 / 18.3 / 9.47 / 1.18 & 12.6 / 11.0 / 3.07 / 0.61 & 25.9 / 35.2 / 22.7 / 12.5 & 36.2 / 56.4 / 28.0 / 7.20 \\
Eagle2.5-8B & 16.3 / 24.8 / 14.0 / 7.01 & 5.12 / 7.69 / 2.96 / 1.18 & 9.41 / 7.98 / 3.68 / 0.00 & 10.3 / 15.3 / 8.33 / 3.24 & 29.2 / 47.6 / 20.8 / 5.60 \\
InternVL3-8B & 11.8 / 15.3 / 8.92 / 3.82 & 4.76 / 5.92 / 2.37 / 1.18 & 7.86 / 6.79 / 2.47 / 0.00 & 8.61 / 10.6 / 3.70 / 1.39 & 3.41 / 4.40 / 2.00 / 0.40 \\
InternVL3-14B & 12.7 / 17.8 / 5.73 / 3.18 & 3.59 / 5.33 / 1.18 / 0.59 & 7.38 / 7.36 / 3.07 / 0.00 & 8.41 / 10.6 / 3.70 / 0.46 & 8.71 / 13.2 / 5.20 / 2.80 \\
InternVL3.5-8B & 9.54 / 11.5 / 7.64 / 1.27 & 4.06 / 2.96 / 1.18 / 0.00 & 4.49 / 6.13 / 3.68 / 1.84 & 7.07 / 7.41 / 2.78 / 0.93 & 2.03 / 3.60 / 1.60 / 0.80 \\
Gemma4-12B & 18.3 / 27.4 / 14.0 / 8.28 & 8.67 / 11.2 / 5.92 / 3.55 & 8.02 / 9.20 / 1.84 / 1.23 & 17.3 / 25.0 / 15.7 / 7.87 & 27.5 / 43.2 / 22.0 / 6.40 \\
\bottomrule
\end{tabular}}
\end{table*}

\begin{table*}[t]
\centering
\caption{\textbf{Zero-shot results of VLMs on the Spatial Video Grounding (SVG) task.}
Each cell reports $m_{\text{s}}\mathrm{IoU}$ / $s\mathrm{IoU}@0.3$ / $s\mathrm{IoU}@0.5$ (\%).
\textbf{Bold} and \underline{underlined} fonts represent the best and second-best results, respectively.}
\label{tab:vlm_svg_details}
\resizebox{\textwidth}{!}{
\begin{tabular}{l *{5}{c}}
\toprule
\textbf{Models} & \textcolor{AnimalColor}{\textbf{Animal}} & \textcolor{IndustryColor}{\textbf{Industry}} & \textcolor{SportsColor}{\textbf{Sports}} & \textcolor{SurgeryColor}{\textbf{Surgery}} & \textcolor{SecurityColor}{\textbf{Public Security}} \\
\midrule
\multicolumn{6}{c}{\textit{Proprietary VLMs}} \\
\midrule
GPT-4o & 7.90 / 7.64 / 1.91 & 9.75 / 14.8 / 2.96 & 0.63 / 0.00 / 0.00 & 1.88 / 0.00 / 0.00 & 6.57 / 4.40 / 1.60 \\
GPT-5.1 & 30.7 / 55.4 / 22.9 & 17.7 / 28.4 / 14.8 & 1.79 / 1.84 / 1.23 & 5.81 / 0.67 / 0.00 & 22.7 / 35.6 / 12.0 \\
GPT-5.4 & 41.2 / 65.6 / 40.1 & 20.6 / 33.1 / 15.4 & 5.87 / 5.52 / 1.84 & 9.45 / 2.68 / 0.00 & 30.7 / 53.6 / 16.4 \\
GPT-5.6-Sol & \underline{72.8} / \underline{92.4} / \underline{86.0} & \underline{34.6} / \underline{47.9} / \underline{37.9} & \underline{30.2} / \underline{45.4} / \underline{33.1} & 19.8 / 27.5 / 2.68 & \underline{70.6} / \underline{88.4} / \underline{81.2} \\
GPT-5.6-Terra & 63.5 / 86.0 / 73.9 & 32.5 / 43.8 / 36.7 & 20.1 / 31.3 / 22.7 & 13.9 / 10.1 / 0.67 & 64.1 / 83.2 / 75.6 \\
GPT-6-Astra & \textbf{81.9} / \textbf{97.5} / \textbf{92.4} & \textbf{44.5} / \textbf{53.8} / \textbf{45.6} & \textbf{44.5} / \textbf{58.3} / \textbf{52.1} & \textbf{29.8} / \textbf{61.7} / \textbf{6.04} & \textbf{81.2} / \textbf{93.6} / \textbf{90.8} \\
Gemini-2.5-Flash & 12.7 / 15.3 / 5.73 & 4.17 / 4.73 / 2.96 & 2.80 / 3.68 / 2.45 & 4.32 / 0.67 / 0.00 & 3.07 / 2.00 / 0.00 \\
Gemini-2.5-Pro & 14.0 / 20.4 / 9.55 & 14.3 / 17.2 / 10.1 & 6.19 / 8.59 / 4.29 & 3.22 / 3.36 / 0.00 & 15.8 / 22.8 / 10.0 \\
Gemini-3-Flash & 37.5 / 52.2 / 36.9 & 15.8 / 20.1 / 10.7 & 4.19 / 2.45 / 1.84 & 12.2 / 13.4 / 1.34 & 29.8 / 45.6 / 22.8 \\
Gemini-3.1-Pro & 49.2 / 70.7 / 49.0 & 31.1 / 41.4 / 32.0 & 13.1 / 16.6 / 7.98 & 19.1 / 34.2 / 2.68 & 33.2 / 52.0 / 28.4 \\
Gemini-3.5-Flash & 46.7 / 66.9 / 46.5 & 32.4 / 47.3 / 31.4 & 12.2 / 15.3 / 8.59 & 18.7 / 29.5 / \textbf{6.04} & 34.1 / 52.0 / 28.0 \\
Gemini-3.8-Flash & 46.1 / 66.2 / 42.7 & 27.3 / 37.3 / 23.7 & 9.68 / 12.9 / 5.52 & \underline{23.9} / \underline{41.6} / \underline{5.37} & 37.0 / 56.0 / 33.6 \\
\midrule
\multicolumn{6}{c}{\textit{Open-source Specialized VLMs}} \\
\midrule
LLaVA-ST & 32.2 / 53.5 / 23.6 & 10.7 / 12.4 / 5.33 & 4.69 / 3.68 / 1.84 & 4.72 / 0.00 / 0.00 & 13.9 / 13.2 / 2.40 \\
\midrule
\multicolumn{6}{c}{\textit{Open-source General-Purpose VLMs}} \\
\midrule
Qwen3-VL-4B & 28.2 / 42.0 / 16.6 & 9.59 / 13.0 / 2.96 & 0.08 / 0.00 / 0.00 & 3.36 / 0.00 / 0.00 & 3.87 / 0.00 / 0.00 \\
Qwen3-VL-8B & 15.2 / 17.8 / 1.27 & 7.90 / 7.69 / 2.37 & 0.44 / 0.00 / 0.00 & 1.89 / 0.00 / 0.00 & 5.05 / 3.60 / 0.00 \\
Qwen3.5-4B & 13.1 / 13.4 / 3.18 & 6.97 / 7.69 / 2.37 & 0.11 / 0.00 / 0.00 & 3.19 / 1.34 / 0.00 & 2.26 / 2.00 / 0.40 \\
Qwen3.5-9B & 17.4 / 20.4 / 7.01 & 8.80 / 12.4 / 1.18 & 0.13 / 0.00 / 0.00 & 4.62 / 1.34 / 0.00 & 5.10 / 4.80 / 0.80 \\
Qwen3.8-27B & 9.00 / 12.7 / 7.64 & 17.8 / 28.4 / 16.6 & 1.94 / 1.23 / 1.23 & 10.9 / 10.1 / 0.67 & 2.54 / 2.80 / 0.40 \\
Eagle2.5-8B & 1.32 / 1.27 / 0.00 & 1.90 / 2.37 / 0.00 & 0.06 / 0.00 / 0.00 & 0.58 / 0.00 / 0.00 & 0.33 / 0.40 / 0.00 \\
InternVL3-8B & 4.64 / 3.82 / 0.00 & 1.26 / 1.18 / 0.00 & 0.01 / 0.00 / 0.00 & 0.69 / 0.00 / 0.00 & 0.61 / 0.80 / 0.00 \\
InternVL3-14B & 7.73 / 7.64 / 1.91 & 2.06 / 1.18 / 0.00 & 0.00 / 0.00 / 0.00 & 0.92 / 0.00 / 0.00 & 0.57 / 0.00 / 0.00 \\
InternVL3.5-8B & 1.34 / 0.00 / 0.00 & 0.81 / 1.18 / 0.00 & 0.00 / 0.00 / 0.00 & 0.44 / 0.00 / 0.00 & 0.21 / 0.00 / 0.00 \\
Gemma4-12B & 2.09 / 3.82 / 1.27 & 12.9 / 18.3 / 12.4 & 0.00 / 0.00 / 0.00 & 2.31 / 2.01 / 1.34 & 0.18 / 0.00 / 0.00 \\
\bottomrule
\end{tabular}}
\end{table*}

\section{Results with Additional Metrics}
\label{app:detailed_results}
We report mean IoUs and stricter thresholds for zero-shot results in \autoref{tab:vlm_stvg_details}, \autoref{tab:vlm_tvg_details}, and \autoref{tab:vlm_svg_details}, and for adaptation results in \autoref{tab:adaptation_stvg_details}, \autoref{tab:adaptation_tvg_details}, and \autoref{tab:adaptation_svg_details}.
For STVG, $m_{\text{t}}\mathrm{IoU}$ denotes the temporal IoU of the predicted tube.

\noindent \textbf{Detailed Zero-Shot Results.}
Stricter thresholds confirm that zero-shot STVG remains far from practical: the best $v\mathrm{IoU}@0.5$ is only 11.1 (surgery, GPT-6-Astra) and at most 1.23 in sports, and all open-source models except Gemma4-12B and Qwen3.8-27B score 0.00 outside animal.
For open-source models, spatial grounding is the main bottleneck; for instance, in public security, the Qwen models achieve TVG $m_{\text{t}}\mathrm{IoU}$ of 24.8--36.2 but $m_{\text{s}}\mathrm{IoU}$ of at most 5.10.
For GPT-6-Astra, the temporal IoU of its STVG predictions is lower than its TVG $m_{\text{t}}\mathrm{IoU}$ in all five domains (\eg, 17.3 vs.\ 32.4 in sports and 36.7 vs.\ 46.2 in public security), indicating that predicting space and time jointly degrades even temporal localization.
Moreover, high scores at the 0.3 threshold partly reflect coarse localization.
In public security, TVG reaches 76.0 at $t\mathrm{IoU}@0.3$ but at most 17.2 at $t\mathrm{IoU}@0.7$.
In surgery, GPT-6-Astra's SVG drops from 61.7 at $s\mathrm{IoU}@0.3$ to 6.04 at $s\mathrm{IoU}@0.5$.

\begin{table*}[t]
\centering
\caption{\textbf{Adaptation results on the Spatio-Temporal Video Grounding (STVG) task.}
Each cell reports $m_{\text{t}}\mathrm{IoU}$ / $m_{\text{v}}\mathrm{IoU}$ / $v\mathrm{IoU}@0.3$ / $v\mathrm{IoU}@0.5$.
The first row shows the zero-shot baseline; subsequent rows show adaptation using the provided per-domain training subsets. \colorbox{gray!12}{Shaded} rows denote fine-tuning-based adaptation, while unshaded adaptation rows are training-free.
\poscol{Purple} denotes performance improvements or ties and \negcol{red} denotes degradations relative to the zero-shot baseline.}
\label{tab:adaptation_stvg_details}
\resizebox{\textwidth}{!}{
\begin{tabular}{l *{5}{c}}
\toprule
\textbf{Models} & \textcolor{AnimalColor}{\textbf{Animal}} & \textcolor{IndustryColor}{\textbf{Industry}} & \textcolor{SportsColor}{\textbf{Sports}} & \textcolor{SurgeryColor}{\textbf{Surgery}} & \textcolor{SecurityColor}{\textbf{Public Security}} \\
\midrule
\multicolumn{6}{c}{\textit{Proprietary VLM (Gemini-3.1-Pro)}} \\
\midrule
Zero-shot & 29.1 / 12.9 / 16.6 / 6.37 & 19.7 / 6.22 / 7.69 / 2.37 & 31.2 / 3.26 / 1.23 / 0.00 & 19.0 / 6.55 / 4.17 / 0.46 & 40.5 / 17.0 / 22.8 / 4.40 \\
~+ICL & \negcol{26.6} / \negcol{11.0} / \negcol{12.7} / \poscol{6.37} & \poscol{20.4} / \poscol{8.37} / \poscol{11.8} / \poscol{5.92} & \negcol{22.8} / \negcol{1.51} / \poscol{1.23} / \poscol{0.00} & \poscol{19.8} / \poscol{6.93} / \poscol{9.72} / \poscol{4.17} & \poscol{42.9} / \poscol{17.8} / \negcol{22.4} / \poscol{6.80} \\
~+GEPA & \poscol{31.0} / \poscol{14.4} / \poscol{17.2} / \poscol{8.28} & \poscol{29.2} / \poscol{10.1} / \poscol{13.0} / \poscol{5.92} & \poscol{34.6} / \poscol{3.60} / \poscol{2.45} / \poscol{0.00} & \poscol{33.9} / \poscol{15.3} / \poscol{21.8} / \poscol{9.72} & \poscol{40.6} / \poscol{17.1} / \negcol{19.2} / \negcol{4.00} \\
\midrule
\multicolumn{6}{c}{\textit{Open-source VLM (Qwen3-VL-8B)}} \\
\midrule
Zero-shot & 18.7 / 4.77 / 3.82 / 1.27 & 6.76 / 0.82 / 0.00 / 0.00 & 11.8 / 0.03 / 0.00 / 0.00 & 10.3 / 0.78 / 0.00 / 0.00 & 27.7 / 2.28 / 0.80 / 0.00 \\
~+ICL & \negcol{17.5} / \poscol{5.62} / \poscol{3.82} / \poscol{1.27} & \poscol{16.8} / \poscol{3.99} / \poscol{3.55} / \poscol{1.78} & \poscol{26.9} / \poscol{0.29} / \poscol{0.00} / \poscol{0.00} & \negcol{9.42} / \poscol{1.52} / \poscol{0.93} / \poscol{0.00} & \poscol{30.6} / \poscol{5.74} / \poscol{4.00} / \poscol{0.40} \\
~+GEPA & \negcol{16.8} / \negcol{3.84} / \negcol{3.18} / \negcol{0.00} & \poscol{6.77} / \negcol{0.20} / \poscol{0.00} / \poscol{0.00} & \negcol{8.94} / \poscol{0.12} / \poscol{0.00} / \poscol{0.00} & \negcol{6.21} / \negcol{0.55} / \poscol{0.00} / \poscol{0.00} & \negcol{24.8} / \poscol{2.28} / \negcol{0.40} / \poscol{0.00} \\
\rowcolor{gray!12}
~+SFT LoRA & \negcol{17.0} / \poscol{7.73} / \poscol{8.92} / \poscol{3.18} & \negcol{6.40} / \poscol{0.99} / \poscol{0.59} / \poscol{0.00} & \poscol{20.0} / \poscol{0.60} / \poscol{0.00} / \poscol{0.00} & \poscol{19.1} / \poscol{2.66} / \poscol{0.46} / \poscol{0.00} & \poscol{30.3} / \poscol{6.63} / \poscol{4.00} / \poscol{0.40} \\
\rowcolor{gray!12}
~+SFT Full Fine-tuning & \negcol{15.3} / \poscol{8.93} / \poscol{10.2} / \poscol{4.46} & \poscol{8.78} / \poscol{2.62} / \poscol{2.96} / \poscol{1.18} & \poscol{18.6} / \poscol{0.64} / \poscol{0.61} / \poscol{0.00} & \poscol{20.4} / \poscol{5.62} / \poscol{4.63} / \poscol{0.46} & \poscol{28.1} / \poscol{8.61} / \poscol{8.00} / \poscol{0.80} \\
\rowcolor{gray!12}
~+GRPO LoRA
& \negcol{18.6} / \poscol{10.6} / \poscol{11.5} / \poscol{5.10}
& \negcol{5.99} / \poscol{1.52} / \poscol{0.59} / \poscol{0.00}
& \poscol{19.6} / \poscol{0.86} / \poscol{0.61} / \poscol{0.00}
& \poscol{19.2} / \poscol{4.62} / \poscol{3.70} / \poscol{0.93}
& \negcol{26.9} / \poscol{8.48} / \poscol{7.20} / \poscol{1.60} \\
\bottomrule
\end{tabular}}
\end{table*}

\begin{table*}[t]
\centering
\caption{\textbf{Adaptation results on the Temporal Video Grounding (TVG) task.}
Each cell reports $m_{\text{t}}\mathrm{IoU}$ / $t\mathrm{IoU}@0.3$ / $t\mathrm{IoU}@0.5$ / $t\mathrm{IoU}@0.7$.
The first row shows the zero-shot baseline; subsequent rows show adaptation using the provided per-domain training subsets. \colorbox{gray!12}{Shaded} rows denote fine-tuning-based adaptation, while unshaded adaptation rows are training-free.
\poscol{Purple} denotes performance improvements or ties and \negcol{red} denotes degradations relative to the zero-shot baseline.}
\label{tab:adaptation_tvg_details}
\resizebox{\textwidth}{!}{
\begin{tabular}{l *{5}{c}}
\toprule
\textbf{Models} & \textcolor{AnimalColor}{\textbf{Animal}} & \textcolor{IndustryColor}{\textbf{Industry}} & \textcolor{SportsColor}{\textbf{Sports}} & \textcolor{SurgeryColor}{\textbf{Surgery}} & \textcolor{SecurityColor}{\textbf{Public Security}} \\
\midrule
\multicolumn{6}{c}{\textit{Proprietary VLM (Gemini-3.1-Pro)}} \\
\midrule
Zero-shot & 26.4 / 37.6 / 24.8 / 15.9 & 18.4 / 21.9 / 13.0 / 8.88 & 19.3 / 26.4 / 10.4 / 3.07 & 25.5 / 32.9 / 23.1 / 14.4 & 41.3 / 69.5 / 39.0 / 12.9 \\
~+ICL & \negcol{24.9} / \negcol{33.1} / \negcol{23.6} / \negcol{14.0} & \poscol{20.5} / \poscol{27.8} / \poscol{16.0} / \poscol{8.88} & \poscol{25.8} / \poscol{39.9} / \poscol{20.2} / \poscol{7.98} & \poscol{31.0} / \poscol{42.1} / \poscol{28.7} / \poscol{18.5} & \poscol{48.0} / \poscol{77.2} / \poscol{50.0} / \poscol{19.6} \\
~+GEPA & \negcol{25.1} / \negcol{34.4} / \negcol{24.2} / \negcol{14.0} & \poscol{26.3} / \poscol{38.5} / \poscol{28.4} / \poscol{12.4} & \poscol{25.1} / \poscol{39.3} / \poscol{21.5} / \poscol{7.98} & \poscol{32.3} / \poscol{39.8} / \poscol{31.0} / \poscol{23.1} & \poscol{47.8} / \poscol{78.4} / \poscol{50.0} / \poscol{16.8} \\
\midrule
\multicolumn{6}{c}{\textit{Open-source VLM (Qwen3-VL-8B)}} \\
\midrule
Zero-shot & 14.1 / 19.7 / 11.5 / 5.73 & 8.34 / 10.7 / 5.92 / 0.59 & 10.6 / 11.7 / 1.23 / 0.00 & 10.6 / 15.7 / 5.09 / 3.70 & 29.4 / 46.0 / 16.0 / 2.00 \\
~+ICL & \poscol{20.7} / \poscol{28.0} / \poscol{19.7} / \poscol{8.92} & \poscol{10.6} / \poscol{14.8} / \poscol{8.28} / \poscol{3.55} & \poscol{16.4} / \poscol{23.3} / \poscol{14.7} / \poscol{2.45} & \poscol{24.3} / \poscol{34.3} / \poscol{18.1} / \poscol{10.6} & \poscol{39.1} / \poscol{65.6} / \poscol{37.6} / \poscol{12.0} \\
~+GEPA & \negcol{12.6} / \negcol{17.2} / \poscol{11.5} / \poscol{6.37} & \poscol{8.46} / \poscol{11.8} / \poscol{7.10} / \poscol{2.96} & \poscol{12.2} / \poscol{17.8} / \poscol{6.13} / \poscol{2.45} & \poscol{29.1} / \poscol{39.4} / \poscol{28.2} / \poscol{14.8} & \poscol{31.1} / \poscol{48.8} / \poscol{22.0} / \poscol{7.20} \\
\rowcolor{gray!12}
~+SFT LoRA & \poscol{14.2} / \poscol{21.0} / \poscol{12.1} / \poscol{7.01} & \negcol{6.46} / \negcol{7.69} / \negcol{2.37} / \poscol{0.59} & \poscol{11.3} / \poscol{19.0} / \poscol{6.75} / \poscol{2.45} & \poscol{24.5} / \poscol{36.1} / \poscol{22.2} / \poscol{9.26} & \poscol{40.3} / \poscol{64.8} / \poscol{40.4} / \poscol{15.6} \\
\rowcolor{gray!12}
~+SFT Full Fine-tuning & \negcol{10.5} / \negcol{16.6} / \negcol{10.2} / \negcol{2.55} & \negcol{5.81} / \negcol{7.69} / \negcol{4.14} / \poscol{2.37} & \negcol{8.93} / \poscol{12.9} / \poscol{7.98} / \poscol{3.07} & \poscol{20.5} / \poscol{27.8} / \poscol{17.1} / \poscol{6.48} & \poscol{45.1} / \poscol{71.2} / \poscol{49.6} / \poscol{17.6} \\
\rowcolor{gray!12}
~+GRPO LoRA
& \poscol{25.2} / \poscol{34.4} / \poscol{23.6} / \poscol{12.7}
& \poscol{12.7} / \poscol{18.3} / \poscol{10.7} / \poscol{5.92}
& \poscol{21.3} / \poscol{33.1} / \poscol{14.7} / \poscol{4.29}
& \poscol{41.0} / \poscol{54.2} / \poscol{44.4} / \poscol{26.4}
& \poscol{50.3} / \poscol{83.6} / \poscol{52.4} / \poscol{18.0} \\
\bottomrule
\end{tabular}}
\end{table*}

\begin{table*}[t]
\centering
\caption{\textbf{Adaptation results on the Spatial Video Grounding (SVG) task.}
Each cell reports $m_{\text{s}}\mathrm{IoU}$ / $s\mathrm{IoU}@0.3$ / $s\mathrm{IoU}@0.5$.
The first row shows the zero-shot baseline; subsequent rows show adaptation using the provided per-domain training subsets. \colorbox{gray!12}{Shaded} rows denote fine-tuning-based adaptation, while unshaded adaptation rows are training-free.
\poscol{Purple} denotes performance improvements or ties and \negcol{red} denotes degradations relative to the zero-shot baseline.}
\label{tab:adaptation_svg_details}
\resizebox{\textwidth}{!}{
\begin{tabular}{l *{5}{c}}
\toprule
\textbf{Models} & \textcolor{AnimalColor}{\textbf{Animal}} & \textcolor{IndustryColor}{\textbf{Industry}} & \textcolor{SportsColor}{\textbf{Sports}} & \textcolor{SurgeryColor}{\textbf{Surgery}} & \textcolor{SecurityColor}{\textbf{Public Security}} \\
\midrule
\multicolumn{6}{c}{\textit{Proprietary VLM (Gemini-3.1-Pro)}} \\
\midrule
Zero-shot & 49.2 / 70.7 / 49.0 & 31.1 / 41.4 / 32.0 & 13.1 / 16.6 / 7.98 & 19.1 / 34.2 / 2.68 & 33.2 / 52.0 / 28.4 \\
~+ICL & \negcol{41.9} / \negcol{60.5} / \negcol{42.7} & \negcol{29.5} / \negcol{39.6} / \poscol{32.5} & \negcol{6.89} / \negcol{7.36} / \negcol{5.52} & \poscol{24.1} / \poscol{34.2} / \poscol{14.1} & \negcol{26.4} / \negcol{41.6} / \negcol{21.2} \\
~+GEPA & \negcol{41.7} / \negcol{60.5} / \negcol{42.0} & \poscol{37.7} / \poscol{51.5} / \poscol{39.6} & \negcol{10.5} / \negcol{12.9} / \poscol{7.98} & \poscol{32.2} / \poscol{45.6} / \poscol{32.2} & \poscol{34.1} / \poscol{53.2} / \poscol{30.0} \\
\midrule
\multicolumn{6}{c}{\textit{Open-source VLM (Qwen3-VL-8B)}} \\
\midrule
Zero-shot & 15.2 / 17.8 / 1.27 & 7.90 / 7.69 / 2.37 & 0.44 / 0.00 / 0.00 & 1.89 / 0.00 / 0.00 & 5.05 / 3.60 / 0.00 \\
~+ICL & \poscol{15.5} / \poscol{20.4} / \poscol{5.73} & \poscol{11.7} / \poscol{15.4} / \poscol{6.51} & \poscol{0.48} / \poscol{0.61} / \poscol{0.00} & \poscol{10.7} / \poscol{13.4} / \poscol{2.01} & \negcol{4.78} / \negcol{3.20} / \poscol{0.40} \\
~+GEPA & \poscol{25.6} / \poscol{35.7} / \poscol{12.7} & \poscol{9.80} / \poscol{11.8} / \poscol{2.37} & \negcol{0.41} / \poscol{0.61} / \poscol{0.00} & \poscol{5.40} / \poscol{0.00} / \poscol{0.00} & \poscol{5.82} / \poscol{4.80} / \poscol{0.80} \\
\rowcolor{gray!12}
~+SFT LoRA & \poscol{18.1} / \poscol{26.1} / \poscol{15.9} & \poscol{13.5} / \poscol{21.3} / \poscol{11.2} & \poscol{0.85} / \poscol{1.23} / \poscol{0.00} & \poscol{11.2} / \poscol{12.8} / \poscol{5.37} & \negcol{2.26} / \negcol{0.40} / \poscol{0.00} \\
\rowcolor{gray!12}
~+SFT Full Fine-tuning & \poscol{22.0} / \poscol{28.7} / \poscol{25.5} & \poscol{16.2} / \poscol{24.3} / \poscol{13.0} & \poscol{3.90} / \poscol{5.52} / \poscol{2.45} & \poscol{22.7} / \poscol{32.9} / \poscol{13.4} & \poscol{13.5} / \poscol{16.0} / \poscol{7.20} \\
\rowcolor{gray!12}
~+GRPO LoRA
& \poscol{23.8} / \poscol{31.2} / \poscol{26.1}
& \poscol{17.3} / \poscol{25.4} / \poscol{14.2}
& \poscol{2.76} / \poscol{3.07} / \poscol{1.23}
& \poscol{22.7} / \poscol{36.2} / \poscol{14.8}
& \negcol{3.05} / \negcol{2.40} / \poscol{0.80} \\
\bottomrule
\end{tabular}}
\end{table*}

\noindent \textbf{Detailed Adaptation Results.}
The inconsistency of training-free adaptation holds across metrics.
For Gemini-3.1-Pro, the gains of GEPA in surgery persist at stricter thresholds, raising $v\mathrm{IoU}@0.5$ from 0.46 to 9.72 and $s\mathrm{IoU}@0.5$ from 2.68 to 32.2.
In contrast, GEPA degrades $m_{\text{v}}\mathrm{IoU}$ of Qwen3-VL-8B in three domains and the temporal IoU of its STVG predictions in four.
Fine-tuning yields SVG gains that also hold at $s\mathrm{IoU}@0.5$, \eg, from 1.27 to 26.1 in animal and from 0.00 to 14.8 in surgery with GRPO LoRA.
However, these gains do not carry over to STVG.
GRPO LoRA raises TVG $m_{\text{t}}\mathrm{IoU}$ in public security from 29.4 to 50.3, yet the temporal IoU of its STVG predictions slightly decreases (27.7 to 26.9), with similar trends in animal and industry.
The gains are also domain-dependent: in public security, both LoRA variants degrade $m_{\text{s}}\mathrm{IoU}$, whereas full fine-tuning improves it from 5.05 to 13.5.
Consequently, $v\mathrm{IoU}@0.5$ of all fine-tuned models remains at most 5.10.

\section{Further Analysis}
\label{app:further_analysis}
\begin{figure*}[t]
\centering
\includegraphics[width=\textwidth]{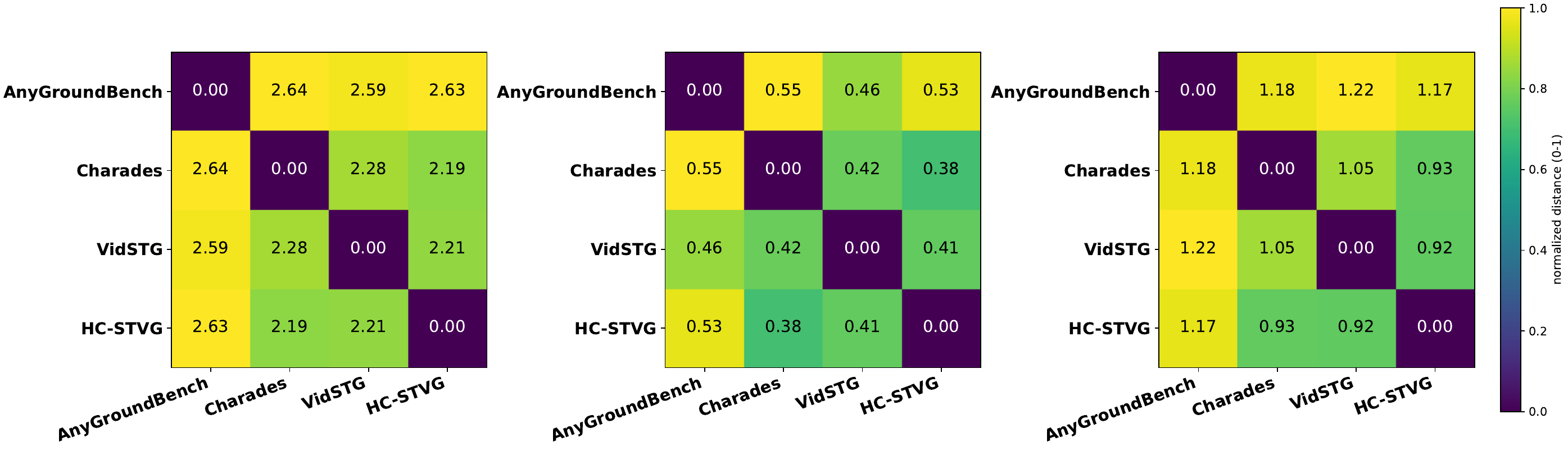}
\caption{\textbf{Dataset distance.}
Pairwise Fr\'echet distances among \paper, Charades-STA, VidSTG, and HC-STVG in three
embedding spaces: joint multimodal (Qwen3-VL-Embedding-8B), visual (InternVideo2), and
text (SentenceBERT). Larger values indicate greater distributional shift.}
\label{fig:dataset_distance}
\end{figure*}

\noindent \textbf{Dataset Distance.}
\label{app:dataset_distance}
We compute Fr\'echet distances between \paper\ and Charades-STA, VidSTG and
HC-STVG in three embedding spaces: joint multimodal (Qwen3-VL-Embedding-8B),
visual (InternVideo2) and text (SentenceBERT). Figure~\ref{fig:dataset_distance} reports all pairwise
distances. The three existing benchmarks
differ in content yet sit close together, so we use the average distance among
them ($2.23$ multimodal, $0.40$ visual, $0.97$ text) as a reference scale for
how far apart daily-life grounding benchmarks normally lie. The average
distance from \paper\ to an existing benchmark is $18\%$, $27\%$ and $23\%$
larger than this reference in the three spaces respectively, and this is not an
effect of averaging: even the existing benchmark closest to \paper\ is farther
away than any two existing benchmarks are from each other ($2.59$ vs. $2.28$,
$0.46$ vs. $0.42$, $1.17$ vs. $1.05$). \paper\ thus lies outside the region the
existing benchmarks jointly occupy, in what is seen, in what is asked, and in
how the two are paired: it covers rare domains that existing grounding datasets
do not provide.

\noindent \textbf{Parse Error Analysis.} To disambiguate whether near-zero scores reflect genuine grounding failure or output-parsing breakdown, we audited every model output across the three evaluation tasks and categorized it as successfully parsed, empty output (\eg "[]", ""), refusal (\eg "I'm unable to determine specific actions or behaviors of birds in images.", "I'm unable to perform this task."), invalid structured output, or other parse error (\eg "The video contains 34 frames."). Parsing succeeded for the large majority of outputs (STVG 89.4\%, TVG 97.9\%, SVG 86.8\%), with TVG showing a negligible combined failure rate of 2.1\% (empty 0.2\%, refusal 0.1\%, invalid structured 0.0\%, other 1.8\%), while STVG and SVG show higher, still-minority failure rates of 10.6\% (empty 5.6\%, refusal 1.3\%, invalid structured 2.5\%, other 1.1\%) and 13.2\% (empty 5.4\%, refusal 1.8\%, invalid structured 4.4\%, other 1.5\%) respectively, with invalid structured output nearly doubling from STVG to SVG (2.5\%$\to$4.4\%) consistent with the added complexity of emitting well-formed spatial coordinates. These results show that the large majority of outputs across all tasks are correctly parsed, confirming that near-zero scores are not an artifact of broken output but instead reflect the models' genuine grounding ability.

\begin{figure*}[t]
\centering
\includegraphics[width=\textwidth]{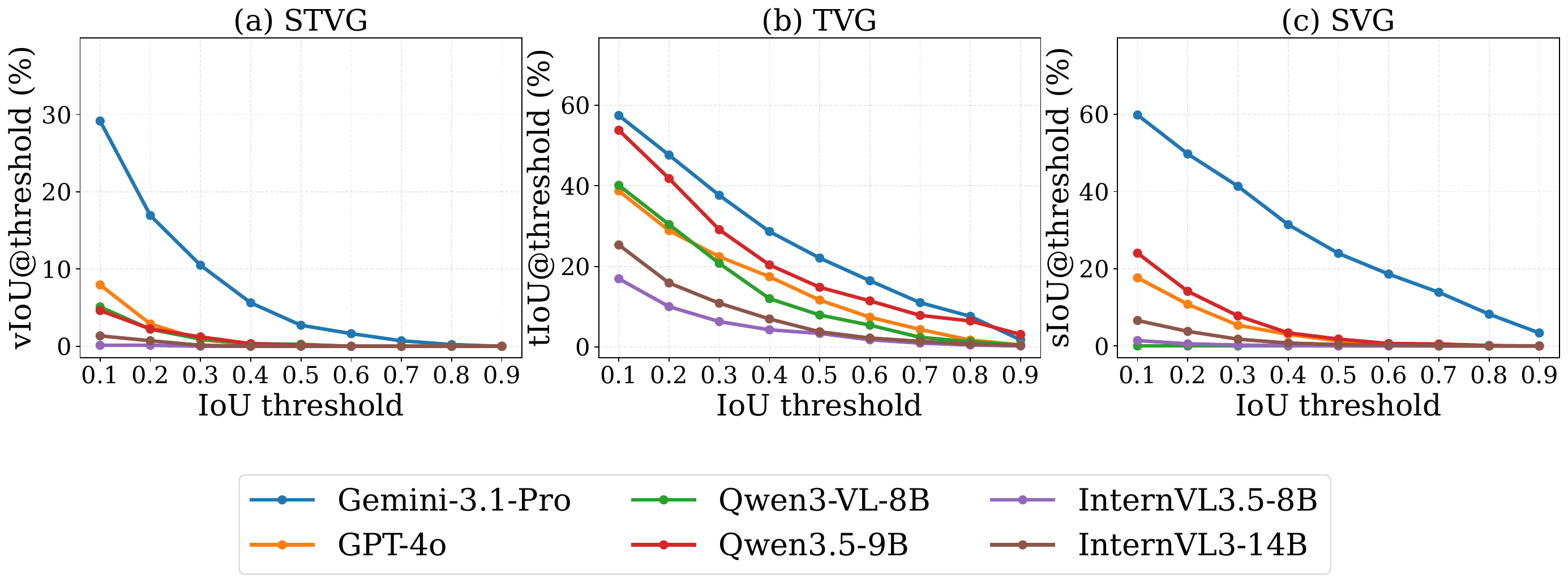}
\caption{\textbf{Model-wise threshold sensitivity across STVG, TVG, and SVG.}
Each curve reports the percentage of examples above each IoU threshold.
STVG and SVG use $v\mathrm{IoU}$ and $s\mathrm{IoU}$ thresholds, respectively, while TVG uses temporal IoU thresholds.
STVG accuracy drops sharply as the threshold increases, showing that coarse success at permissive thresholds rarely becomes precise tube grounding.}
\label{fig:threshold_sensitivity_models}
\end{figure*}

\noindent \textbf{Threshold Sensitivity Analysis.}\label{app:threshold_sensitivity} \autoref{fig:threshold_sensitivity_models} shows how representative models behave as the IoU threshold is varied from $0.1$ to $0.9$.
Across models, STVG accuracy decreases sharply as the threshold becomes stricter.
Gemini-3.1-Pro remains the strongest model over most thresholds, but even its STVG curve falls rapidly from the permissive regime to nearly zero at high thresholds.
TVG degrades more gradually than STVG and SVG, which reinforces the main-table observation that temporal localization is comparatively more robust than precise tube grounding.

\begin{figure*}[t]
\centering
\includegraphics[width=\textwidth]{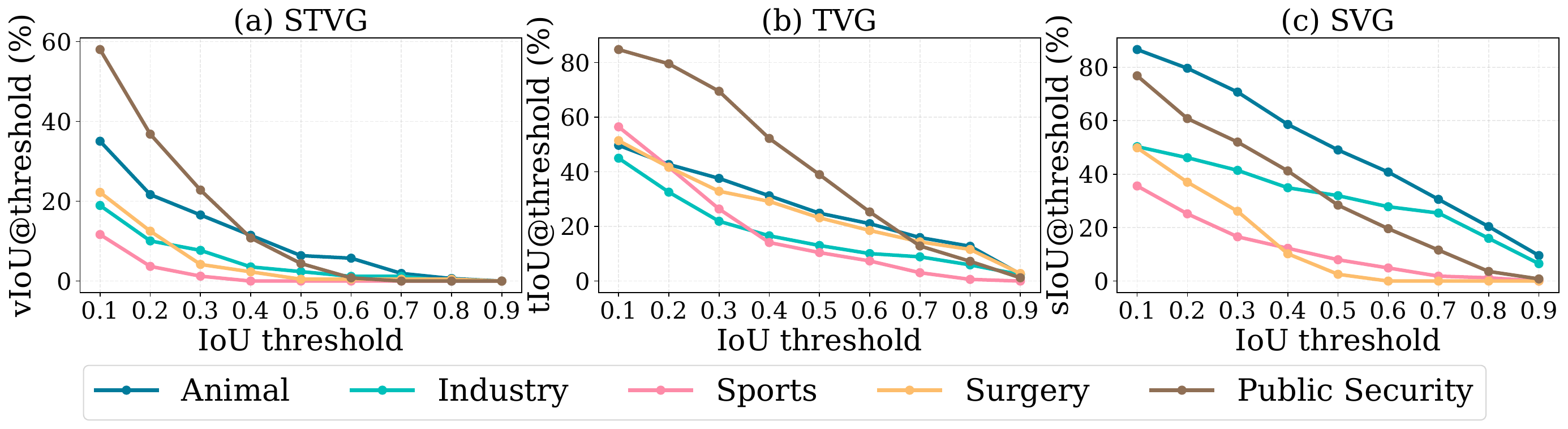}
\caption{\textbf{Domain-wise threshold sensitivity for Gemini-3.1-Pro.}
Each curve aggregates results within one domain across STVG, TVG, and SVG.
Public Security and Animal remain relatively strong at loose thresholds, while Sports and Surgery are consistently difficult.
Across all domains, STVG degrades much more sharply than TVG, confirming that precise spatio-temporal overlap is the central failure mode.}
\label{fig:threshold_sensitivity_domains}
\end{figure*}

\autoref{fig:threshold_sensitivity_domains} presents the same analysis aggregated by domain for Gemini-3.1-Pro.
Animal and Public Security remain the strongest domains under loose thresholds, whereas Sports and Surgery are consistently difficult.
However, all domains show a steep decline in STVG as the threshold increases.
This indicates that domain-specific STVG failures are not only ranking artifacts at a single threshold; they reflect a broader lack of precise spatio-temporal overlap.

\begin{figure*}[tb]
\centering
\includegraphics[width=\textwidth]{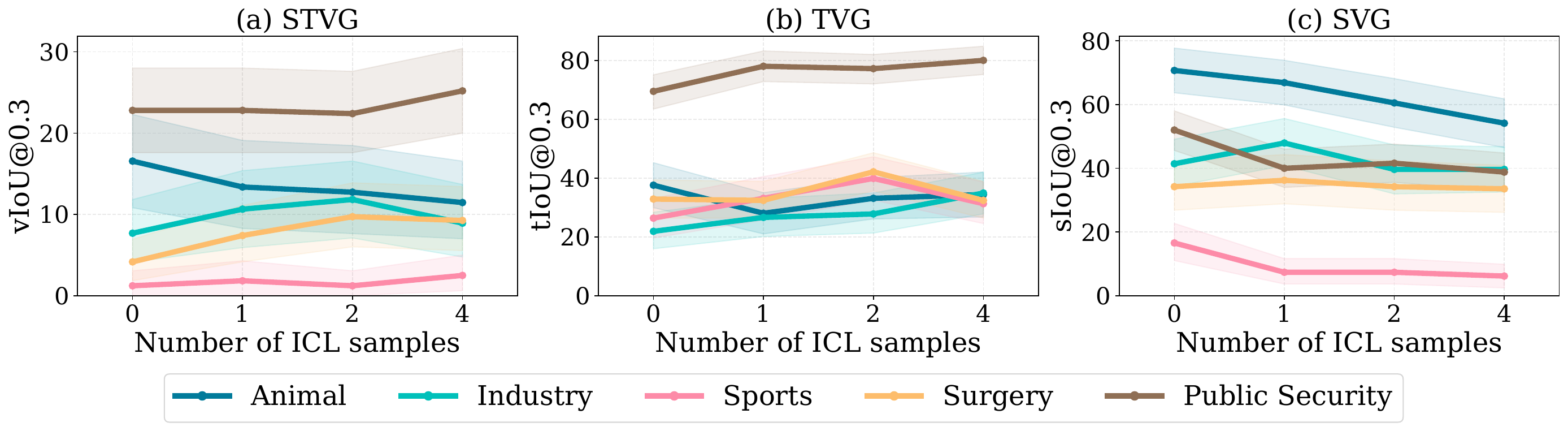}
\caption{\textbf{Effect of the number of in-context demonstrations.}
Performance on (a) STVG, (b) TVG, and (c) SVG as the number of retrieved demonstrations varies from $0$ to $4$. Error bars denote bootstrap $95\%$ confidence intervals computed per domain.
All results use Gemini-3.1-Pro.}
\label{fig:icl_n_shot_ablation}
\end{figure*}

\noindent \textbf{Impact of the Number of In-Context Demonstrations.}
To further analyze the limited and inconsistent ICL gains in the main results, we vary the number of retrieved demonstrations from 0 to 4 for Gemini-3.1-Pro, as shown in \autoref{fig:icl_n_shot_ablation}.
Increasing the number of demonstrations primarily benefits TVG, whereas the average SVG score consistently drops from zero-shot to 4-shot.
Meanwhile, the gains in STVG remain modest, increasing only from $10.5$ (zero-shot) to $11.6$ (2-shot) and $11.5$ (4-shot), suggesting that simply adding more demonstrations is insufficient for adapting specialized-domain STVG.

\noindent \textbf{Effectiveness of In-Context Selection Strategies.}
To clarify whether the instability of ICL depends not only on the number of
demonstrations but also on how examples are retrieved from the training pool, we
compare random selection (no similarity) against text-only, video-only, and text+video
retrieval for Gemini-3.1-Pro, as summarized in
\autoref{tab:icl_selection_strategies_main_style}.
The margins between strategies are small: averaged across
domains, text+video retrieval is highest on STVG ($11.6$ versus $11.2$ for random),
on TVG ($44.0$ versus $43.3$ for text-only), and on SVG ($36.7$ versus $35.3$ for
random). Notably, every retrieval strategy falls below the
zero-shot SVG average ($43.0$).
These results suggest that no retrieval signal reliably improves spatial grounding, and
that the instability of ICL is not resolved by better demonstration selection.

\begin{table*}[tb]
\centering
\caption{\textbf{Comparison of ICL sample selection strategies for Gemini-3.1-Pro on STVG, TVG, and SVG.}
Each cell reports STVG / TVG / SVG, where STVG uses $v\mathrm{IoU}@0.3$, TVG uses $t\mathrm{IoU}@0.3$, and SVG uses $s\mathrm{IoU}@0.3$ (\%).
Rows compare the zero-shot baseline with four 2-shot retrieval strategies (\colorbox{gray!12}{shaded}): random, text-only, video-only, and text+video similarity over the per-domain training pool; text+video corresponds to the +ICL setting in \autoref{tab:VLM_adaptation_results}.
\textbf{Bold} marks the best value within each task--domain pair.}
\label{tab:icl_selection_strategies_main_style}
\resizebox{\textwidth}{!}{%
\begin{tabular}{l *{5}{c}}
\toprule
\textbf{Selection Strategy} & \textcolor{AnimalColor}{\textbf{Animal}} & \textcolor{IndustryColor}{\textbf{Industry}} & \textcolor{SportsColor}{\textbf{Sports}} & \textcolor{SurgeryColor}{\textbf{Surgery}} & \textcolor{SecurityColor}{\textbf{Public Security}} \\
\midrule
Zero-shot       & 16.6 / \textbf{37.6} / \textbf{70.7} & 7.69 / 21.9 / \textbf{41.4} & 1.23 / 26.4 / \textbf{16.6} & 4.17 / 32.9 / \textbf{34.2} & 22.8 / 69.5 / \textbf{52.0} \\
\rowcolor{gray!12}
Random       & \textbf{20.4} / \textbf{37.6} / 68.2 & 8.28 / 27.8 / 36.7 & \textbf{2.45} / 27.0 / 7.39 & 4.63 / 39.8 / 23.1 & 20.4 / 74.4 / 41.2 \\
\rowcolor{gray!12}
Text only    & 14.6 / 36.3 / 54.8 & 10.1 / \textbf{33.1} / 37.9 & 0.61 / 35.0 / 6.13 & 8.33 / 34.7 / 27.4 & 22.0 / 77.2 / 40.0 \\
\rowcolor{gray!12}
Video only   & 15.3 / 33.1 / 63.1 & 8.77 / 25.4 / 37.3 & 0.61 / 33.7 / 5.53 & 4.16 / 31.9 / 16.4 & \textbf{25.6} / \textbf{78.8} / 38.4 \\
\rowcolor{gray!12}
Text + Video & 12.7 / 33.1 / 60.5 & \textbf{11.8} / 27.8 / 39.6 & 1.23 / \textbf{39.9} / 7.36 & \textbf{9.72} / \textbf{42.1} / \textbf{34.2} & 22.4 / 77.2 / 41.6 \\
\bottomrule
\end{tabular}}
\end{table*}

\begin{figure*}[tb]
\centering
\includegraphics[width=\textwidth]{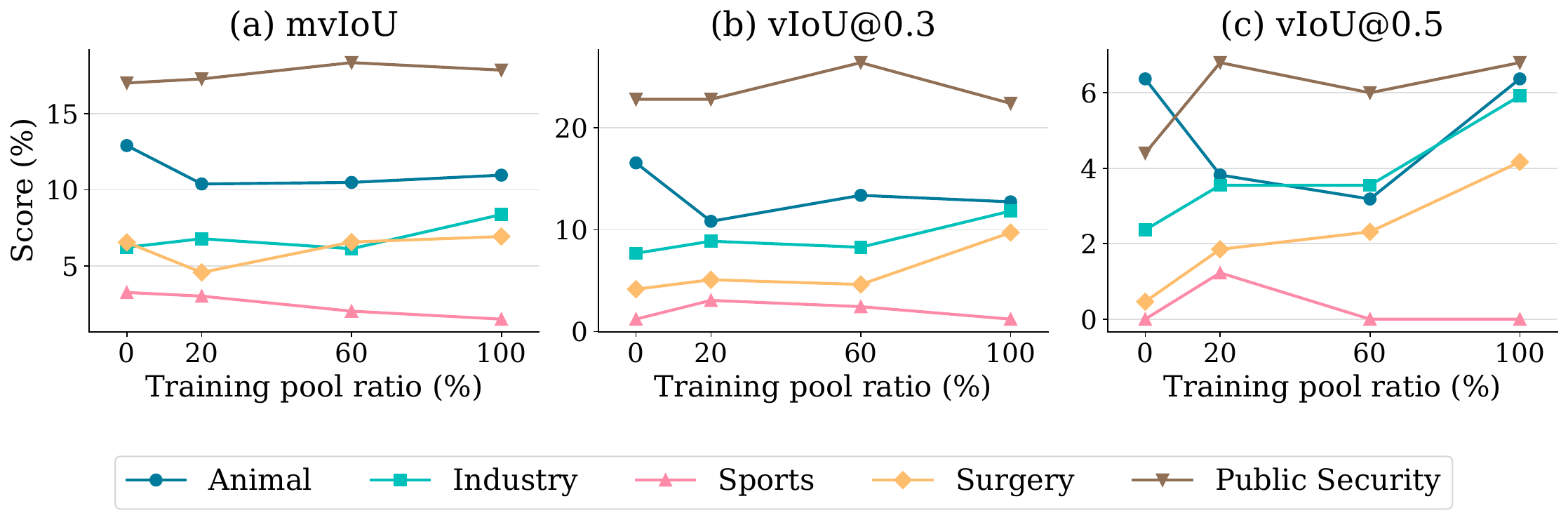}
\caption{\textbf{Effect of the training pool size for ICL retrieval.}
STVG performance of Gemini-3.1-Pro with ICL when the retrieval pool is restricted to a given ratio of each per-domain training subset, measured by (a) $m_{\text{v}}\mathrm{IoU}$, (b) $v\mathrm{IoU}@0.3$, and (c) $v\mathrm{IoU}@0.5$.
A ratio of 0 denotes the zero-shot baseline, and 100\% corresponds to the setting in the main paper.}
\label{fig:icl_train_pool_ablation}
\end{figure*}

\noindent \textbf{Effect of Training Data Size on ICL.}
\label{app:icl_pool}
We analyze how the amount of training data available for ICL affects STVG performance, using Gemini-3.1-Pro and restricting the retrieval pool to $N\%$ of each per-domain training subset, with $N \in \{0, 20, 60, 100\}$.
Here, $N=0$ corresponds to the zero-shot baseline and $N=100$ to the setting in the main paper. As shown in \autoref{fig:icl_train_pool_ablation}, no pool size yields a clear improvement in STVG: $m_{\text{v}}\mathrm{IoU}$ and $v\mathrm{IoU}@0.3$ stay close to the zero-shot baseline, and $v\mathrm{IoU}@0.5$ improves only with the full pool in three of five domains.
The effects also vary across domains.
Industry and surgery improve mainly with the full pool, whereas animal degrades in $m_{\text{v}}\mathrm{IoU}$ and $v\mathrm{IoU}@0.3$ for all pool sizes, and sports degrades monotonically in $m_{\text{v}}\mathrm{IoU}$ as the pool grows.
These results indicate that more training data alone does not reliably improve STVG through ICL, consistent with the limited and inconsistent gains of training-free adaptation in the main paper.

\section{Limitations and Future Work}
\label{sec:limitations}

\noindent \textbf{Task Scope.}
\paper\ centers on single-query grounding: each query is generated from a single target object tube and its bounding boxes.
This design supports diagnosis of spatial and temporal localization failures, but not open-ended expert QA, causal reasoning, or multi-object relational understanding.
Extensions to multi-query, multi-object, and instruction-following settings would cover other aspects of specialized video workflows.

\noindent \textbf{Video Length.}
\paper\ uses short clips to support dense tube annotation and broad MLLM evaluation.
The average video is $47.13$s and the average target segment is $7.51$s, so the benchmark does not test long-form video search.
Extensions to longer videos would require coarse-to-fine retrieval, adaptive frame selection, and memory over sparse events.

%% file: iclr2027_conference.bib
@article{meng2025openo3,
  title={{Open-o3 Video: Grounded Video Reasoning with Explicit Spatio-Temporal Evidence}}, 
  author={Meng, Jiahao and Li, Xiangtai and Wang, Haochen and Tan, Yue and Zhang, Tao and Kong, Lingdong and Tong, Yunhai and Wang, Anran and Teng, Zhiyang and Wang, Yujing and Wang, Zhuochen},
  journal={arXiv preprint arXiv:2510.20579},
  year={2025}
}

@article{cheng2025vstar,
  title={{V-STaR: Benchmarking Video-LLMs on Video Spatio-Temporal Reasoning}},
  author={Cheng, Zixu and Hu, Jian and Liu, Ziquan and Si, Chenyang and Li, Wei and Gong, Shaogang},
  journal={arXiv preprint arXiv:2503.11495},
  year={2025}
}

@inproceedings{
Zeng2025futuresightdrive,
title={{FutureSightDrive: Thinking Visually with Spatio-Temporal CoT for Autonomous Driving}},
author={Shuang Zeng and Xinyuan Chang and Mengwei Xie and Xinran Liu and Yifan Bai and Zheng Pan and Mu Xu and Xing Wei},
booktitle={NeurIPS},
year={2025},
}

@inproceedings{gu2024context,
  title={{Context-Guided Spatio-Temporal Video Grounding}},
  author={Gu, Xin and Fan, Heng and Huang, Yan and Luo, Tiejian and Zhang, Libo},
  booktitle={CVPR},
  year={2024}
}

@article {Segalin2021MARS,
title = {{The Mouse Action Recognition System (MARS) software pipeline for automated analysis of social behaviors in mice}},
author = {Segalin, Cristina and Williams, Jalani and Karigo, Tomomi and Hui, May and Zelikowsky, Moriel and Sun, Jennifer J and Perona, Pietro and Anderson, David J and Kennedy, Ann},
journal = {eLife},
volume = 10,
year = 2021,
pages = {e63720},
}

@article{brown2020language,
  title={{Language Models are Few-Shot Learners
}},
  author={Brown, Tom and Mann, Benjamin and Ryder, Nick and Subbiah, Melanie and Kaplan, Jared D and Dhariwal, Prafulla and Neelakantan, Arvind and Shyam, Pranav and Sastry, Girish and Askell, Amanda and others},
  journal={NeurIPS},
  year={2020}
}

@inproceedings{
xie2022an,
title={{An Explanation of In-context Learning as Implicit Bayesian Inference}},
author={Sang Michael Xie and Aditi Raghunathan and Percy Liang and Tengyu Ma},
booktitle={ICLR},
year={2022},

}

@inproceedings{
rubin2022learning,
title={{Learning To Retrieve Prompts for In-Context Learning}},
author={Ohad Rubin and Jonathan Herzig and Jonathan Berant},
booktitle={NAACL},
year={2022},
}

@InProceedings{Kim2025VideoICL,
    author    = {Kim, Kangsan and Park, Geon and Lee, Youngwan and Yeo, Woongyeong and Hwang, Sung Ju},
    title     ={ {VideoICL: Confidence-based Iterative In-context Learning for Out-of-Distribution Video Understanding}},
    booktitle = {CVPR},
    year      = {2025},
}

@article{Fujii2026Viola,
  title={{VIOLA: Towards Video In-Context Learning with Minimal Annotations}},
  author={Fujii, Ryo and Saito, Hideo and Hachiuma, Ryo},
  journal={arXiv preprint arXiv:2601.15549},
  year={2026}
}

@InProceedings{Wasim2024VideoGrounding-DINO,
    author    = {Wasim, Syed Talal and Naseer, Muzammal and Khan, Salman and Yang, Ming-Hsuan and Khan, Fahad Shahbaz},
    title     = {{VideoGrounding-DINO: Towards Open-Vocabulary Spatio-Temporal Video Grounding}},
    booktitle = {CVPR},
    year      = {2024},
}

@InProceedings{Su2021STVGBert,
    author    = {Su, Rui and Yu, Qian and Xu, Dong},
    title     = {{STVGBert: A Visual-Linguistic Transformer Based Framework for Spatio-Temporal Video Grounding}},
    booktitle = {ICCV},
    year      = {2021},
}

@inproceedings{Gu2026ARTSTVG,
  title     = {{Towards Long-Form Spatio-Temporal Video Grounding}},
  author    = {Gu, Xin and Fan, Bing and Yao, Jiali and Zhang, Zhipeng and Huang, Yan and Han, Cheng and Fan, Heng and Zhang, Libo},
  booktitle = {ECCV},
  year      = {2026}
}

@inproceedings{
Gu2025Knowing,
title={{Knowing Your Target: Target-Aware Transformer Makes Better Spatio-Temporal Video Grounding}},
author={Xin Gu and Yaojie Shen and Chenxi Luo and Tiejian Luo and Yan Huang and Yuewei Lin and Heng Fan and Libo Zhang},
booktitle={ICLR},
year={2025},
}

@article{Qwen3-VL,
      title={{Qwen3-VL Technical Report}}, 
      author={Shuai Bai and Yuxuan Cai and Ruizhe Chen and Keqin Chen and Xionghui Chen and Zesen Cheng and Lianghao Deng and Wei Ding and Chang Gao and Chunjiang Ge and Wenbin Ge and Zhifang Guo and Qidong Huang and Jie Huang and Fei Huang and Binyuan Hui and Shutong Jiang and Zhaohai Li and Mingsheng Li and Mei Li and Kaixin Li and Zicheng Lin and Junyang Lin and Xuejing Liu and Jiawei Liu and Chenglong Liu and Yang Liu and Dayiheng Liu and Shixuan Liu and Dunjie Lu and Ruilin Luo and Chenxu Lv and Rui Men and Lingchen Meng and Xuancheng Ren and Xingzhang Ren and Sibo Song and Yuchong Sun and Jun Tang and Jianhong Tu and Jianqiang Wan and Peng Wang and Pengfei Wang and Qiuyue Wang and Yuxuan Wang and Tianbao Xie and Yiheng Xu and Haiyang Xu and Jin Xu and Zhibo Yang and Mingkun Yang and Jianxin Yang and An Yang and Bowen Yu and Fei Zhang and Hang Zhang and Xi Zhang and Bo Zheng and Humen Zhong and Jingren Zhou and Fan Zhou and Jing Zhou and Yuanzhi Zhu and Ke Zhu},
	  journal={arXiv preprint arXiv:2511.21631},
      year={2025}
}

@misc{qwen3.5,
    title  = {{{Qwen3.5}: Towards Native Multimodal Agents}},
    author = {{Qwen Team}},
    month  = {February},
    year   = {2026},
    url    = {https://qwen.ai/blog?id=qwen3.5}
}

@misc{qwen38,
    title = {{Qwen3.8-Max}: A New Bar for Coding and Cowork},
    url = {https://qwen.ai/blog?id=qwen3.8},
    author = {{Qwen Team}},
    month = {August},
    year = {2026}
}

@article{Vidi2025vidi2,
  title={{Vidi2: Large Multimodal Models for Video Understanding and Creation}},
  author={{Vidi Team} and Celong Liu and Chia-Wen Kuo and Chuang Huang and Dawei Du and Fan Chen and Guang Chen and Haoji Zhang and Haojun Zhao and Lingxi Zhang and Lu Guo and Lusha Li and Longyin Wen and Qihang Fan and Qingyu Chen and Rachel Deng and Sijie Zhu and Stuart Siew and Tong Jin and Weiyan Tao and Wen Zhong and Xiaohui Shen and Xin Gu and Zhenfang Chen and Zuhua Lin},
  journal={arXiv preprint arXiv:2511.19529},
  year={2025}
}

@article{zhu2025internvl3,
  title={{InternVL3: Exploring Advanced Training and Test-Time Recipes for Open-Source Multimodal Models}},
  author={Zhu, Jinguo and Wang, Weiyun and Chen, Zhe and Liu, Zhaoyang and Ye, Shenglong and Gu, Lixin and Tian, Hao and Duan, Yuchen and Su, Weijie and Shao, Jie and others},
  journal={arXiv preprint arXiv:2504.10479},
  year={2025}
}

@article{wang2025internvl3_5,
  title={{InternVL3.5: Advancing Open-Source Multimodal Models in Versatility, Reasoning, and Efficiency}},
  author={Wang, Weiyun and Gao, Zhangwei and Gu, Lixin and Pu, Hengjun and Cui, Long and Wei, Xingguang and Liu, Zhaoyang and Jing, Linglin and Ye, Shenglong and Shao, Jie and others},
  journal={arXiv preprint arXiv:2508.18265},
  year={2025}
}

@misc{google2025gemini3flash,
  title= {{Gemini 3 Flash Model Card}},
  author = {{Google DeepMind}},
  institution = {Google DeepMind},
  year = {2025},
  howpublished = {\url{https://storage.googleapis.com/deepmind-media/Model-Cards/Gemini-3-Flash-Model-Card.pdf}}
}

@article{comanici2025gemini2_5,
  title={{Gemini 2.5: Pushing the Frontier with Advanced Reasoning, Multimodality, Long Context, and Next Generation Agentic Capabilities}},
  author={Comanici, Gheorghe and Bieber, Eric and Schaekermann, Mike and Pasupat, Ice and Sachdeva, Noveen and Dhillon, Inderjit and Blistein, Marcel and Ram, Ori and Zhang, Dan and Rosen, Evan and others},
  journal={arXiv preprint arXiv:2507.06261},
  year={2025}
}

@article{gemmateam2026gemma4,
      title={{Gemma 4 Technical Report}},
      author={{Gemma Team}},
      journal={arXiv preprint arXiv:2607.02770},
      year={2026},
}

@article{achiam2023gpt,
  title={{GPT-4 Technical Report}},
  author={Achiam, Josh and Adler, Steven and Agarwal, Sandhini and Ahmad, Lama and Akkaya, Ilge and Aleman, Florencia Leoni and Almeida, Diogo and Altenschmidt, Janko and Altman, Sam and Anadkat, Shyamal and others},
  journal={arXiv preprint arXiv:2303.08774},
  year={2023}
}

@article{singh2025gpt5,
  title={{OpenAI GPT-5 System Card}},
  author={Singh, Aaditya and Fry, Adam and Perelman, Adam and Tart, Adam and Ganesh, Adi and El-Kishky, Ahmed and McLaughlin, Aidan and Low, Aiden and Ostrow, AJ and Ananthram, Akhila and others},
  journal={arXiv preprint arXiv:2601.03267},
  year={2025}
}

@misc{openai2026gpt5_4,
  title        = {{GPT-5.4 Thinking System Card}},
  author       = {{OpenAI}},
  year         = {2026},
  month        = {March},
  howpublished = {\url{https://deploymentsafety.openai.com/gpt-5-4-thinking/gpt-5-4-thinking.pdf}}
}

@misc{openai2026gpt5_6,
  title        = {{GPT-5.6 System Card}},
  author       = {{OpenAI}},
  year         = {2026},
  month        = {June},
  howpublished = {\url{https://deploymentsafety.openai.com/gpt-5-6}}
}

@misc{openai2026gpt6astra,
  title        = {{GPT-6 Astra System Card}},
  author       = {{OpenAI}},
  year         = {2026},
  howpublished = {\url{https://deploymentsafety.openai.com/gpt-6-astra}}
}

@misc{google2026gemini3_1pro,
  title        = {{Gemini 3.1 Pro Model Card}},
  author       = {{Google DeepMind}},
  year         = {2026},
  month        = {February},
  howpublished = {\url{https://deepmind.google/models/model-cards/gemini-3-1-pro/}}
}

@misc{google2026gemini3_5flash,
  title        = {{Gemini 3.5 Flash Model Card}},
  author       = {{Google DeepMind}},
  year         = {2026},
  howpublished = {\url{https://storage.googleapis.com/deepmind-media/Model-Cards/Gemini-3-5-Flash-Model-Card.pdf}}
}

@misc{google2026gemini3_8flash,
  title        = {{Gemini 3.8 Flash Model Card}},
  author       = {{Google DeepMind}},
  year         = {2026},
  month        = {September},
  howpublished = {\url{https://storage.googleapis.com/deepmind-media/Model-Cards/Gemini-3-8-Flash-Model-Card.pdf}}
}

@inproceedings{
chen2025eagle2_5,
title={{Eagle 2.5: Boosting Long-Context Post-Training for Frontier Vision-Language Models}},
author={Guo Chen and Zhiqi Li and Shihao Wang and Jindong Jiang and Yicheng Liu and Lidong Lu and De-An Huang and Wonmin Byeon and Matthieu Le and Max Ehrlich and Tong Lu and Limin Wang and Bryan Catanzaro and Jan Kautz and Andrew Tao and Zhiding Yu and Guilin Liu},
booktitle={NeurIPS},
year={2026},

}

@article{Vidi2026vidi25,
  title={{Vidi2.5: Large Multimodal Models for Video Understanding and Creation}},
  author={Vidi Team and Chia-Wen Kuo and Chuang Huang and Dawei Du and 
          Fan Chen and Fanding Lei and Feng Gao and Guang Chen and 
          Haoji Zhang and Haojun Zhao and Jin Liu and Jingjing Zhuge and 
          Lili Fang and Lingxi Zhang and Longyin Wen and Lu Guo and 
          Lu Xu and Lusha Li and Qihang Fan and Rachel Deng and 
          Shaobo Fang and Shu Zhang and Sijie Zhu and Stuart Siew and 
          Weiyan Tao and Wen Zhong and Xiaohui Shen and Xin Gu and 
          Ye Yuan and Yicheng He and Yiming Cui and Zhenfang Chen and 
          Zhihua Wu and Zuhua Lin},
  journal={arXiv preprint arXiv:2511.19529},
  year={2026}
}

@inproceedings{Li2024groundinggpt,
    title = {{{G}rounding{GPT}: Language Enhanced Multi-modal Grounding Model}},
    author = "Li, Zhaowei  and
      Xu, Qi  and
      Zhang, Dong  and
      Song, Hang  and
      Cai, YiQing  and
      Qi, Qi  and
      Zhou, Ran  and
      Pan, Junting  and
      Li, Zefeng  and
      Tu, Vu  and
      Huang, Zhida  and
      Wang, Tao",
    booktitle = {ACL},
    year = "2024",
}

@article{Wang2026SpaceVLLM,
title={{SpaceVLLM: Endowing Multimodal Large Language Model with Spatio-Temporal Video Grounding Capability}},
author={Wang, Jiankang and Zhang, Zhihan and Liu, Zhihang and Li, Yang and Ge, Jiannan and Xie, Hongtao and Zhang, Yongdong},
journal={AAAI},
year={2026},
}

@inproceedings{
Yang2025Unleashing,
title={{Unleashing the Potential of Multimodal {LLM}s for Zero-Shot Spatio-Temporal Video Grounding}},
author={Zaiquan Yang and Yuhao LIU and Gerhard Petrus Hancke and Rynson W. H. Lau},
booktitle={NeurIPS},
year={2025},
}

@article{gu2025thinking,
  title={{Thinking With Bounding Boxes: Enhancing Spatio-Temporal Video Grounding via Reinforcement Fine-Tuning}},
  author={Gu, Xin and Zhang, Haoji and Fan, Qihang and Niu, Jingxuan and Zhang, Zhipeng and Zhang, Libo and Chen, Guang and Chen, Fan and Wen, Longyin and Zhu, Sijie},
  journal={arXiv preprint arXiv:2511.21375},
  year={2025}
}

@article{ahmad2025videomolmo,
  title={{VideoMolmo: Spatio-Temporal Grounding Meets Pointing}},
  author={Ahmad, Ghazi Shazan and Heakl, Ahmed and Gani, Hanan and Shaker, Abdelrahman and Shen, Zhiqiang and Khan, Fahad Shahbaz and Khan, Salman},
  journal={arXiv preprint arXiv:2506.05336},
  year={2025}
}

@inproceedings{
zhang2026stvgr,
title={{{STVG}-R1: Incentivizing Instance-Level Reasoning and Grounding in Videos via Reinforcement Learning}},
author={Xiaowen Zhang and Zhi Gao and Licheng Jiao and Lingling Li and Qing Li},
booktitle={ICLR},
year={2026},
}

@INPROCEEDINGS{Gao2018Charades-STA,
  author={Gao, Jiyang and Sun, Chen and Yang, Zhenheng and Nevatia, Ram},
  booktitle={ICCV}, 
  title={{TALL: Temporal Activity Localization via Language Query}}, 
  year={2017}}

@inproceedings{Zhou2018YouCook2,
author = {Zhou, Luowei and Xu, Chenliang and Corso, Jason J.},
title = {{Towards Automatic Learning of
Procedures from Web Instructional Videos}},
year = {2018},
booktitle = {AAAI},
}

@article{Regneri2013TACoS,
author = {Regneri, Michaela and Rohrbach, Marcus and Wetzel, Dominikus and Thater, Stefan and Schiele, Bernt and Pinkal, Manfred},
title = {{Grounding Action Descriptions in Videos}},
year = {2013},
journal = {TACL},
}

@inproceedings{Zhan2025UAV-SVG,
 author = {Zhan, Yang and Yuan, Yuan},
 booktitle = {NeurIPS},
 title = {{Where Does It Exist from the Low-Altitude: Spatial Aerial Video Grounding}},
 year = {2025}
}

@InProceedings{Zhang2020VidSTG,
author = {Zhang, Zhu and Zhao, Zhou and Zhao, Yang and Wang, Qi and Liu, Huasheng and Gao, Lianli},
title = {{Where Does It Exist: Spatio-Temporal Video Grounding for Multi-Form Sentences}},
booktitle = {CVPR},
year = {2020}
}

@ARTICLE{Tang2022HC-STVGv2,
  author={Tang, Zongheng and Liao, Yue and Liu, Si and Li, Guanbin and Jin, Xiaojie and Jiang, Hongxu and Yu, Qian and Xu, Dong},
  journal={TCSVT}, 
  title={{Human-Centric Spatio-Temporal Video Grounding With Visual Transformers}}, 
  year={2022},}

@InProceedings{Liang2025EgoMask,
    author    = {Liang, Shuo and Zhong, Yiwu and Hu, Zi-Yuan and Tao, Yeyao and Wang, Liwei},
    title     = {{Fine-grained Spatiotemporal Grounding on Egocentric Videos}},
    booktitle = {ICCV},
    year      = {2025},
}

@InProceedings{gao2025omniground,
    title={{OmniGround: A Comprehensive Spatio-Temporal Grounding Benchmark for Real-World Complex Scenarios}},
  author={Gao, Hong and Wu, Jingyu and Xu, Xiangkai and Xie, Kangni and Zhang, Yunchen and Zhong, Bin and Gao, Xurui and Zhang, Min-Ling},
    booktitle = {CVPR},
    year      = {2026},
}

@article{xu2025tog,
  title={{ToG-Bench: Task-Oriented Spatio-Temporal Grounding in Egocentric Videos}},
  author={Xu, Qi'ao and Qian, Tianwen and Fu, Yuqian and Li, Kailing and Jiao, Yang and Zhang, Jiacheng and Wang, Xiaoling and He, Liang},
  journal={arXiv preprint arXiv:2512.03666},
  year={2025}
}

@inproceedings{
zhou2025tumtraf,
title={{{TUMT}raf Video{QA}: Dataset and Benchmark for Unified Spatio-Temporal Video Understanding in Traffic Scenes}},
author={Xingcheng Zhou and Konstantinos Larintzakis and Hao Guo and Walter Zimmer and Mingyu Liu and Hu Cao and Jiajie Zhang and Venkatnarayanan Lakshminarasimhan and Leah Strand and Alois Knoll},
booktitle={ICML},
year={2025},
}

@inproceedings{
cao2025when,
title={{When One Moment Isn't Enough: Multi-Moment Retrieval with Cross-Moment Interactions}},
author={Zhuo Cao and Heming Du and Bingqing Zhang and Xin Yu and Xue Li and Sen Wang},
booktitle={NeurIPS},
year={2025},
}

@inproceedings{Ding2023MeViS,
  title={{{MeViS}: A Large-scale Benchmark for Video Segmentation with Motion Expressions}},
  author={Ding, Henghui and Liu, Chang and He, Shuting and Jiang, Xudong and Loy, Chen Change},
  booktitle={ICCV},
  year={2023}
}

@inproceedings{Seo2020URVOS,
author = {Seo, Seonguk and Lee, Joon-Young and Han, Bohyung},
title = {{URVOS: Unified Referring Video Object Segmentation Network with a Large-Scale Benchmark}},
year = {2020},
booktitle = {ECCV},
}

@INPROCEEDINGS{Gu2018AVA,
  author={Gu, Chunhui and Sun, Chen and Ross, David A. and Vondrick, Carl and Pantofaru, Caroline and Li, Yeqing and Vijayanarasimhan, Sudheendra and Toderici, George and Ricco, Susanna and Sukthankar, Rahul and Schmid, Cordelia and Malik, Jitendra},
  booktitle={CVPR}, 
  title={{AVA: A Video Dataset of Spatio-Temporally Localized Atomic Visual Actions}}, 
  year={2018},}

@inproceedings{shang2019VidOR,
    title={{Annotating Objects and Relations in User-Generated Videos}},
    author={Shang, Xindi and Di, Donglin and Xiao, Junbin and Cao, Yu and Yang, Xun and Chua, Tat-Seng},
    booktitle={ICMR},
    year={2019},
}

@INPROCEEDINGS{Li2025LLaVA-ST,
  author={Li, Hongyu and Chen, Jinyu and Wei, Ziyu and Huang, Shaofei and Hui, Tianrui and Gao, Jialin and Wei, Xiaoming and Liu, Si},
  booktitle={CVPR}, 
  title={{LLaVA-ST: A Multimodal Large Language Model for Fine-Grained Spatial-Temporal Understanding}}, 
  year={2025},}

@InProceedings{Fujii2024Egosurgeryphase,
  title        = {{EgoSurgery-Phase: A Dataset of Surgical Phase Recognition from Egocentric Open Surgery Videos}},
  author       = {Ryo Fujii and Masashi Hatano and Hideo Saito and Hiroki Kajita},
  year         = {2024},
  booktitle    = {MICCAI},
}

@article{Fujii2024Egosurgerytool,
  title        = {{EgoSurgery-Tool: A Dataset of Surgical Tool and Hand Detection from Egocentric Open Surgery Videos}},
  author       = {Fujii, Ryo and Saito, Hideo and Kajita, Hiroki},
  journal      = {arXiv preprint arXiv:2406.03095},
  year         = {2024},
}

@article{Fujii2022Surgicaltool,
  title        = {{Surgical Tool Detection in Open Surgery Videos}},
  author       = {Fujii, Ryo and Hachiuma, Ryo and Kajita, Hiroki and Saito, Hideo},
  journal      = {Applied Sciences},
  year         = {2022},
}

@INPROCEEDINGS{Nwoye2025CholecTrack20,
  author={Nwoye, Chinedu Innocent and Elgohary, Kareem and Srinivas, Anvita and Zaid, Fauzan and Lavanchy, Joël L. and Padoy, Nicolas},
  booktitle={CVPR}, 
  title={{CholecTrack20: A Multi-Perspective Tracking Dataset for Surgical Tools}}, 
  year={2025},}

@InProceedings{Li2021MultiSports,
    author    = {Li, Yixuan and Chen, Lei and He, Runyu and Wang, Zhenzhi and Wu, Gangshan and Wang, Limin},
    title     = {{MultiSports: A Multi-Person Video Dataset of Spatio-Temporally Localized Sports Actions}},
    booktitle = {ICCV},
    year      = {2021},
}

@INPROCEEDINGS{Sultani2018UCFCrime,
  author={Sultani, Waqas and Chen, Chen and Shah, Mubarak},
  booktitle={CVPR}, 
  title={{Real-World Anomaly Detection in Surveillance Videos}}, 
  year={2018},}

@INPROCEEDINGS{Yuan2024UCA,
  author={Yuan, Tongtong and Zhang, Xuange and Liu, Kun and Liu, Bo and Chen, Chen and Jin, Jian and Jiao, Zhenzhen},
  booktitle={CVPR}, 
  title={{Towards Surveillance Video-and-Language Understanding: New Dataset, Baselines, and Challenges}}, 
  year={2024},
 }

@inproceedings{Ragusa2024Enigma51,
  title={{ENIGMA-51: Towards a Fine-Grained Understanding of Human Behavior in Industrial Scenarios}},
  author={Ragusa, Francesco and Leonardi, Rosario and Mazzamuto, Michele and Bonanno, Claudia and Scavo, Rosario and Furnari, Antonino and Farinella, Giovanni Maria},
  booktitle={WACV},
  year={2024}
}

@article{Francesco2023MECCANO,
title = {{MECCANO: A Multimodal Egocentric Dataset for Humans Behavior Understanding in the Industrial-like Domain }},
journal = {{CVIM}},
volume = {235},
pages = {103764},
year = {2023},
issn = {1077-3142},
doi = {https://doi.org/10.1016/j.cviu.2023.103764},
author = {Francesco Ragusa and Antonino Furnari and Giovanni Maria Farinella},}

@InProceedings{
    Ng2022AnimalKingdom,
    author    = {Ng, Xun Long and Ong, Kian Eng and Zheng, Qichen and Ni, Yun and Yeo, Si Yong and Liu, Jun},
    title     = {{Animal Kingdom: A Large and Diverse Dataset for Animal Behavior Understanding}},
    booktitle = {CVPR},
    year      = {2022},
 }

@ARTICLE{Yao2023DoTA,
  author={Yao, Yu and Wang, Xizi and Xu, Mingze and Pu, Zelin and Wang, Yuchen and Atkins, Ella and Crandall, David J.},
  journal={TPAMI}, 
  title={{DoTA: Unsupervised Detection of Traffic Anomaly in Driving Videos}}, 
  year={2023},
  volume={45},
  number={1},
  pages={444-459},}

@InProceedings{Yamaguchi2017STPR,
author = {Yamaguchi, Masataka and Saito, Kuniaki and Ushiku, Yoshitaka and Harada, Tatsuya},
title ={{Spatio-Temporal Person Retrieval via Natural Language Queries}},
booktitle = {ICCV},
year = {2017}
}

@INPROCEEDINGS{Kurita2023RefEgo,
  author={Kurita, Shuhei and Katsura, Naoki and Onami, Eri},
  booktitle={ICCV}, 
  title={{RefEgo: Referring Expression Comprehension Dataset from First-Person Perception of Ego4D}}, 
  year={2023},}

@inproceedings{
yao2026omnistvg,
title={{Omni{STVG}: Toward Spatio-Temporal Omni-Object Video Grounding}},
author={Jiali Yao and Xin Gu and Xinran Deng and Mengrui Dai and Bing Fan and Zhipeng Zhang and Yan Huang and Heng Fan and Libo Zhang},
booktitle={ICLR},
year={2026},
}

@inproceedings{Chen2019Weakly,
    title = {{Weakly-Supervised Spatio-Temporally Grounding Natural Sentence in Video}},
    author = "Chen, Zhenfang  and
      Ma, Lin  and
      Luo, Wenhan  and
      Wong, Kwan-Yee Kenneth",
    booktitle = "ACL",
    year = "2019",

}

@inproceedings{wang2024internvideo2,
  title={{InternVideo2: Scaling Foundation Models for Multimodal Video Understanding}},
  author={Wang, Yi and Li, Kunchang and Li, Xinhao and Yu, Jiashuo and He, Yinan and Chen, Guo and Pei, Baoqi and Zheng, Rongkun and Wang, Zun and Shi, Yansong and others},
  booktitle={ECCV},
  year={2024},
}

@inproceedings{reimers-gurevych-2019-sentence,
    title = {{Sentence-{BERT}: Sentence Embeddings using {S}iamese {BERT}-Networks}},
    author = "Reimers, Nils  and
      Gurevych, Iryna",
    booktitle = "EMNLP-IJCNLP",
    year = "2019",

}

@INPROCEEDINGS{Yang2022TubeDETR,
  author={Yang, Antoine and Miech, Antoine and Sivic, Josef and Laptev, Ivan and Schmid, Cordelia},
  booktitle={CVPR}, 
  title={{TubeDETR: Spatio-Temporal Video Grounding with Transformers}}, 
  year={2022},}

@INPROCEEDINGS{Huang2024VTimeLLM,
  author={Huang, Bin and Wang, Xin and Chen, Hong and Song, Zihan and Zhu, Wenwu},
  booktitle={CVPR}, 
  title={{VTimeLLM: Empower LLM to Grasp Video Moments}}, 
  year={2024},}

@INPROCEEDINGS{Ren2024TimeChat,
  author={Ren, Shuhuai and Yao, Linli and Li, Shicheng and Sun, Xu and Hou, Lu},
  booktitle={CVPR}, 
  title={{TimeChat: A Time-sensitive Multimodal Large Language Model for Long Video Understanding}}, 
  year={2024},}

@inproceedings{Liu2024GroundingDINO,
author = {Liu, Shilong and Zeng, Zhaoyang and Ren, Tianhe and Li, Feng and Zhang, Hao and Yang, Jie and Jiang, Qing and Li, Chunyuan and Yang, Jianwei and Su, Hang and Zhu, Jun and Zhang, Lei},
title = {{Grounding DINO: Marrying DINO with Grounded Pre-Training for Open-Set Object Detection}},
year = {2024},
booktitle = {ECCV},
}

@inproceedings{
ravi2025sam,
title={{{SAM} 2: Segment Anything in Images and Videos}},
author={Nikhila Ravi and Valentin Gabeur and Yuan-Ting Hu and Ronghang Hu and Chaitanya Ryali and Tengyu Ma and Haitham Khedr and Roman R{\"a}dle and Chloe Rolland and Laura Gustafson and Eric Mintun and Junting Pan and Kalyan Vasudev Alwala and Nicolas Carion and Chao-Yuan Wu and Ross Girshick and Piotr Dollar and Christoph Feichtenhofer},
booktitle={ICLR},
year={2025},
}

@InProceedings{Heo2025CVPR,
    author    = {Heo, Miran and Chen, Min-Hung and Huang, De-An and Liu, Sifei and Radhakrishnan, Subhashree and Kim, Seon Joo and Wang, Yu-Chiang Frank and Hachiuma, Ryo},
title     = {{Omni-RGPT: Unifying Image and Video Region-level Understanding via Token Marks}},
    booktitle = {CVPR},
    year      = {2025}
}

@article{shen2025zoomzero,
      title={{Zoom-Zero: Reinforced Coarse-to-Fine Video Understanding via Temporal Zoom-in}}, 
      author={Xiaoqian Shen and Min-Hung Chen and Yu-Chiang Frank Wang and Mohamed Elhoseiny and Ryo Hachiuma},
      year={2025},
      journal={arXiv preprint arXiv:2512.14273},
}

@InProceedings{Xiao_2024_CVPR,
    author    = {Xiao, Junbin and Yao, Angela and Li, Yicong and Chua, Tat-Seng},
    title     = {{Can I Trust Your Answer? Visually Grounded Video Question Answering}},
    booktitle = {CVPR},
    year      = {2024}
}

@article{yan2025videochatr15,
      title={{VideoChat-R1.5: Visual Test-Time Scaling to Reinforce Multimodal Reasoning by Iterative Perception}}, 
      author={Ziang Yan and Xinhao Li and Yinan He and Zhengrong Yue and Xiangyu Zeng and Yali Wang and Yu Qiao and Limin Wang and Yi Wang},
      year={2025},
      journal={arXiv preprint arXiv:2509.21100},
}

@InProceedings{Kim_2023_WACV,
    author    = {Kim, Dahye and Park, Jungin and Lee, Jiyoung and Park, Seongheon and Sohn, Kwanghoon},
    title     = {{Language-Free Training for Zero-Shot Video Grounding}},
    booktitle = {WACV},
    year      = {2023}
}

@InProceedings{Pramanick_2025_ICCV,
    author    = {{Pramanick, Shraman and Mavroudi, Effrosyni and Song, Yale and Chellappa, Rama and Torresani, Lorenzo and Afouras, Triantafyllos}},
    title     = {{Enrich and Detect: Video Temporal Grounding with Multimodal LLMs}},
    booktitle = {ICCV},
    year      = {2025}
}

@InProceedings{liu2024doraweightdecomposed,
      title={{DoRA: Weight-Decomposed Low-Rank Adaptation}}, 
      author={Shih-Yang Liu and Chien-Yi Wang and Hongxu Yin and Pavlo Molchanov and Yu-Chiang Frank Wang and Kwang-Ting Cheng and Min-Hung Chen},
      year={2024},
      booktitle={ICML},
}

@inproceedings{hu2022lora,
title={{Lo{RA}: Low-Rank Adaptation of Large Language Models}},
author={Edward J Hu and yelong shen and Phillip Wallis and Zeyuan Allen-Zhu and Yuanzhi Li and Shean Wang and Lu Wang and Weizhu Chen},
booktitle={ICLR},
year={2022},
}

@inproceedings{
agrawal2025gepa,
title={{{GEPA}: Reflective Prompt Evolution Can Outperform Reinforcement Learning}},
author={Lakshya A Agrawal and Shangyin Tan and Dilara Soylu and Noah Ziems and Rishi Khare and Krista Opsahl-Ong and Arnav Singhvi and Herumb Shandilya and Michael J Ryan and Meng Jiang and Christopher Potts and Koushik Sen and Alexandros G. Dimakis and Ion Stoica and Dan Klein and Matei Zaharia and Omar Khattab},
booktitle={ICLR},
year={2026},
}

@article{shao2024deepseekmath,
  title={{Deepseekmath: Pushing the limits of mathematical reasoning in open language models}},
  author={Shao, Zhihong and Wang, Peiyi and Zhu, Qihao and Xu, Runxin and Song, Junxiao and Bi, Xiao and Zhang, Haowei and Zhang, Mingchuan and Li, YK and Wu, Yang and others},
  journal={arXiv preprint arXiv:2402.03300},
  year={2024}
}

@misc{li2026videothinkerbuilds,
      title={{VideoThinker: Building Agentic VideoLLMs with LLM-Guided Tool Reasoning}}, 
      author={Chenglin Li and Qianglong Chen and Feng Han and Yikun Wang and Xingxi Yin and Yan Gong and Ruilin Li and Yin Zhang and Jiaqi Wang},
      year={2026},
      journal={arXiv preprint 2601.15724}
}

@article{kalueff2016neurobiology,
  title={{Neurobiology of Rodent Self-Grooming and Its Value for Translational Neuroscience}},
  author={Kalueff, Allan V and Stewart, Adam Michael and Song, Cai and Berridge, Kent C and Graybiel, Ann M and Fentress, John C},
  journal={Nature Reviews Neuroscience},
  volume={17},
  number={1},
  pages={45--59},
  year={2016},
  publisher={Nature Publishing Group UK London}
}

@inproceedings{jia2022vpt,
  title={{Visual Prompt Tuning}},
  author={Jia, Menglin and Tang, Luming and Chen, Bor-Chun and Cardie, Claire and Belongie, Serge and Hariharan, Bharath and Lim, Ser-Nam},
  booktitle={ECCV},
  year={2022},
}

@inproceedings{fu2025videomme,
      title={{Video-MME: The First-Ever Comprehensive Evaluation Benchmark of Multi-modal LLMs in Video Analysis}}, 
      author={Chaoyou Fu and Yuhan Dai and Yongdong Luo and Lei Li and Shuhuai Ren and Renrui Zhang and Zihan Wang and Chenyu Zhou and Yunhang Shen and Mengdan Zhang and Peixian Chen and Yanwei Li and Shaohui Lin and Sirui Zhao and Ke Li and Tong Xu and Xiawu Zheng and Enhong Chen and Caifeng Shan and Ran He and Xing Sun},
      year={2025},
      booktitle={CVPR}
}

@inproceedings{zhou2024mlvu,
    author    = {Zhou, Junjie and Shu, Yan and Zhao, Bo and Wu, Boya and Liang, Zhengyang and Xiao, Shitao and Qin, Minghao and Yang, Xi and Xiong, Yongping and Zhang, Bo and Huang, Tiejun and Liu, Zheng},
    title     = {{MLVU: Benchmarking Multi-task Long Video Understanding}},
    booktitle = {CVPR},
    year      = {2025},
}

@inproceedings{yu2024eliciting,
    title = {{Eliciting In-Context Learning in Vision-Language Models for Videos Through Curated Data Distributional Properties}},
    author = "Yu, Keunwoo Peter  and
      Zhang, Zheyuan  and
      Hu, Fengyuan  and
      Storks, Shane  and
      Chai, Joyce",
    booktitle = "EMNLP",
    year = "2024",
}

@article{dong2026demoicl,
  title={{Demo-ICL: In-Context Learning for Procedural Video Knowledge Acquisition}}, 
  author={Dong, Yuhao and Tian, Shulin and Liu, Shuai and Ding, Shuangrui and Zang, Yuhang and Dong, Xiaoyi and Cao, Yuhang and Wang, Jiaqi and Liu, Ziwei},
  journal={arXiv preprint arXiv:2602.08439},
  year={2026}
}

@article{xue2026personal,
  title={{Personal Visual Context Learning in Large Multimodal Models}},
  author={Xue, Zihui and Baid, Ami and Kim, Sangho and Luo, Mi and Grauman, Kristen},
  journal={arXiv preprint arXiv:2605.10936},
  year={2026}
}

@inproceedings{
shi2026videoloom,
title={{VideoLoom: A Video Large Language Model for Joint Spatial-Temporal Understanding}},
author={Jiapeng Shi and Junke Wang and Zuyao You and Bo He and Zuxuan Wu},
booktitle={ICML},
year={2026},
}

@article{gwet2008computing,
  title={{Computing Inter-Rater Reliability and Its Variance in the Presence of High Agreement}},
  author={Kilem L. Gwet},
  journal={The British journal of mathematical and statistical psychology},
  year={2008},
  volume={61 Pt 1},
  pages={
          29-48
        },
}

@article{cohen1960coefficient,
  title={{A Coefficient of Agreement for Nominal Scales}},
  author={Jacob Cohen},
  journal={Educational and Psychological Measurement},
  year={1960},
  volume={20},
  pages={37 - 46},
}

@article{landis1977measurement,
  title={{The Measurement of Observer Agreement for Categorical Data}},
  author={J. Richard Landis and Gary G. Koch},
  journal={Biometrics},
  year={1977},
  volume={33},
  number={1},
  pages={159--174},
}

@article{feinstein1990high,
  title={High Agreement but Low Kappa: I. The Problems of Two Paradoxes},
  author={Alvan R. Feinstein and Domenic V. Cicchetti},
  journal={Journal of Clinical Epidemiology},
  year={1990},
  volume={43},
  number={6},
  pages={543--549},
}
